\documentclass[12pt, final]{l4dc2023}

\title[Hybrid Systems Neural Control]{Hybrid Systems Neural Control with Region-of-Attraction Planner}
\usepackage{times}

\author{%
 \Name{Yue Meng} \Email{mengyue@mit.edu} \AND
 \Name{Chuchu Fan} \Email{chuchu@mit.edu}\\
 \addr Massachusetts Institute of Technology %
}

\usepackage{algorithm}
\usepackage[noend]{algpseudocode}
\usepackage{amsfonts}

\usepackage{wrapfig}

\usepackage{hyperref}

\usepackage{bbm}
\newcommand{\figref}[1]{Fig.~\ref{#1}}
\renewcommand{\eqref}[1]{Eq.~\ref{#1}}

\newcommand{\appref}[1]{\href{https://mit-realm.github.io/hybrid-clf/static/appendix.pdf}{\citep{l4dcappendix}[Appx.~\ref*{#1}]}}
\newcommand{\apprefref}[2]{\href{https://mit-realm.github.io/hybrid-clf/static/appendix.pdf}{\citep{l4dcappendix}[Appx.~\ref*{#1}$\sim$\ref*{#2}]}}
\newcommand{\secref}[1]{Sec.~\ref{#1}}
\newcommand{\defref}[1]{Def.~\ref{#1}}
\newcommand{\propref}[1]{Prop.~\ref{#1}}
\newcommand{\theoref}[1]{Theo.~\ref{#1}}
\newcommand{\algorref}[1]{Algor.~\ref{#1}}

\newcommand{\Noindent}{\noindent}
\newcommand{\partitle}[1]{\Noindent\textbf{#1: }}

\newcommand{\postac}[1]{\color{black}{#1}\color{black}}

\begin{document}

\maketitle

\begin{abstract}
Hybrid systems are prevalent in robotics. However, ensuring the stability of hybrid systems is challenging due to sophisticated continuous and discrete dynamics. A system with all its \postac{system modes } stable can still be unstable. Hence special treatments are required at mode switchings to stabilize the system. In this work, we propose a hierarchical, neural network (NN)-based method to control general hybrid systems. For each \postac{system mode}, we first learn an NN Lyapunov function and an NN controller to ensure the states within the region of attraction (RoA) can be stabilized. Then an RoA NN estimator is learned across different modes. Upon mode switching, we propose a differentiable planner to ensure the states after switching can land in next mode's RoA, hence stabilizing the hybrid system. We provide novel theoretical stability guarantees and conduct experiments in car tracking control, pogobot navigation, and bipedal walker locomotion. Our method only requires 0.25X of the training time as needed by other learning-based methods. With low running time (10$\sim$50X faster than model predictive control (MPC)), our controller achieves a higher stability/success rate over other baselines such as MPC, reinforcement learning (RL), common Lyapunov methods (CLF), linear quadratic regulator (LQR), quadratic programming (QP) and Hamilton-Jacobian-based methods (HJB). The project page is on \url{https:/mit-realm.github.io/hybrid-clf}.
\end{abstract}

\begin{keywords}%
  Hybrid system, Control Lyapunov functions, Region of Attraction%
\end{keywords}
\section{Introduction}
Learning how to control hybrid systems is critical in the realm of robotics and artificial intelligence, given the wide variety of hybrid systems in autonomous driving~\citep{ning2021survey}, locomotion for bipedal robots, and UAVs~\citep{gillula2011applications}. 
However, it remains challenging to analyze the stability and design controllers for general nonlinear hybrid systems, due to the intricate dynamics involving both the continuous flows and discrete jumps~\citep{chen2021learning}.

Various approaches in classic control emerged to analyze the stability for a certain type of hybrid systems, such as piecewise affine (PWA) systems~\citep{johansson2003piecewise,pettersson1996stability,pettersson1999exponential,prajna2003analysis} and periodic systems~\citep{poincare1885courbes,clark2018poincar,manchester2011transverse,manchester2011regions}. However, these methods either work on simple systems in low dimensions or rely on computation-heavy methods for verification and synthesis~\citep{abate2020formal, jarvis2003some,topcu2008local,majumdar2013control}. 

The pivot hurdle for stabilizing general hybrid systems is to properly handle the system at the mode switching~\citep{de2009survey}. A hybrid system, provided with all its \postac{system modes stable}, can still be unstable if the mode switching is too fast~\citep{branicky1998multiple}. Using multiple Lyapunov functions, some works constraint the average dwell time (ADT)~\citep{hespanha1999stability,zhai2000piecewise} so the system switching is ``slow-on-the-average'', but they cannot handle discrete jumps. Other methods enforce the sub-sequence of each Lyapunov function at switch-in instants decreasing~\citep{branicky1998multiple}. Those methods track each Lyapunov function's switching sequence, which is hard to implement when there are many system modes. 

\begin{figure}[!tbp]
\floatconts{fig:roa-teaser}
{\caption{We learn the stabilizing control and the RoA for each \postac{system mode}. In testing phase, our method (red lines) plans the configuration (the exiting state and the mode) to ensure the next entering state is always within the RoA of the next equilibrium, whereas traditional methods (gray lines) directly tracking the next equilibrium will diverge after the jump.}}
{
\includegraphics[width=0.95\textwidth]{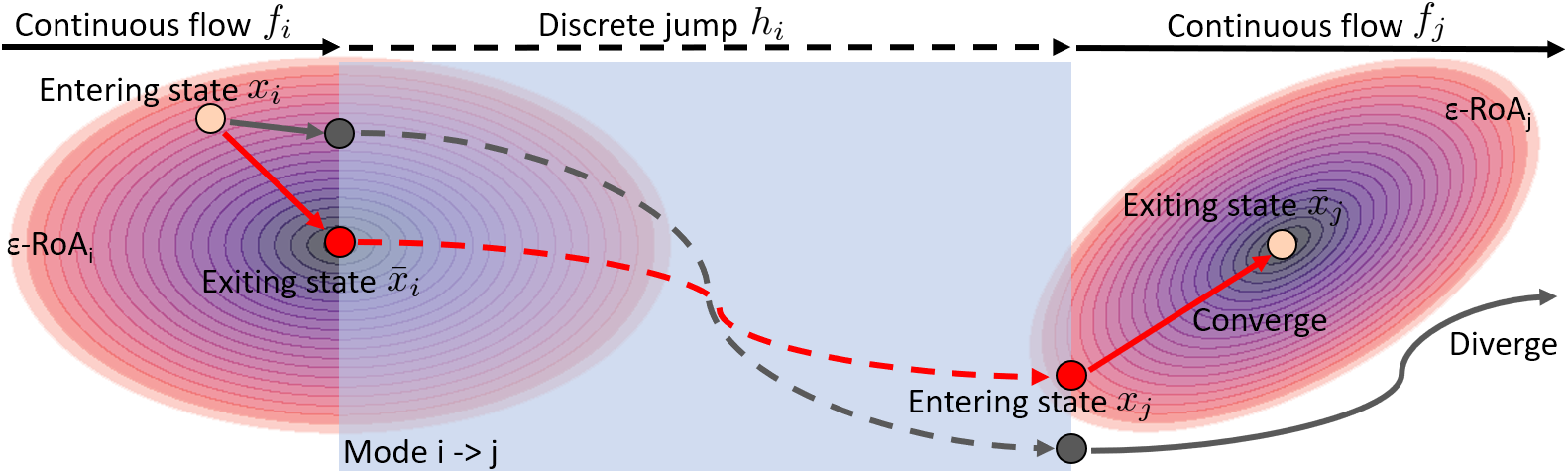}
}
\end{figure}

Motivated by region of attraction (RoA)-based planning methods~\citep{tedrake2010lqr}, we propose an RoA-based approach to stabilize hybrid systems. The idea is to let the state always enter next mode's RoA after switching. We ensure \postac{the stability of the system under each mode } by using control Lyapunov functions (CLF), and we use the invariant set provided by the Lyapunov function to represent the RoA for \postac{the system under each mode}. In this way, we don't need to optimize the whole sequence of the Lyapunov functions but only consider the Lyapunov function (and RoA) in consecutive modes, which is more efficient to implement for stability analysis and controller synthesis. Constructing Lyapunov functions used to be time-consuming and is limited to low-dimension systems~\citep{abate2020formal, jarvis2003some,topcu2008local,majumdar2013control}. Recently there is a trend in using neural networks to construct control Lyapunov and Barrier certificates~\citep{chang2019neural,dawson2022safecontrol, qin2021learning, meng2021reactive, sun2020learning}. We follow this direction and further use neural networks for RoA estimation. 

The whole pipeline is as follows: we first collect samples under \postac{each system mode}, use the neural network controller to generate trajectories, and construct neural Lyapunov certificates to ensure the stability for each mode. Then we estimate the RoA by the Lyapunov level-set and learn \postac{each system mode's } stable level-set using the Neural RoA estimator. Finally, upon deployment, we use a differentiable planner to find the optimal configuration before mode switching to ensure the next state will be within the RoA of the next mode, hence achieving the stability of the hybrid system. Although there might exist a gap between theoretical guarantees and simulation performance due to  imperfect learning of neural networks, in practice we observe strong empirical results.

We conduct experiments on three challenging scenarios (car tracking control on different road conditions, pogobot navigation, and bipedal walker locomotion). We achieve the best performance (mean square error, failure rate, etc) compared to other baselines (MPC, RL, LQR, CLF, QP, HJB). Our learned method requires less training time than RL-based methods or the HJB approach, and at the evaluation stage, the runtime is only 1/50X$\sim$1/10X the time for MPC.

Our contributions are:
(1) we are the first to use a neural-network RoA estimator, planner and controller to stabilize hybrid systems \postac{within certain RoA}. (2) we \postac{define } a new stability for hybrid systems and derive sufficient conditions to enforce the stability with theoretical guarantees (3) our approach can be applied to different hybrid systems \postac{even if the dynamics is unknown}, where the controller can be state-feedback control or apex-to-apex control for periodic systems. (4) we conduct challenging experiments that involve complicated hybrid systems and outperform other baselines.

\section{Related Work}

Controlling hybrid systems is challenging and has been studied for decades. Various formulations for system modelling and control strategies have been extensively presented in ~\citep{decarlo2000perspectives,de2009survey,davrazos2001review,sanfelice2013control}. We only list a few closely related works in stability analysis and controller synthesis.

\partitle{Lyapunov stability analysis} 
The common Lyapunov function is used to prove hybrid system stability in ~\citep{dogruel1994stability,fierro1997hybrid}
. Multiple Lyapunov functions are proposed in ~\citep{peleties1992asymptotic} for linear systems and ~\citep{branicky1998multiple,michel1999towards} for more general cases where the monotonous decreasing condition is relaxed. In~\citep{malmborg1998analysis}, non-smooth Lyapunov functions are used for hybrid controller synthesis. 
For piecewise affine (PWA) systems, linear matrix inequality based approaches are proposed to construct Lyapunov certificates~\citep{johansson2003piecewise,pettersson1996stability}. 
For switched systems, average dwell time (ADT) is introduced in~\citep{hespanha1999stability,zhai2000piecewise} to tie the stability with the ratio between stable and unstable modes. For periodic systems, Poincar\'{e} map~\citep{clark2018poincar} and transverse Lyapunov functions~\citep{manchester2011transverse} are used to analysis the limit cycle stability. Although equipped with theoretical guarantees for the stability, the methods above are either only suited for a specified type of systems (linear or PWA) or require expert knowledge to design the Lyapunov functions, or rely on tedious numerical methods such as Satisfiability Modulo Theories (SMT) solvers~\citep{abate2020formal} and sum-of-squares (SoS) optimization~\citep{jarvis2003some,topcu2008local}. For complicated systems, recently there is a growing trend to approximate the Lyapunov functions and the controllers using data-driven methods like Gaussian Process~\citep{zhai2019region,xiao2020learning}, SVM~\citep{sun2005support} and neural networks~\citep{richards2018lyapunov,chang2019neural,mehrjou2020neural,dawson2022safecontrol,dawson2022safe}. We are aligned with this and further study to stabilize hybrid systems using neural networks.

\partitle{Hybrid system control designs} Upon grounding work on Lyapunov theory~\citep{sontag1983lyapunov,artstein1983stabilization}, a wide body of literature exists on synthesizing feedback controllers for switched linear and affine systems~\citep{wicks1997solution,johansson1999piecewise,mignone2000stability}. There are also many works beyond Lyapunov controllers for more general hybrid systems, such as optimal control~\citep{cassandras2001optimal,cho2001forward}, model predictive control (MPC)~\citep{slupphaug1997model,lazar2006model,camacho2010model,marcucci2020warm} and Hamilton-Jacobian reachability-based (HJB) methods~\citep{choi2022computation}, and region-of-attraction (RoA) based approaches~\citep{tedrake2010lqr,bhounsule2018switching,zamani2019feedback}. However, these methods are often computational expensive for high dimension hybrid systems. There exist reinforcement learning (RL) methods to control hybrid systems like legged robots~\citep{benbrahim1997biped,morimoto2004simple,neunert2020continuous,mastrogeorgiou2020slope}, but finding the appropriate RL methods and rewards are extremely challenging and may cause undesired behaviors learnt to hack for high returns~\citep{clark2016faulty}. Our method shares similar philosophy with RoA-based works~\citep{tedrake2010lqr,bhounsule2018switching,zamani2019feedback}, whereas ours is computation-efficient, suits general nonlinear hybrid systems and can represent RoAs for arbitrary number of modes.

\section{Problem Formulation}

\begin{definition}[Controlled Hybrid Systems]
A \textit{controlled hybrid system} is defined as: 
\begin{equation}
    \begin{cases}
    \dot{x}=f_i(x, u; p_i), \quad x\in \mathcal{C}_i \\
    x^{+} = h_i(x, u; p_i, p_j), \quad x\in \mathcal{D}_{i,j} \\
    \end{cases}
\end{equation}
where $x\in\mathbb{R}^{n}$ denotes the system state, $u\in\mathbb{R}^m$ denotes the control input, and $p_i\in\mathcal{P}$ denotes the system configuration (e.g. the set point, reference velocity). The index $i=1,...,I$ denotes the system mode. $\mathcal{C}_i$ is the flow set where states follow continuous flow map $f_i:\mathbb{R}^n\times\mathbb{R}^m\times \mathcal{P} \to\mathbb{R}^n$, and $\mathcal{D}_{i,j}$ is the jump set where states follow discrete jump map $h_i:\mathbb{R}^n\times \mathbb{R}^m\times \mathcal{P}\times\mathcal{P}\to\mathbb{R}^n$. 
\label{def:hybrid-sys}
\end{definition}


We aim to stabilize the hybrid system in \defref{def:hybrid-sys}. For each \postac{mode of system } $\dot{x}=f_i(x,u;p_i)$ with equilibrium $x^*_i$, we consider the stability defined in~\citep{khalil2009lyapunov}. Then the sufficient conditions for \postac{each mode } system stability are as follows (we omit the index $i$ and the configuration $p$ for brevity).

\begin{proposition}[Control Lyapunov Conditions for System Stability] The system $\dot{x} = f(x,u)$ with equilibrium $x^*$ is asymptotically stable at $x^*$ if there exist a differentiable function $V: \mathcal{C}\to\mathbb{R}$ and a control policy $u=\pi(x)$ such that: $V(x^*)=0;\, \text{and}\, \forall x\in \mathcal{C} \backslash \{x^*\},\, V(x)>0,\, \text{and}\, \frac{dV}{dx} f(x, u)< 0$.
The $V$ is called a \textit{control Lyapunov function (CLF)}. The system is exponentially stable at $x^*$ if the $V$ further satisfies: $\exists \gamma>0,\, \forall x\in \mathcal{C} \backslash \{x^*\},\, \frac{dV}{dx} f(x, u)< -\gamma V$.
\label{prop-1}
\end{proposition}
The proof can be found in~\citep{isidori1985nonlinear}[Theorem 9.4.1]. If each \postac{system mode } is stable with valid Lyapunov functions, will the hybrid system converge? Unfortunately, the answer is no with a counter-example in~\citep{branicky1998multiple}. 
The culprit is that the system switches too fast: although the Lyapunov value is decreasing in each mode, the distance toward equilibrium is not decreasing yet. 
For systems with jumps, it is more unlikely to ensure the stability, as there might be jumps making $||x(t)-x^*||\geq \epsilon$ infinitely often. Thus, 
we propose a new stability for the hybrid systems.

\begin{definition}[Hybrid System Stability] Given $\epsilon=\{\epsilon_i\}$, a hybrid system is $\epsilon$-asymptotically stable (exponentially stable), if each \postac{system mode } is asymptotically stable (exponentially stable), and all states $\bar{x}_i$ exiting the mode $i$ are within $\epsilon_i$-ball of the $x^*$, i.e., $||\bar{x}_i-x^*||\leq \epsilon_i$. We call $\bar{x}_i$ as \textit{exiting state} (and call state $x_i$ that enters the mode $i$ as \textit{entering state}).
\label{def:e-stable}
\end{definition}
If $\epsilon$ is constant, the exiting states will be at most $\epsilon$-far away from the equilibrium. If $\epsilon$ converges to zero, the sequence of the exiting states will converge to the equilibrium \postac{hence asymptotic stability is achieved}. In this paper, we consider the former case. 
With constant $\epsilon$, we define $\epsilon$-RoA as follows.

\begin{definition}[$\epsilon$-Region of Attraction] The $\epsilon$-RoA for $x^*$ in mode $i$  is defined as:
$\mathcal{R}^\epsilon = \{x_0|, x(0)=x_0, ||\bar{x}_i-x^*||\leq \epsilon \}$
, where $\bar{x}_i$ is the exiting state (for mode $i$) starting at $x_0$. 
\label{def:e-roa}
\end{definition}
\postac{Using $\epsilon$-RoA, we do not need to check dwell time conditions like~\citep{hespanha1999stability}}. To efficiently check if $x\in\epsilon$-RoA, \postac{we seek a set representation for $\epsilon$-RoA and a scalar function/index to tell whether $x$ is in the set}. We use Lyapunov level set to (conservatively) represent $\epsilon$-RoA.
 \begin{definition}[Maximum $\epsilon$-Stable Level Set] For mode $i$ and configuration $p_i$, the largest level set $\mathcal{S}^{c_i}=\{x|V_i(x, p_i)\leq c_i(p_i)\}$ within $\epsilon$-RoA is called \textit{Maximum $\epsilon$-Stable Level Set}, and $c_i(p_i)$ is called \textit{Maximum $\epsilon$-Stable Level Set index}. \label{def:e-level-set}
\end{definition}
For an entering state $x_j$, we have $V_j(x_j,p_j)\leq c_j(p_j) \to x_j\in\epsilon$-RoA. One step ahead, at mode $i$ with switching $i\to j$, it requires $V_j(h_i(\bar{x}_i,u;p_i,p_j))\leq c_j(p_j)$. Propagating from $\bar{x}_i$ to $x^*$, we derive sufficient conditions for hybrid system stability proved in \footnote{The appendix is at \url{https://mit-realm.github.io/hybrid-clf/static/appendix.pdf}}{\appref{appendix-prop2}}.

 

 
  \begin{theorem}[Lyapunov Conditions for Hybrid System $\epsilon$-Stability] A hybrid system in \defref{def:hybrid-sys} is $\epsilon$-exponentially stable, if there exists a Lyapunov function $V_i$ for each mode $i$ (and configuration $p_i$) satisfying all conditions in \propref{prop-1} and $\alpha ||x-x^*||\leq V_i(x, p_i)\leq \beta ||x-x^*||$, and for each entering state $x_i$ that moves toward the mode $j$, we have the $p_i, c_i, p_j, c_j$ to satisfy: \begin{equation}V_i(x_i, p_i) \leq c_i(p_i) \text{ and } V_j(h_i(x^*, u; p_i, p_j), p_j)\leq \Upsilon
  \label{eq:v-condition}
  \end{equation}
  where $\Upsilon=\frac{\alpha_j}{\beta_j} c_j(p_j) - \alpha_j K_i \epsilon$, and $c_i$, $c_j$ are the maximum $\epsilon$-stable level set indices for modes $i$ and $j$, and $K_i$ is the local Lipschitz constant for function $h_i$ within $\epsilon$-ball of $x^*$. 
  \label{theo-2}
  \end{theorem}
 
 In short, \theoref{theo-2} guarantees each \postac{system mode } is exponentially stable, and at switching $i\to j$, the entering state $x_i$ is within $\epsilon$-RoA for mode $i$, and the next entering state $x_j$ is also inside $\epsilon$-RoA for mode $j$. The Lipschitz constant $K_i$ can be approximated by $|\frac{\partial h_i}{\partial x}|$ at $(x^*,u;p_i,p_j)$ for small $\epsilon$. 
 In the next section, we will use neural networks to satisfy \theoref{theo-2} for hybrid system stability.
\section{Methodology}
To control the system in \defref{def:hybrid-sys}, we propose a learning framework in \algorref{alg:hybrid-control}. Guided by \theoref{theo-2}, we first learn the controller and the CLF to stabilize the \postac{system } within each mode. Then we perform the RoA estimation, and finally, we bring in the differentiable planner that leverages the controller and RoA estimator to enforce the hybrid system stability.

\subsection{Learning neural Lyapunov functions and controllers}
We train neural networks (NN) to synthesize the control Lyapunov functions $V(x,p)$ and controllers $u=\pi(x,p)$ for each \postac{system mode}. During training, we use the controller to roll out trajectories from uniformly sampled initial states and compute the CLF values. 
To satisfy the CLF conditions in \theoref{theo-2}, 
we design the Lyapunov function as $V(x,p)=||P_{\text{NN}}(p)(x-x^*)||+V_\text{NN}(x,p)^2 ||x-x^*||^2$, where $P_\text{NN}(p)$ is an NN that takes $p$ as input and outputs a matrix, and $V_{\text{NN}}(x,p)$ is an NN taking $x,p$ as inputs and outputs a scalar. In this way, 
the positive definiteness and the norm inequalities of the Lyapunov functions are naturally satisfied. 
Finally, to enforce the Lyapunov value decreasing (exponentially) along the trajectories $\mathcal{S}$, we design the loss: 
\begin{equation}
    \mathcal{L}_{clf}=\sum\limits_{x\in\mathcal{S}, p\in\mathcal{P}} \text{ReLU}\Big(\gamma V(x, p)+\frac{\partial V(x,p)}{\partial x} f(x, u)\Big)
    \label{eq:loss}
\end{equation}
where $u=\pi(x,p)$ and we compute the derivative term $\frac{\partial V(x_t,p)}{\partial x_t} f(x_t,u)\approx (V_{t+1}-V_t)/dt$.

\subsection{Learning neural RoA estimator}
\label{sec:roa-est}
After training for the CLF and controllers, we estimate the maximum $\epsilon$-stable level set in \defref{def:e-level-set} for any given $p$ using: $c^*(p)=\max\Big\{c \,\Big|\, \forall x_0\in \mathcal{C}_i,\, V(x_0,p)\leq c\rightarrow x_0 \in \mathcal{R}^{\epsilon}  \Big\}$
, where $\mathcal{R}^{\epsilon}$ is the $\epsilon$-RoA in \defref{def:e-roa}. For each $p_i$, we set $\epsilon=10^{-2}$ and uniformly sample $10^3 \sim 10^4$ initial states, roll out trajectories and find the largest Lyapunov value $c_i^*$ for all the initial states that have exiting states within the $\epsilon$-ball of the equilibrium. We train the NN RoA Estimator $R_{\text{NN}}(p)$ with:
\begin{equation}
    \mathcal{L}_{roa}=\sum\limits_{(p_i, c_i^*)\sim\mathcal{Z}} \Big(R_{\text{NN}}(p_i) - c_i^*\Big)^2 
    \label{eq:loss-roa}
\end{equation}
where $\mathcal{Z}$ is the set for simulated trajectories and maximum $\epsilon$-Stable level set indices. In this way, the RoA under configuration $p$ can be approximated by $\big\{x \ | \ V(x,p)\leq R_{\text{NN}}(p)\big\}$.

\subsection{Differentiable configuration planner}
We use a differentiable planner to ensure switching stability by satisfying the last condition in \eqref{eq:v-condition} of \theoref{theo-2}. Assume $p_j$ is given at mode $i$. We use gradient descent to find $p_i$ to minimize:
\begin{equation}
\begin{aligned}
    \mathcal{L}_{e}& =\text{ReLU}\Big(V_i(x_i, p_i) - R_{\text{NN}}(p_i)\Big) +\text{ReLU}\Big(V_j(h_i(x^*, u, p_i), p_j) - \eta R_{\text{NN}}(p_j) + \kappa \Big)
    \label{eq:loss-e}
\end{aligned}
\end{equation}
where $\eta=\frac{\alpha_j}{\beta_j}$ and $\kappa=\alpha_j K_i \epsilon$. It is time-consuming to evaluate $\alpha_j$, $\beta_j$ and $K_i$ for each $p_i$ and $p_j$. Empirically, we set $\eta$=0.9 and $\kappa$=$10^{-2}$ for $\epsilon$=$10^{-2}$ (ablation in~\appref{appendix-abl-hyper}).  The optimization converges after 3$\sim$10 steps, adding only little runtime overhead. 
We can also use a new loss $\mathcal{L}_{e}^+$ for periodic systems with decomposable jump map (common in error-state dynamics):
\begin{equation}
    h_i(x,u;p_i,p_j)=h^+(x,p_i)+p_i-p_j,\, h^+(x^*,p_i)=x^*
    \label{eq:special-sys}
\end{equation}
where the heuristic is to guide the configuration gradually approaches the target configuration $p_j$:  
\begin{equation}
    \mathcal{L}_{e}^+ = \text{ReLU}\Big(V_i(x_i, p_i) - R_{\text{NN}}(p_i)\Big) + \lambda ||p_j-p_i||
    \label{eq:loss-e-heur}
\end{equation}
We guarantee it converges to the target $p_j$ in finite steps (proved in \appref{appendix-prop3}):

\begin{theorem}[Finite-step converging toward the target configuration]
For special hybrid systems with $h_i$ defined in \eqref{eq:special-sys}. If $c_m(p_m)\geq \beta_m K_m \epsilon$ for all modes $m$ and $p_m$, and assume the first term in \eqref{eq:loss-e-heur} is zero for the optimal $p$, then the system is $\epsilon$-stable and the configuration is guaranteed to reach the target configuration $p_j$ in finite switches $N\leq \left\lceil\frac{||p_j-p_i||}{\min\limits_{m}\frac{c_m}{\beta_m}-K_{m} \epsilon}\right\rceil$. 
\label{theo-3}
\end{theorem}

\begin{algorithm}[H]
\scriptsize
    \SetAlgoNoLine
    \SetKwFor{While}{while}{do}{}
    \SetKwFor{ForAll}{foreach}{do}{}
    \SetKwProg{Fn}{Function}{}{}
    

    


    

    Sample $\{f_i, p_i\}_{i=1}^N$ from the hybrid system in \defref{def:hybrid-sys} \tcp*{Training phase}

    \ForAll{$f_i$, $p_i$}{
        Train $ V_i$ and $\pi_i$ with the loss $\mathcal{L}_{clf}$ in \eqref{eq:loss}
        
        Find $c_i^*$ for $V_i$ and $\pi_i$ (according to~\secref{sec:roa-est})
    }
    Train $R_{NN}$ on $\{(p_i, c_i^*)\}_{i=1}^N$ with the loss $\mathcal{L}_{roa}$ in \eqref{eq:loss-roa}

    Initialize state $x(0)\gets x_0$, mode $k$ and time $t\gets 0$ \tcp*{Testing phase}

    \While{$t\leq T$}{
    \uIf{ right before entering mode $i$ (with next mode $j$)}{

        Optimize $p_i$ with the loss $\mathcal{L}_e$ ($\mathcal{L}_e^+$) in \eqref{eq:loss-e} (\eqref{eq:loss-e-heur})

        $x(t)\gets h_k(x(t), u; p_k, p_i), \,\, k\gets i$
    }
    \uElse{
        $x(t+\Delta t)\gets x(t) + f_k(x(\tau), \pi_k(x(\tau), p_k), p_k) \Delta t$, \quad $t\gets t+\Delta t$
    }
    
    }
    
\caption{Hybrid System Control Algorithm}\label{alg:hybrid-control}
\end{algorithm}

\begin{figure}[!htbp]
\floatconts{fig:env}
{\caption{Simulation environment visualization and reward comparisons under varied training sizes: (a) car control on icy roads (25 maps) (b) pogobot navigation in 2D mazes (25 maps) (c) Bipedal walker control. The RL reward is averaged across 3 random seeds.}}
{\subfigure[Car tracking control]{  
\includegraphics[width=0.28\textwidth]{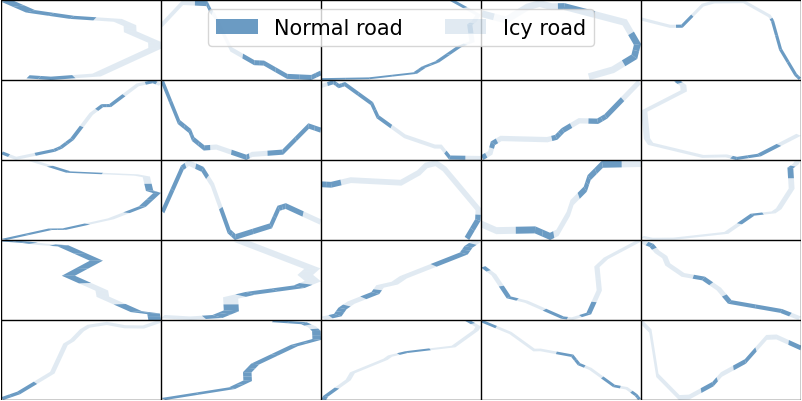} 
}%
\hfill
\subfigure[Pogobot navigation]{  
\includegraphics[width=0.28\textwidth]{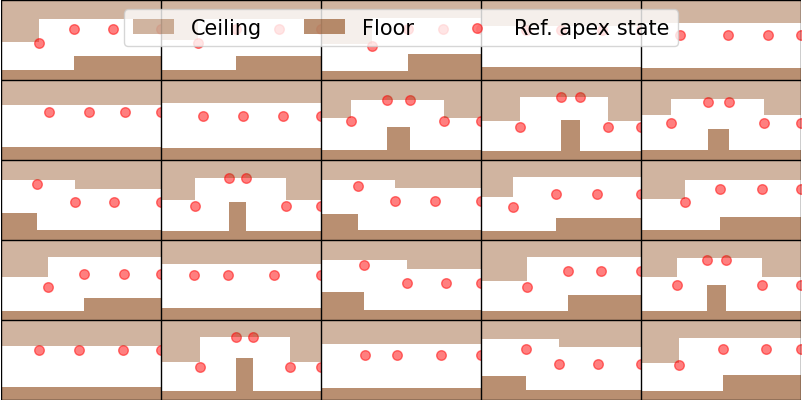} 
}%
\hfill
\subfigure[Bipedal gait tracking]{  
\includegraphics[width=0.28\textwidth]{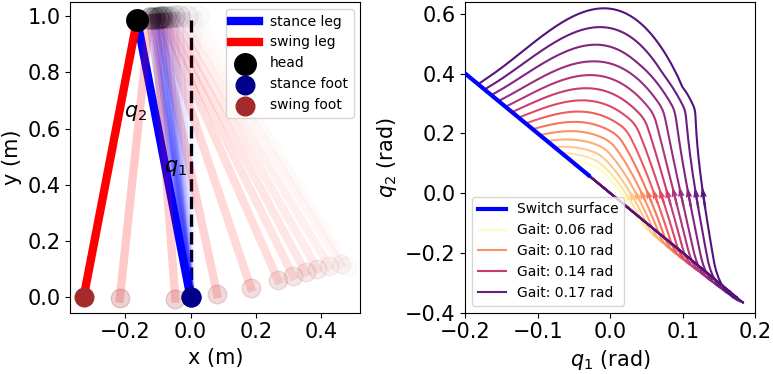} 
}%
\hfill
\subfigure[Car control reward]{  
\includegraphics[width=0.30\textwidth]{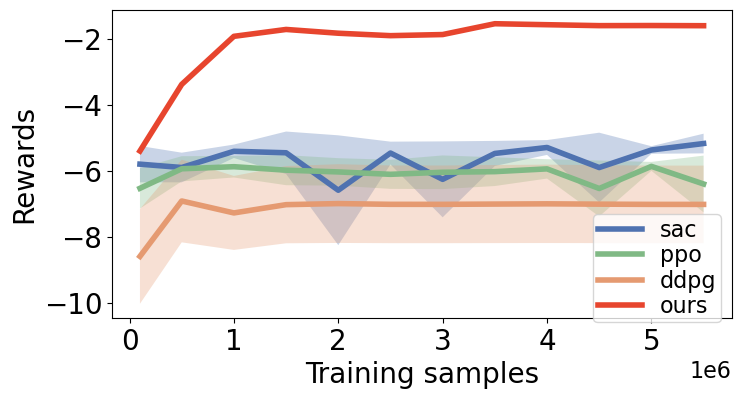}
}
\hfill
\subfigure[Pogobot reward]{  
\includegraphics[width=0.29\textwidth]{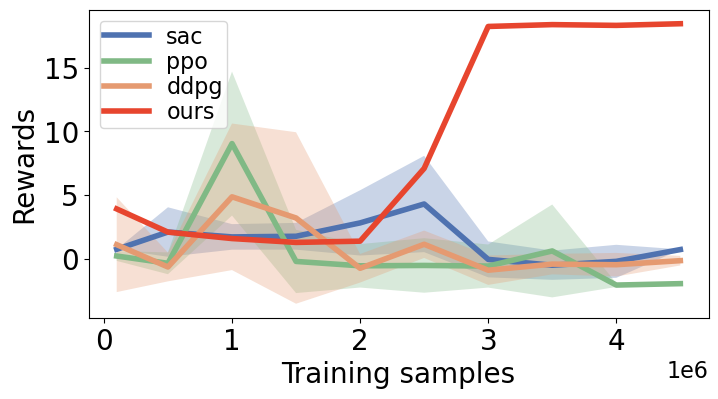}
}
\hfill
\subfigure[Bipedal walker reward]{  
\includegraphics[width=0.30\textwidth]{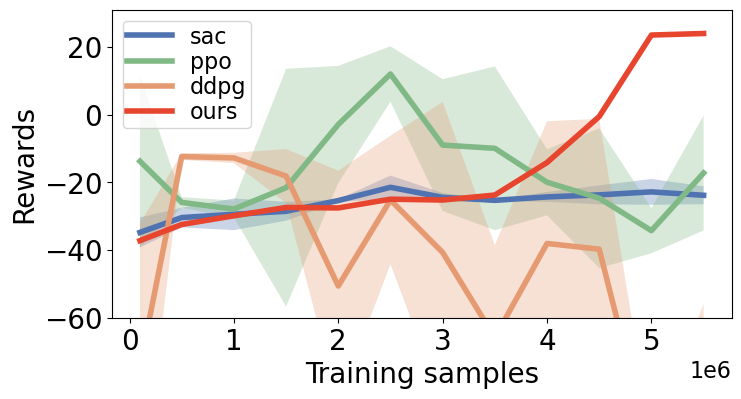}
}
}
\end{figure}

\section{Experiments}
\begin{wrapfigure}{r}{0.55\textwidth}
\floatconts{fig:car-roa-comparisons}
{\caption{RoA comparison with LQR.}}
{
\subfigure[RoA]{  
\includegraphics[width=0.15\textwidth]{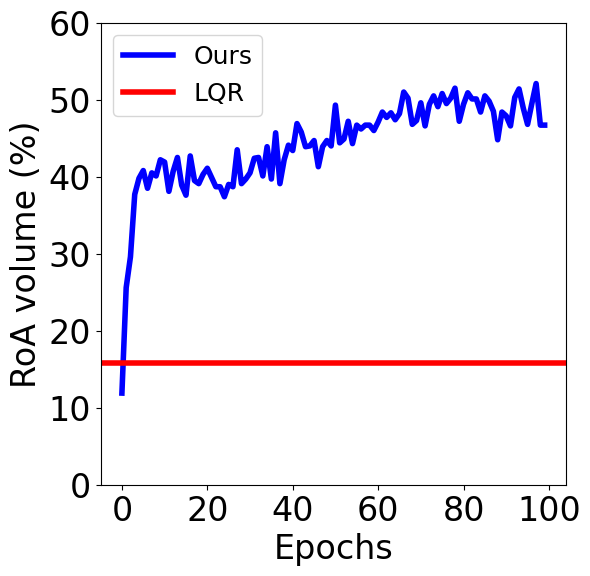} \hfill
}
\hfill
\subfigure[x-y plane]{  
\includegraphics[width=0.17\textwidth]{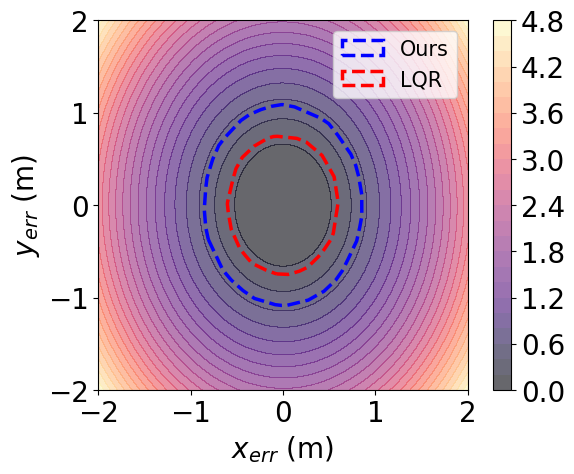} \hfill
}
\hfill
\subfigure[v-$\delta$ plane]{  
\includegraphics[width=0.17\textwidth]{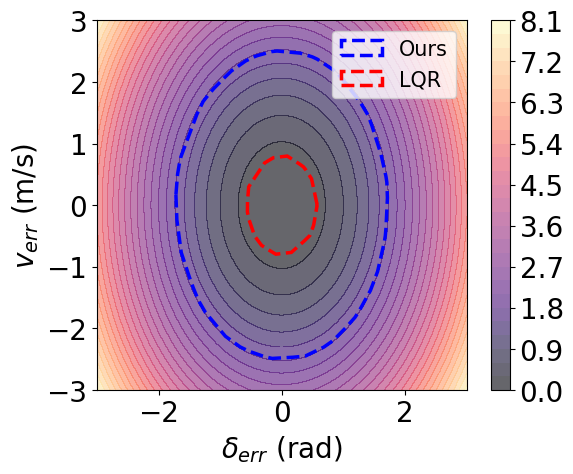} \hfill
}
}
\end{wrapfigure}
We conduct three challenging experiments shown in \figref{fig:env}. Our method achieves the best performance (success rate, RMSE, etc) than most baselines and is 10$\sim$50X faster than MPC. Our approach also results in shorter training time (0.5X of RL methods and 0.1X of HJB approaches).
\partitle{Baselines} For all cases we compare with: 
 model-base RL (MBPO~\citep{janner2019trust}), model-free RL(SAC~\citep{haarnoja2018soft}, PPO~\citep{schulman2017proximal} and DDPG~\citep{lillicrap2015continuous}) and model predictive control (MPC). Besides, for the car case, we compare with Linear Quadratic Regulator (LQR) and single CLF~\citep{chang2019neural}. For the bipedal we compare with quadratic program (QP) and Hamilton-Jacobian (HJB)~\citep{choi2022computation}.
We did not compare with HJB for (9-dim) car or (8-dim) pogobot as the state dimension is too high for HJB to handle. 
\partitle{Implementation details}
For CLF, the controller and the RoA estimator, we use 2-layer NNs with 256 units in each layer and ReLU~\citep{nair2010rectified} in intermediate layers. The controller uses TanH~\citep{lecun2012efficient} in the last layer to clip the output in a reasonable range. We implement our method in PyTorch~\citep{paszke2019pytorch}. The training takes 3$\sim$6 hours on an RTX2080 Ti GPU. More details are in \appref{appendix-bsl-impl}.

\partitle{Remarks on sample efficiency}
As shown in \figref{fig:env}, compared to RL under the same sample size, we achieve the highest rewards. This is because RL directly interacts with the hybrid systems, whereas ours learns to control the \postac{system } under each mode, which is easier. 


\begin{figure}[!htbp]
\floatconts{fig:car-comparisons}
{\caption{Quantitative ((a)$\sim$(d)) and qualitative ((e)$\sim$(g)) comparison for car control experiment. At T=2.5s, our approach learns to decelerate and turn left a bit to prepare for the incoming turn at icy road. At T=3.8s, our method is on the road, whereas other baselines diverge.}}
{
\subfigure[Lane deviation]{  
\includegraphics[width=0.23\textwidth]{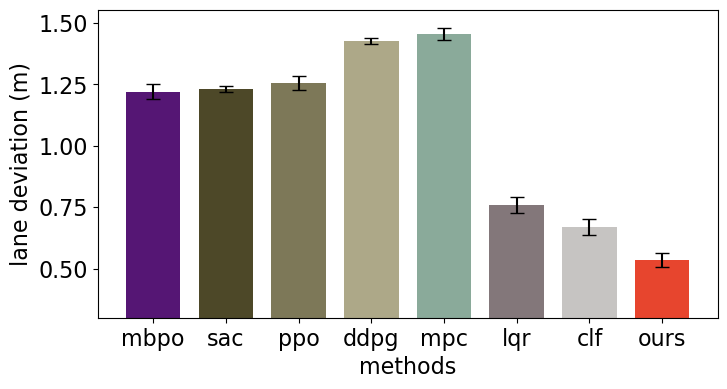}
}
\subfigure[Mean square error]{  
\includegraphics[width=0.23\textwidth]{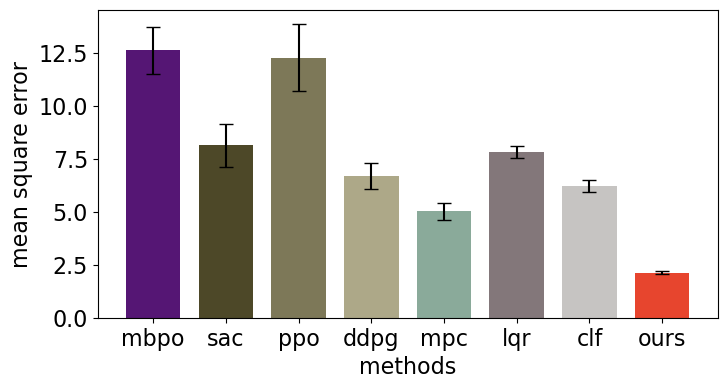} 
}
\subfigure[Distance to goal]{  
\includegraphics[width=0.23\textwidth]{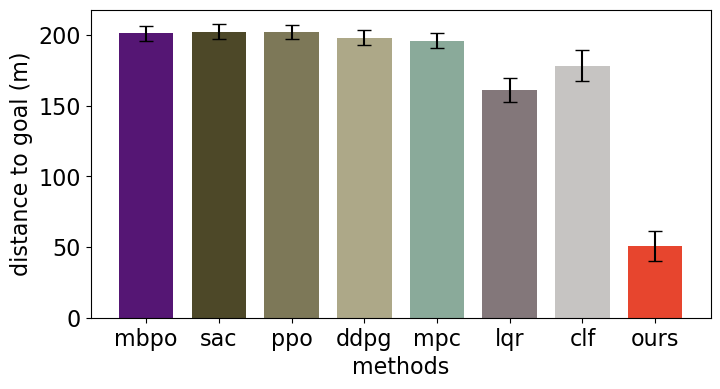}
}
\subfigure[Computation time]{  
\includegraphics[width=0.23\textwidth]{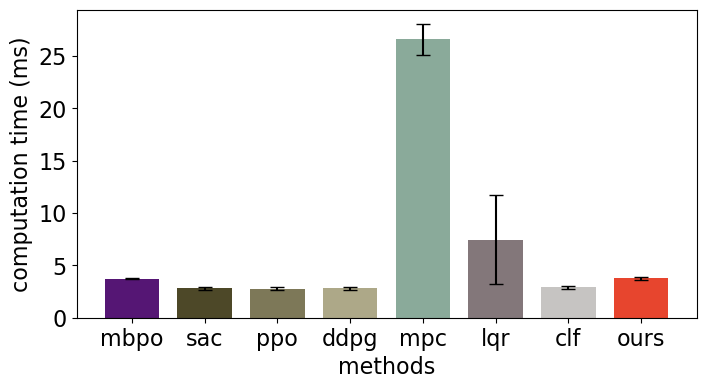}
}
\subfigure[T=2.5s]{  
\includegraphics[width=0.315\textwidth]{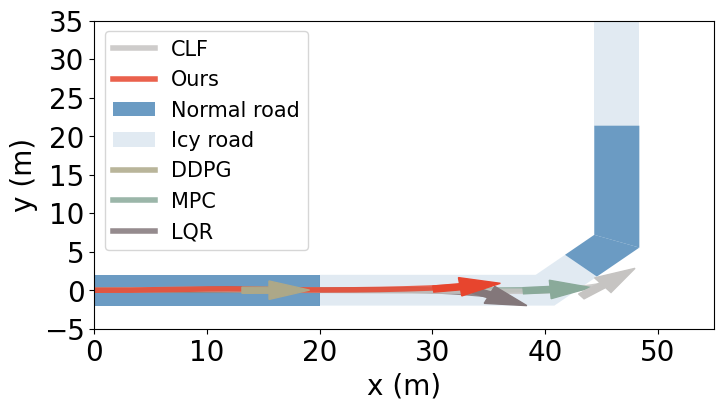}
}
\subfigure[T=3.8s]{  
\includegraphics[width=0.315\textwidth]{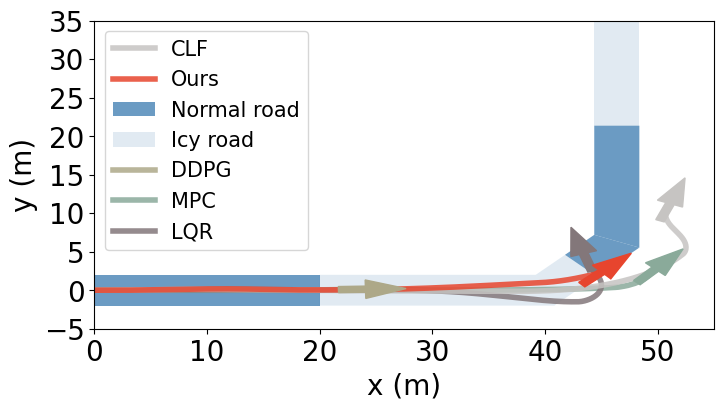} 
}
\subfigure[T=5.8s]{  
\includegraphics[width=0.315\textwidth]{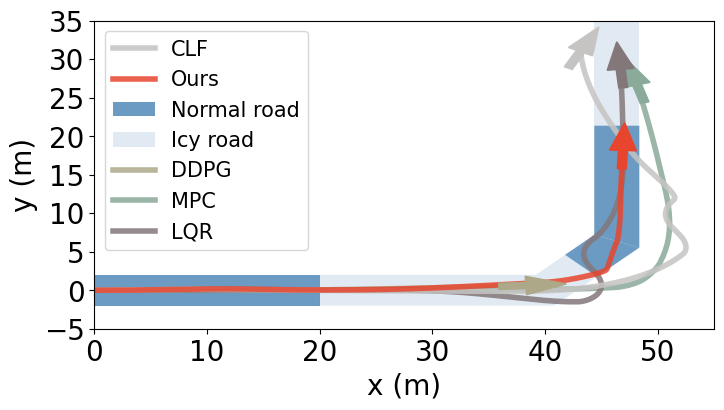}
}
}
\end{figure}

\subsection{Car control under different road conditions}
We consider the tracking control problem for the single-track model in~\citep{althoff2017commonroad} under varied frictions. The system is hybrid as varied frictions and tracking velocities lead to different dynamics. The challenge is that when the friction is low (e.g., on icy roads), the road cannot provide enough traction to keep the car on the lane. To stabilize the system, the controller needs to track proper configurations (speeds and waypoints). Our method ensures the entering state at the connection of segments is always in the RoA of the next segment. \partitle{Setups} We generate 25 maps with 10 randomly sampled segments. Each segment has friction $\in\{0.1,1\}$ and length $\in[7.5\text{m},37.5\text{m}]$. The angle between consecutive segments $\in[\frac{3\pi}{4},\pi]$. \partitle{Metrics} We measure average lane deviation, mean square error (MSE) to the reference trajectory, distance to the goal (before driving out of lane), and the run time per step.

As shown in \figref{fig:car-roa-comparisons}(a), the RoA of our approach surpasses the LQR method in just a few epochs and converges in 100 epochs, becoming nearly 2X larger than the LQR RoA. From the RoA visualization in \figref{fig:car-roa-comparisons}(b, c), we show that the gain is mainly from velocity error, tire angle error, longitudinal and lateral errors. This shows our method can enlarge the RoA to stabilize more states. The rest visualizations can be found in \appref{appendix-roa-viz}.

Next, we compare with other baselines. As shown in~\figref{fig:car-comparisons}, we achieve the lowest lane deviation and MSE, 67\%$\sim$75\% reduction in the distance to goal, and low run time on par of RL, which is less than 0.1X of the time used by MPC. Qualitatively, as shown in \figref{fig:car-comparisons} (here we omit MBPO, SAC and PPO because they cannot provide reasonable trajectories), our approach is the only method that can keep the car on the road. The reason is that our method learns to decelerate and turn left to prepare for the next icy road segment (\figref{fig:car-comparisons}(a)), so that the car can gain more traction on the icy road for a normal left turn (\figref{fig:car-comparisons}(b)) where other methods fail to keep the car on the road (\figref{fig:car-comparisons}(c)).


\begin{figure}[!htbp]
\floatconts{fig:pogo-comparisons}
{\caption{Quantitative ((a)$\sim$(d)) and qualitative ((e)$\sim$(h)) comparison for the pogobot. 
\postac{Our approach is the only one that can safely jump through the maze. DDPG and PPO methods start to jump to the left afterwards, and MPC results in collisions.}
}}
{
\subfigure[Velocity error]{  
\includegraphics[width=0.225\textwidth]{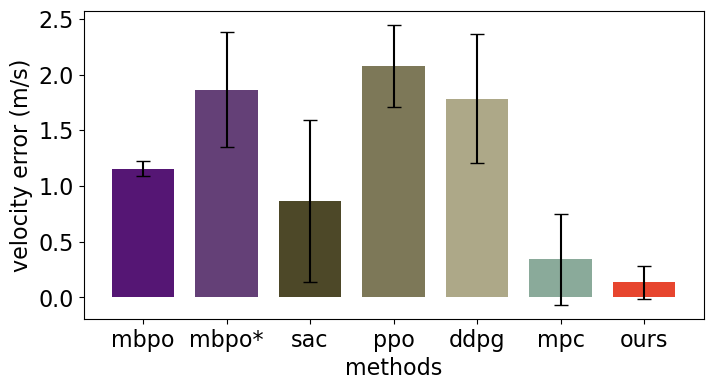}
}
\subfigure[Distance to goal]{  
\includegraphics[width=0.225\textwidth]{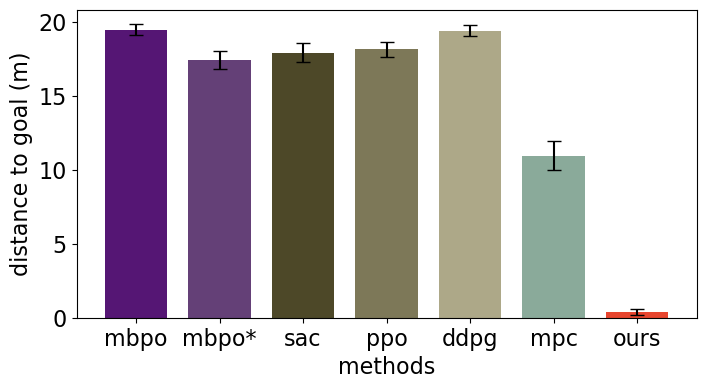}
}
\subfigure[Collision rate]{  
\includegraphics[width=0.225\textwidth]{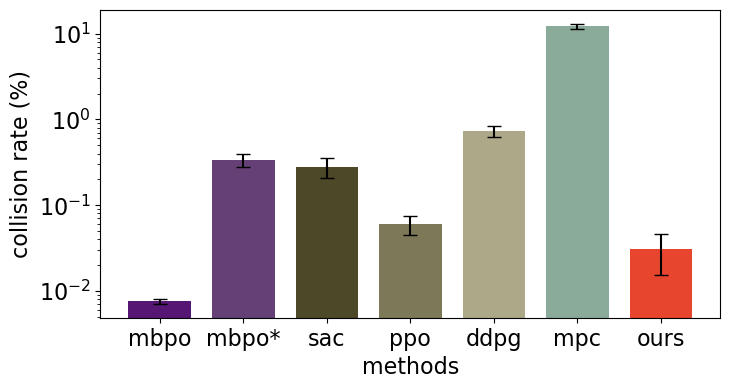} 
}
\subfigure[Runtime]{  
\includegraphics[width=0.225\textwidth]{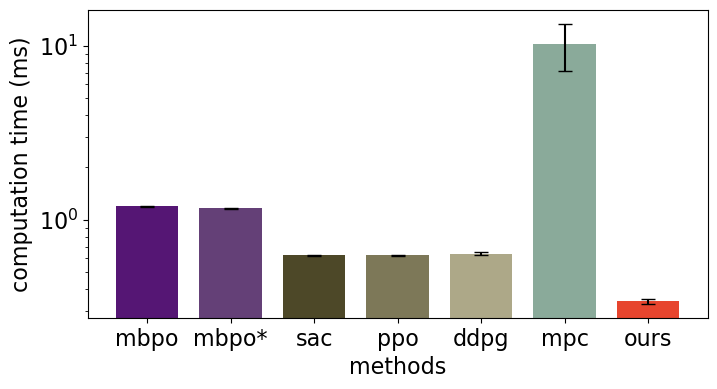}
}
\subfigure[RL method (DDPG)]{  
\includegraphics[width=0.23\textwidth]{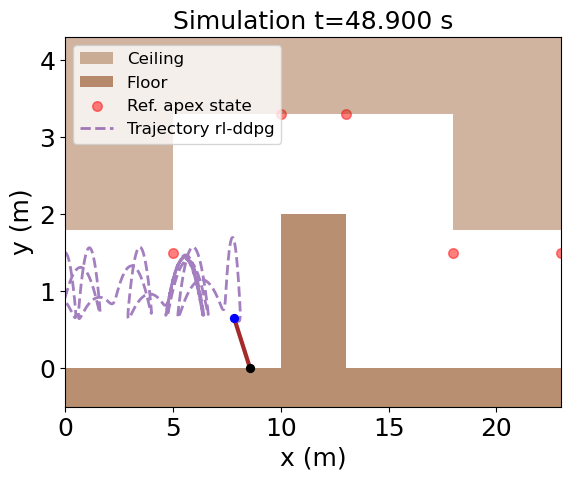}
}
\subfigure[RL method (PPO)]{  
\includegraphics[width=0.23\textwidth]{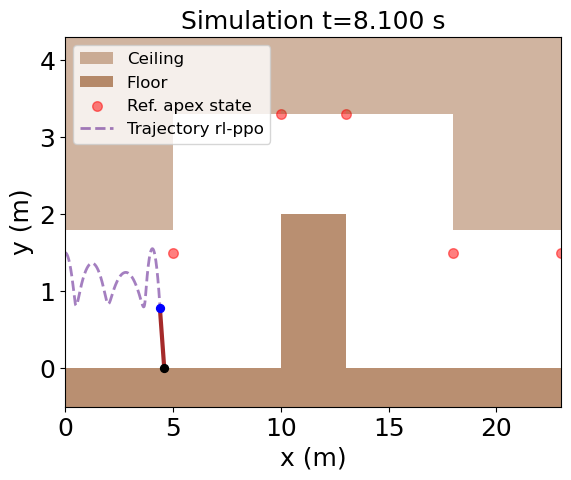}
}
\subfigure[MPC method]{  
\includegraphics[width=0.23\textwidth]{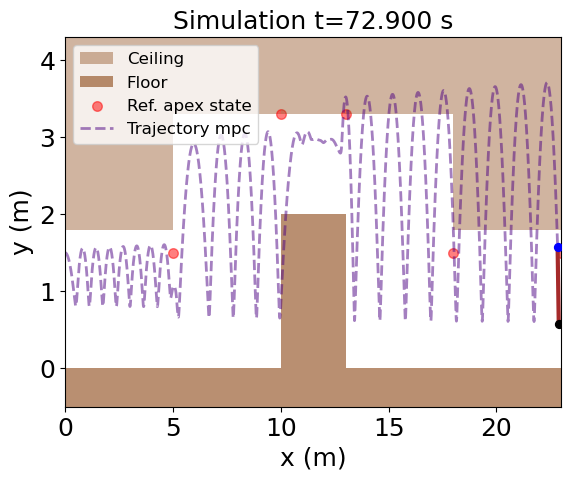} 
}
\subfigure[Our approach]{  
\includegraphics[width=0.23\textwidth]{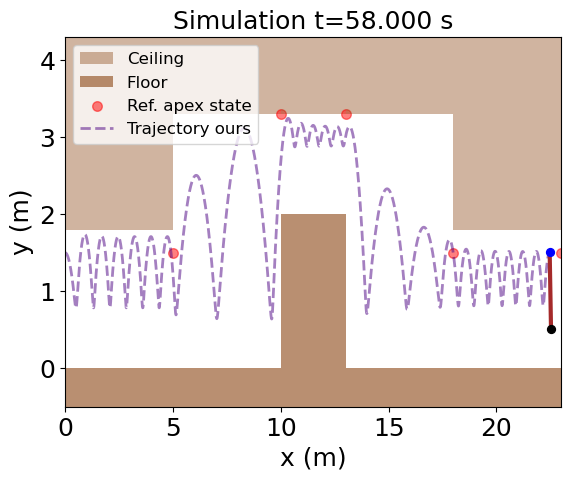}
}
}
\end{figure}

\subsection{Pogobot navigation}
We control a pogobot (with the model in~\citep{zamani2019feedback}) to jump through 2d mazes shown in \figref{fig:pogo-comparisons}. The pogobot alternates between flight and stance phases. 
The apex state is when the pogobot is at the top of a trajectory in the flight phase. Given the reference apex states, the goal is to jump through the maze safely. Our method \postac{first learns the unknown apex state dynamic then } plans configurations (reference apex state) to ensure apex states are within the RoA of the next reference apex state. \partitle{Setups} We generate 25 maps by randomly sampling 3$\sim$5 segments with a segment length $\in[3m,6m]$, a floor height $\in[-0.5m,2m]$, a ceiling height $\in[1.5m,3.5m]$, and a reference speed $\in[0.5m/s,1.5m/s]$. \partitle{Metrics} We measure the velocity error, the remaining distance to goal, the collision rate, and the computation runtime.

Since we use a learned dynamics in this case, we also compare with MBPO with pretrained dynamic model learnt from our approach (denoted as $\text{MBPO}^*$). As shown in \figref{fig:pogo-comparisons}, our approach achieves the lowest velocity tracking error, remaining distance to goal, low collision rate and computation time (MBPO achieves the lowest collision rate as it often jumps out of valid region before crashing - which explains its high distance-to-goal metric). Our approach uses just 1/70X of the computation time as needed for MPC. The simulation in \figref{fig:pogo-comparisons} shows that ours is the only approach to safely jump through the 2d maze, whereas DDPG and PPO methods start to jump to the left afterwards (here we omit the SAC result because it cannot plan for one cycle), and MPC method results in collisions. More details can be found in \apprefref{appendix-sim-env}{appendix-our-impl}. 

\begin{figure}[!htbp]
\floatconts{fig:bipedal-comparisons}
{\caption{Bipedal walker comparison under same ((a)$\sim$(d)) and different target gaits ((e)$\sim$(h)).\postac{Our method achieves the closest performance than the second-best method HJB on the same gait, and achieves much better performance on different gaits setup.}}}
{
\subfigure[RMSE]{  
\includegraphics[width=0.23\textwidth]{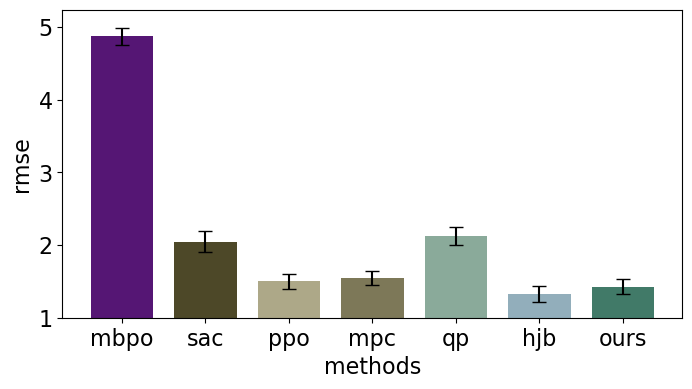}
}
\subfigure[Failure rate]{  
\includegraphics[width=0.23\textwidth]{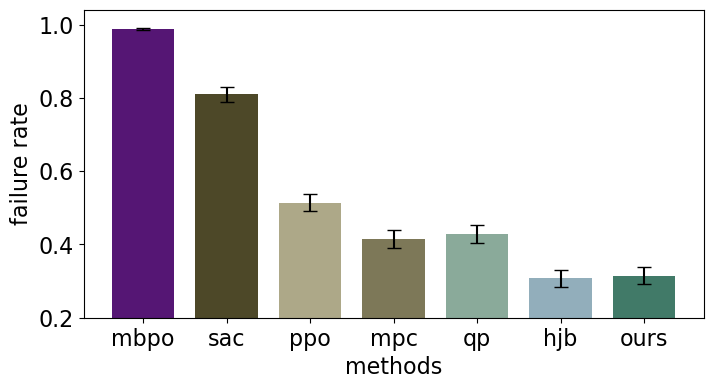}
}
\subfigure[Invalid rate]{  
\includegraphics[width=0.23\textwidth]{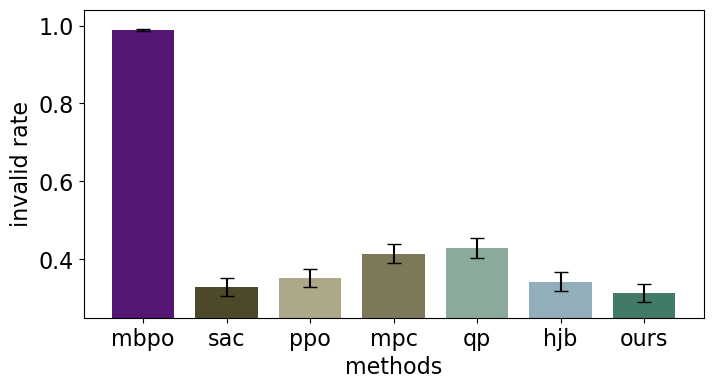}
}
\subfigure[Computation time]{  
\includegraphics[width=0.23\textwidth]{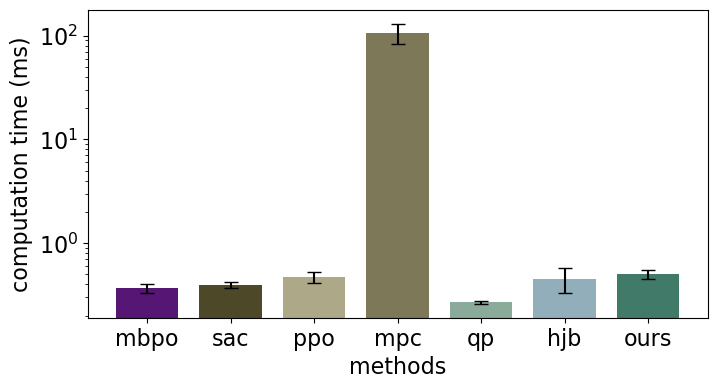}
}
\subfigure[RMSE]{  
\includegraphics[width=0.23\textwidth]{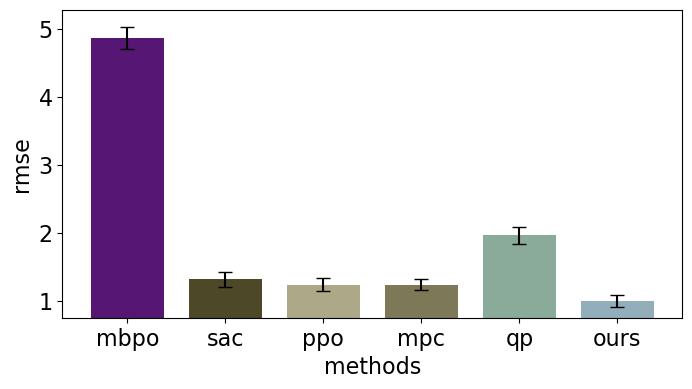}
}
\subfigure[Failure rate]{  
\includegraphics[width=0.23\textwidth]{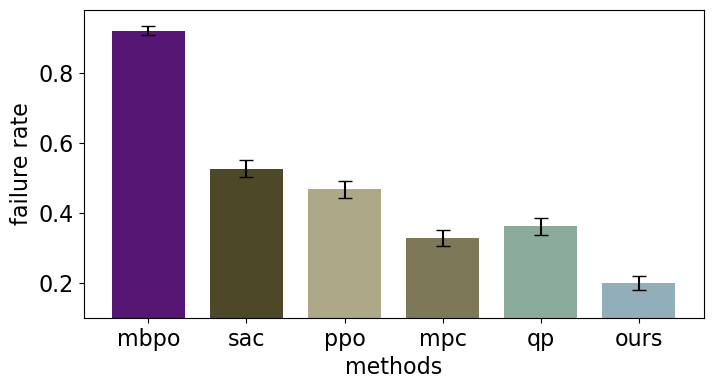}
}
\subfigure[Invalid rate]{  
\includegraphics[width=0.23\textwidth]{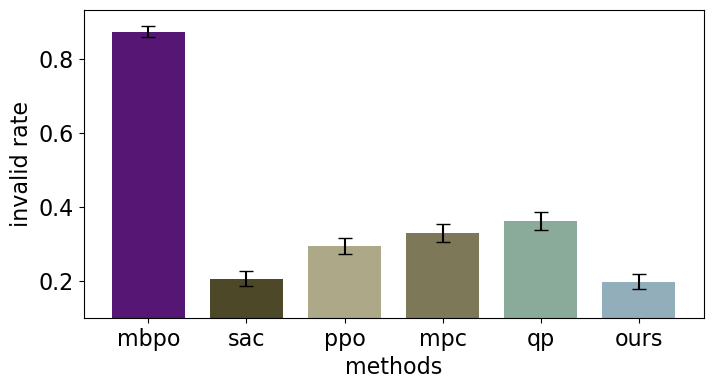}
}
\subfigure[Computation time]{  
\includegraphics[width=0.23\textwidth]{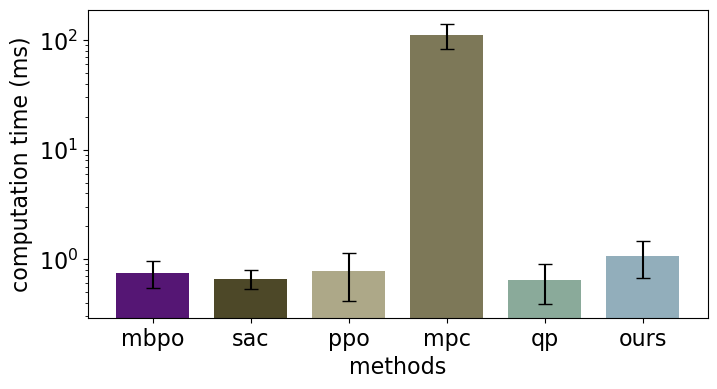}
}
}
\end{figure}

\subsection{Bipedal walker locomotion}

We control the bipedal robot~\citep{choi2022computation} to converge to a target gait. The configuration here is the leg angle $q_1$ of the next gait at the switch surface shown in \figref{fig:env}(c). It is hard to converge directly to the goal gait when the initial gait is far away. Thus, we use the loss in \eqref{eq:loss-e-heur} in planning to ensure the gait at each switching is closer to the target gait and our controller guarantees the state converges to the intermediate gait. \partitle{Setups} We calculate the gaits using Frost Library~\citep{hereid2017frost}, uniformly sample the initial states around each gait with angle $\in[0.04,0.18]$ and target gait~$\in[0.04,0.18]$. When compared to HJB, we set the target gait$=$0.13$\text{rad}$ since HJB only handles that gait. We compare other baselines with different target gaits. \partitle{Metrics} We measure RMSE towards reference gaits, failure rate for convergence, invalid trajectories rate and run time.

As shown in \figref{fig:bipedal-comparisons}(a)-(d), our approach achieves the closest RMSE, failure rate, and invalid trajectory rate to HJB approach. The third best method for failure rate is MPC, which takes a much longer computation time. MBPO has the worst performance probably due to the complicated dynamics which is hard to learn. Compared to HJB, our advantage is in the quick adaptation to other targeted gaits. Trained in less than 6 hours, our method can also learn to converge to other target gaits, whereas the learning time for the HJB method to converge to one target gait is 36 hours. We compare our method with other non-HJB approaches for converging to different gaits. As shown in \figref{fig:bipedal-comparisons}(e)-(h), we achieve the lowest RMSE, failure rate, invalid rate, and low running time.

\subsection{Limitations}
\label{sec:limitations}
Firstly, the RoA estimation is after the controller training, thus a refinement for the RoA estimator is needed if the controller is updated. One thought is to design a ``robust'' RoA estimator that can tolerate mild changes in controller parameters.
Besides, since the certificates are NNs with limited amount of neurons, we cannot guarantee the certificates being satisfied in the whole state space. Also we might bring in errors when numerically approximate the continuous dynamic, though we show in~\appref{appendix-abl-hyper} that the performance is consistent across varied $\Delta t$. In addition, we assume we can obtain full information for the system state and assume no disturbances or noises, which might not hold in real-world experiments. We aim to solve these in future works.

\section{Conclusion}
We propose a learning-based hierarchical approach to stabilize nonlinear hybrid systems. The learned low-level controller can stabilize the system states within each mode. Upon switching, the high-level planner finds the optimal configuration that guarantees the next entering state is within the RoA of the next mode. Experimental results show that our approach achieves the best overall performance on multiple hybrid systems compared to other approaches. For future works, we aim to tackle some of the limitations discussed in \secref{sec:limitations}, and extend this framework to further control hybrid systems with perception modules and the presence of estimation error and disturbances.

\section*{Acknowledgement}
The Defense Science and Technology Agency in Singapore and the C3.ai Digital Transformation Institute provided funds to assist the authors with their research. However, this article solely reflects the opinions and conclusions of its authors and not DSTA Singapore, the Singapore Government, or C3.ai Digital Transformation Institute.



\bibliography{ref}

\appendix
\clearpage
\renewcommand\thesection{\Alph{section}}
\setcounter{page}{1}
\setcounter{section}{0}
\setcounter{figure}{0}

{\LARGE\textbf{Supplementary Materials}}

\section{Proof for \theoref{theo-2}}
\label{appendix-prop2}
From $V_i(x_i, p_i)\leq c_i(p_i)$ we know that the entering state $x_i$ is within the maximum $\epsilon$-stable level set of equilibrium point $x^*$, hence the entering state $x_i$ is within the $\epsilon$-RoA of mode $i$. Next, we show that the next entering state $x_j = h_i(\bar{x}_i,u;p_i,p_j)$ is also within the $\epsilon$-RoA of mode $j$.

Since $x_i$ is within $\epsilon$-RoA of mode $i$,  we know  $||\bar{x}_i-x^*||\leq \epsilon$. Then from $\alpha_j||x-x^*||\leq V_j(x, p_j)\leq \beta_j||x-x^*||$ we have
\begin{equation}
\begin{aligned}
    V_j(x_j, p_j) & = V_j(h_i(\bar{x}_i,u;p_i,p_j), p_j)  \\
    & \text{(Definitions of jump maps and entering/exiting states)}\\
    & \leq \beta_j ||h_i(\bar{x}_i,u;p_i,p_j)|| \\ 
    & \text{(Lyapunov bounding condition)}\\
    & \leq \beta_j ||h_i(x^*,u;p_i,p_j)|| +\beta_j K_i ||\bar{x}_i-x^*|| \\ 
    & \text{(Local Lipschitz condition for $h_i$ at $x^*$)}\\
    & \leq \beta_j||h_i(x^*,u;p_i,p_j)|| + \beta_j K_i \epsilon \\ 
    & \text{(Definition of $\epsilon$-RoA for mode $i$)}\\
    & \leq \frac{\beta_j}{\alpha_j} V_j(h(x_i^*, u; p_i, p_j), p_j) + \beta_j K_i \epsilon \quad \\
    & \text{(Lyapunov function bounding condition)}
    \\
    & \leq \frac{\beta_j}{\alpha_j} \Big(\frac{\alpha_j}{\beta_j} c_j(p_j)-\alpha_jK_i\epsilon\Big) + \beta_j K_i \epsilon \\ 
    & \text{(The condition in \eqref{eq:v-condition})} \\ 
    & \leq c_j(p_j)
\end{aligned}
\end{equation}
therefore we derive that $V_j(x_j,p_j)\leq c_j(p_j)$, which means $x_j$ is within the $\epsilon$-RoA for mode $j$. So the whole hybrid system is $\epsilon$-stable according to Def.~\ref{def:e-stable}.

\section{Proof for \theoref{theo-3}}
\label{appendix-prop3}
We consider the lower bound of $||p_i-p_k||$ for every jump. We know that the Lyapunov value at the entering state of mode $k$ (denote the switching $i\to k$) is:
\begin{equation}
\begin{aligned}
& V_k(h_i(x_i,u;p_i,p_k), p_k) \\
& = V_k(h^+(x_i, p_i)+p_i-p_k, u; p_k) \\ 
& \text{(definition of the special system)} \\
& \leq \beta_k ||h^+(x_i, p_i)+p_i-p_k|| \\ 
& \text{(Lyapunov bounding condition)}\\
& \leq \beta_k ||h^+(x_i, p_i)||+\beta_k||p_i-p_k|| \\ 
& \text{(Triangle inequality)}\\
& \leq \beta_k ||h^+(x^*, p_i)||+K_m\beta_k||x_i-x^*||+\beta_k||p_i-p_k|| \\ 
& \text{(Local Lipschitz condition)} \\
& = K_m\beta_k||x_i-x^*||+\beta_k||p_i-p_k|| \\ 
& \text{(Since $h^+(x^*, p_i)=x^*$)}\\
& \leq \beta_k K_m \epsilon + \beta_k ||p_i-p_k|| \\ 
& \text{(Definition of $\epsilon$-RoA)} \\
 \end{aligned}
\label{eq:appd-dist}
\end{equation}

If the optimization is feasible and the optimal $p$ exists, then from the assumption we know that $V_k(h^+(x_i,p_i)+p_i-p_k, p_k)$ for the optimal $p$ must be no larger than $c_k(p_k)$ (zero the first loss in \eqref{eq:loss-e-heur}). We are going to show that $V_k(h^+(x_i,p_i)+p_i-p_k, p_k)$ must strictly equal to $c_k(p_k)$. If not, based on the continuity of the $V_k$, there must exist a $\tilde{p}_k$ around $p_k$ that can also zero the first loss term in \eqref{eq:loss-e-heur}, and make $||\tilde{p}_k-p_j||\leq ||p_k-p_j||$ which brings contradiction. Thus we have $V_k(h^+(x_i, p_i)+p_i-p_k, p_k)=c_k(p_k)$, hence based on \eqref{eq:appd-dist}, we have:
\begin{equation}
    \begin{aligned}
    ||p_i-p_k||\geq \frac{c_k(p_k)}{\beta_k} - K_m \epsilon
    \end{aligned}
\label{eq:appd-dist2}
\end{equation}
For each jump, the step length is lower bounded as shown in \eqref{eq:appd-dist2}. Thus we have the number of jumps is:
\begin{equation}
    \begin{aligned}
    N\leq \left\lceil\frac{||p_j-p_i||}{\min\limits_{m}\frac{c_m}{\beta_m}-K_{m} \epsilon}\right\rceil 
    \end{aligned}
\end{equation}

\section{Details for the simulation environments}
\label{appendix-sim-env}
\subsection{Car tracking control}

The goal here is to make sure the car can drive on the road under different road conditions. Given a reference state $(x, y, v, \psi)^T$ for a Dubins car model, the state of the car model is $(x_e, y_e, \delta, v_e, \psi_e, \dot{\psi}_e, \beta)^T$, where $x_e, y_e$ represent the Cartesian error, $\delta$ denotes the steering angle, $v_e$ denotes the velocity error, $\psi_e$ and $\dot{\psi}_e$ are the heading angle error and angular velocity error, and $\beta$ is the slip angle. The dynamics are given by $\dot{x}=f(x)+g(x)u$, with 
\begin{equation}
    f(x)=\begin{pmatrix}
    v\cos(\psi_e) - v_{ref} + \omega_{ref} y_e \\
    v\sin(\psi_e)-\omega_{ref} x_e\\
    0\\
    0\\
    \dot{\psi}_e\\
    C_1(\dot{\psi}_e+\omega_{ref}) + C_2 \beta + C_3 \delta \\ 
    C_4 (\dot{\psi}_e-\omega_{ref}) + C_5\beta + C_6 \delta \\
    \end{pmatrix}
\end{equation}
with
\begin{equation}
    \begin{cases}
    C_1=-\frac{\mu m}{v I_x (l_r+l_f)} (l_f^2 C_{Sf} g l_r + l_r ^2 C_{Sr} g l_f) \\
    C_2=\frac{\mu m}{I_x(l_r+l_f)} (l_r C_{Sr} g l_f - l_f C_{Sf} g l_r) \\
    C_3=\frac{\mu m}{I_x(l_r+l_f)} (l_fC_{Sf}gl_r)\\
    C_4=\frac{\mu}{v^2(l_r+l_f)}(C_{Sr}gl_fl_r - C_{Sf} g l_r l_f) - 1 \\
    C_5=-\frac{\mu}{v(l_r+l_f)}(C_{Sr}gl_f + C_{Sf}gl_r)\\
    C_6=\frac{\mu}{v(l_r+l_f)}(C_{Sf} g l_r)\\
    \end{cases}
\end{equation}
and 
\begin{equation}
    g(x)=\begin{pmatrix}
    0 & 0 \\
    0 & 0 \\
    1 & 0 \\
    0 & 1 \\
    0 & 0
    \end{pmatrix}
\end{equation}
where $u$ is the acceleration and the steering angle output, $I_x, l_r, l_f, C_{Sf}, C_{Sr}, g$ are constant factors, and $\mu$ is the road friction factor. More details can be found in~\citep{althoff2017commonroad}.

The road consists of multiple segments with different road conditions (different $\mu$). Each segment belongs to a system mode with the configuration of reference waypoint $(X^E, Y^E)$, reference velocity $v^E$ and the friction factor $\mu$. Different combinations of friction factor and the velocity will give different traction force for the vehicle. At junctions of the two segments, the mode switching causes the system state jump because of the change of the reference waypoint.

\subsection{Pogobot navigation} The state of the pogobot is $s=(x, \dot{x}, y, \dot{y})^T$, where $x,y$ are the 2d coordinate of the pogobot head, and the $\dot{x}, \dot{y}$ are the corresponding velocities. The movement of a pogobot involves two phases. In the flight phase, the pogobot follows a ballistic dynamics $\dot{s}=f(s)$ where:
\begin{equation}
    f(s)= \begin{pmatrix}\dot{x} \\ 0 \\ \dot{y} \\ -g  \end{pmatrix}
\end{equation}
here $g$ is the gravity and the stance foot is determined by the pogobot pose $\theta$ (which can be controlled instantly, since we assume a mass-less leg). In the stance phase, together with the stance foot position $(x_f, y_f)^T$, the dynamics becomes
\begin{equation}
    \dot{s}=\begin{bmatrix}  
    \dot{x}\\
    \frac{x-x_f}{L} \left(k\left(L-l_0\right) + F\right)\\
    \dot{y}\\
    \frac{y-y_f}{L} \left(k\left(L-l_0\right)+F\right)-g
    \end{bmatrix}
\end{equation}
where $L=\sqrt{(x-x_f)^2+(y-y_f)^2}$ denotes the length of the current pogobot, $l_0$ denotes the original length of the pogobot, $k$ denotes the spring constant factor, and $F$ and $\theta$ are the control inputs (the stance force and the swing leg angle). Here we consider the apex-to-apex control strategy. We first collect the simulation data to learn an apex-to-apex dynamic estimator. Then we use this dynamic estimator to train our Lyapunov function and controllers (as well as the RL-based methods).

\subsection{Bipedal walker locomotion}
The state of the Bipedal walker is $s=(q, \dot{q})^T$ where $q=(q_1, q_2)^T$, and $q_1$ is the angle between the normal vector of the ground and the stance leg, and the $q_2$ is the angle between the stance leg and the swing leg. $\dot{q}_1$ and $\dot{q}_2$ are the corresponding angular velocities. Within each mode, the continuous dynamics of the system follows the manipulator equation:
\begin{equation}
    \dot{s}=\begin{bmatrix} \dot{q} \\ D^{-1}_s(q)[-C_s(q,\dot{q})-G_s(q)+B_s(q)u] \end{bmatrix}
\end{equation}
where $D_s, C_s, G_s, B_s$ are functions of $q$ (and $\dot{q}$), $u$ is the control input (torque in this case), and a state jump will occur when $q_2 + 2 q_1 = 0$. Here $B=(1, 0)^T$. We denote the leg mass $m$, the original leg length $l$ with center of mass (CoM) location $l_c$, the acceleration due to gravity $g_0$, and the leg inertial about leg CoM as $I$. Then the $D_s(q)$ can be written as:
\begin{equation}
    \begin{cases}
        (D_s(q))_{1,1}=(l-l_c)^2 m + I\\
        (D_s(q))_{1,2}= ml(l-l_c)\cos(q_2) - (l-l_c)^2 m - I\\
        (D_s(q))_{2,2}=-2ml (l-l_c) \cos(q_2) +\\
        \quad \quad \quad \quad\quad \quad \quad \quad\quad \quad (2(l_c^2 + l^2)-2l_c l) m + 2 I
    \end{cases}
\end{equation}
and the nonzero entries in $C_s$ are:
\begin{equation}
    \begin{cases}
      (C_s(q,\dot{q}))_{1,2}=-ml\sin(q_2) (l-l_c) \dot{q}_1 \\
      (C_s(q_2,\dot{q}_1))_{2,1}=-ml\sin(q_2) (l-l_c) (\dot{q}_2 - \dot{q}_1) \\
      (C_s(q_2,\dot{q}_1))_{2,2}=-ml\sin(q_2) (l-l_c) \dot{q}_2 \\
    \end{cases}
\end{equation}
and the nonzero entries in $G_s$ are:
\begin{equation}
    \begin{cases}
        (G_s(q_1,q_2))_1 = m g_0 \sin(q_2-q_1) (l-l_c) \\
        (G_s(q_1, q_2))_2 = m g_0 ((l_c-l) \sin (q2-q1)-\sin(q_1)(l_c+l))
    \end{cases}
\end{equation}

We recommend readers to~\citep{choi2022computation} for more details.

\section{Implementation of our approach}
\label{appendix-our-impl}

As mentioned in Algorithm. 1, we first learn the control Lyapunov function (CLF) and the NN controller, then we estimate the RoA, and finally we conduct online optimization for the planner in the deployment phase. 

For the CLF and controller learning phase (take the car tracking experiment as an example), we uniformly sample 1000 states from an initial set, then at every epoch we sample trajectories for 100 time steps from the corresponding environment simulator given the NN controller, We use the trajectories to train our CLF and NN controller. The CLF and the NN controller are 2-hidden-layer NNs with 256 hidden units in each layer and ReLU activation in the intermediate layers. The last layer of the controller uses TanH activation function to clip the output signal in a reasonable range. We train the CLF and NN controller via the loss in \eqref{eq:loss}. We use RMSprop gradient descent method for the optimization with the learning rate of $10^{-4}$. We train CLF and controller for (at maximum) 1000 epochs, where inside each epoch the CLF and the controller will be updated for 500 steps. We stop the training when the validation loss is not dropping significantly.

For the RoA Estimation phase, we use the CLF and the NN controller trained in the previous step to generate $10^3\sim 10^4$ trajectories in 100 time steps. Using CLF, we are able to find the largest Lyapunov value $c_i^*$  for all the sampled initial states that having the exiting states within the $\epsilon$-ball of the equilibrium. We set the $\epsilon=10^{-2}$. Then we train the RoA estimator using the loss in \eqref{eq:loss-roa} for 50000 iterations, with RMSprop optimizer and learning rate of $10^{-4}$.

For the online planning (deployment) phase, we use the differentiable planner to plan for valid configurations, and use the controller to "follow" that configuration to maintain stability. For the differentiable planner, at every mode switching instant, we first randomly generate 1000 configuration hypothesis. Then we use RMSprop optimizer and conduct gradient descent with learning rate of 0.05 for 5 steps. Then we pick the updated configuration hypothesis with the lowest loss.

\subsection{Car tracking control}

Before entering the $i$-th segment, we optimize for the configuration $p_i$, which is the waypoint $W_i=(x^{ref}_i, y^{ref}_i)$ at the junction between the $i$-th segment and the $i+1$-th segment, and the reference velocity $v^{ref}_i$ to track on the $i$-th segment. And at the $i$-th segment, we use the environment reference waypoint $(x^E_{i+1}, x^E_{i+1})$ and the reference velocity $v^E_{i+1}$ for the next configuration. We make sure:
(1) the current entering state $x_i$ is within the RoA of the current system under the configuration of $p_i=(W_i,v^{ref}_i)$. (2) the next entering state $x_{i+1}$ is within the RoA of the system at segment $i+1$ with configuration $p_{i+1}$.

\subsection{Pogobot navigation}
Before entering the $i$-th segment, we optimize for the configuration $p_i$, which is the reference apex state height and reference apex state horizontal velocity at the next cycle (during the $i$-the segment). We can find the last apex state $\tilde{x}_i$ before exiting the $i$-the segment using the dynamics estimator, and make sure $\tilde{x}_i$ is within the RoA for the $i+1$-th segment under the reference apex state $X^E_{i+1}$ given from the environment.

\subsection{Bipedal walker locomotion}
In this case, due to the difficulty to synthesize a control Lyapunov function, for the low-level controller, we directly use the QP controller derived from~\citep{choi2022computation} and the corresponding Lyapuonv function is replaced by an RoA classifier, which outputs value $<$0.9 if the entering state is within the RoA, and outputs value $>$1.1 for the rest case. During the planning, we use the differentiable planner but with the loss \eqref{eq:loss-e-heur} in to find the optimal configuration (in this case, the reference gait).

\section{Implementation of baseline approaches}
\label{appendix-bsl-impl}
\partitle{Model-based reinforcement learning approaches} We compare with the Model-based Policy Optimization (MBPO) method~\citep{janner2019trust} implemented by an open-sourced library~\citep{pineda2021mbrl}. Based on the hyper-parameters provided in the library configuration files, we then searched for 3$\sim$ 10 different hyper-parameters (learning rates, policy update frequency, etc) and picked the one with the highest task-performance. We finalize the hyper-parameters and train the MBPO policy under 3 different random seeds for 24 hours for each experiment. 

\partitle{Model-free reinforcement learning approaches} We modify the RL implementation code from \url{https://github.com/RITCHIEHuang/DeepRL_Algorithms}, created the RL environments for each experiment, train each method with 3 random seeds for 24 hours each and take the average performance in the testing. For the car experiment, we use the reward as a combination of Root Mean Square Error (RMSE) penalty with the reference state and the valid rate of the trajectory segments. For the pogobot experiment, we use the RMSE, the collision rate with the ceiling/floor, and the distance to the goal as the rewards/penalties. For the Bipedal walker, we use the L2-distance from the current state to the reference state on the target gait as the penalty to guide the controller to converge to the target gait.

\partitle{Model predictive control approaches} We use the CasADi~\citep{andersson2019casadi} to implement non-linear optimization for each tasks. In each case, the system is simulated under some parameters (controls) and the cost function is computed to optimize the control input. For the car experiment, the cost function is the tracking error within the prediction horizon (T=20). For the pogobot experiment, the cost function is a penalty term with collisions and a tracking error term to the horizontal reference velocity. For the bipedal walker experiment, the cost function is the L2-norm between the leg angle $q_1$ after the switching and the target gait leg angle $q_1^{ref}$. \postac{Due to the difficulty/scalability to encode the RoA conditions (as they are the outputs from the neural networks) in the traditional solver, we did not use RoA condition constraints in the MPC optimization.}

\partitle{Linear quadratic control approaches} At each segment, we require the car to track the segment endpoint and the designed reference velocity on the current segment given the friction factors. We compute the error dynamics, and synthesize the controller by solving the Algebraic Ricatti Equation similar as in ~\citep{dawson2022safe}. 

\partitle{Control Lyapunov function approaches} Followed by \citep{chang2019neural}, we jointly train a single NN Lyapunov function for all the system modes with a NN controller (that can also take modes as inputs), with the same amount of training time used as for our approach. 

\partitle{CLF-QP approach} We directly use the QP controller derived from~\citep{choi2022computation} for the Bipedal walker comparison.

\partitle{Hamilton Jacobian based approach} We directly use the computed result (the value function) from~\citep{choi2022computation} for the comparison for the target gait with the leg angle $q_1=0.13$ rad.

\section{Ablation studies for our method in the car experiment}
\label{appendix-abl-hyper}
We first study the selection of hyperparameters in differentiable planning process. We tuned for different possible values for the $\eta$ and $\kappa$ in the differential planning process. As shown in Table~\ref{table:abl-eta} and Table~\ref{table:abl-kappa}, the performance is not sensitive to the hyperparameter selections. For all our experiments, we choose $\eta=0.9$ and $\kappa=10^{-2}$.
\begin{table}[!htbp]
\begin{center}
\begin{tabular}{|c c c c c|} 
 \hline
 $\eta$ & $\kappa$ & Lane deviation & RMSE & Distance to goal \\ 
 \hline
1.2 & $10^{-2}$ & 2.082 & 0.53018 & 0.117 \\
\hline
1.0 & $10^{-2}$ & 2.100 & 0.523 & 0.166 \\
\hline
0.9 & $10^{-2}$ & 2.087 & 0.514 & 0.117 \\
\hline
0.8 & $10^{-2}$ & 2.084 & 0.514 & 0.117 \\
\hline
0.5 & $10^{-2}$ & 2.121 & 0.538 & 0.167 \\
\hline
\end{tabular}
\caption{How different $\eta$ will affect the performance in car experiment.}
\label{table:abl-eta}
\end{center}
\end{table}
\begin{table}[!htbp]
\begin{center}
\begin{tabular}{|c c c c c|} 
 \hline
 $\eta$ & $\kappa$ & Lane deviation & RMSE & Distance to goal \\ 
 \hline
0.9 & 0.0 & 2.107 & 0.519 & 0.166 \\
\hline
0.9 & $10^{-3}$ & 2.116 & 0.516 & 0.166 \\
\hline
0.9 & $10^{-2}$ & 2.087 & 0.514 & 0.117 \\
\hline
0.9 & $10^{-1}$ & 2.147 & 0.545 & 0.217 \\
\hline
0.9 & $10^0$ & 2.106 & 0.521 & 0.117 \\
\hline
\end{tabular}
\caption{How different $\kappa$ will affect the performance in car experiment.}
\label{table:abl-kappa}
\end{center}
\end{table}

Our method uses Euler method to approximate the continuous dynamics, which might lead to estimation error. Now we study how different approximations for continuous ODE will affect the control performance. We choose different simulation time duration $\Delta t$ (from 10ms to 0.16ms) and conduct the testing for our pretrained controller on the car benchmark. As shown in Table~\ref{table:abl-dt}, the performance is consistent across varied $\Delta t$ and all the metrics converges as $\Delta t\to 0$. As the time duration $\Delta t\to 0$, the estimation error will also decrease and the control performance will gradually converge to the performance on the real continuous dynamics. \postac{A larger $\Delta t $($>$10ms) might result in great estimation error in forward Euler approximation, thus we do not perform those experiments. We aim to tackle the controller synthesis problem under imperfectly estimated system dynamics in future works.}

\begin{table}[!htbp]
\begin{center}
\begin{tabular}{|c c c c|} 
 \hline
 Simulation $\Delta t$ (ms) & Lane deviation & RMSE & Distance to goal \\ 
 \hline
10.0 & 2.114 & 0.533 & 0.166 \\
\hline
5.0 & 2.125 & 0.522 & 0.216 \\
\hline
2.5 & 2.118 & 0.528 & 0.216 \\
\hline
1.25 & 2.106 & 0.531 & 0.166 \\
\hline
0.625 & 2.104 & 0.531 & 0.166 \\
\hline
0.3125 & 2.105 & 0.531 & 0.166 \\
\hline
0.15625 & 2.104 & 0.531 & 0.166 \\
\hline
\end{tabular}
\caption{How different simulation $\Delta t$ will affect the performance in car experiment.}
\label{table:abl-dt}
\end{center}
\end{table}

\postac{\section{Comments about data distribution}
In all our experiments, we use uniform sampling techniques to sample data from the state space (and configuration space) to learn the Control Lyapunov function, the controller, and the RoA estimator. This might not be the most efficient way to sample the data, as in high dimensional space, the feasible (stable) trajectories normally reside in a small volume of reachable sets. Commonly-used method to tackle this issue is to use a counter-example guided approach~\citep{chang2019neural} or to use rare-case event estimation to reweight the sampling distribution~\citep{sinha2020neural}. Improving the sampling quality and efficiency is beyond the scope of this project and we hope to address those in future work.}

\section{Success rate for Bipedal walker locomotion under different initial conditions}

\begin{figure}[!htbp]
\centering
\includegraphics[width=0.5\textwidth]{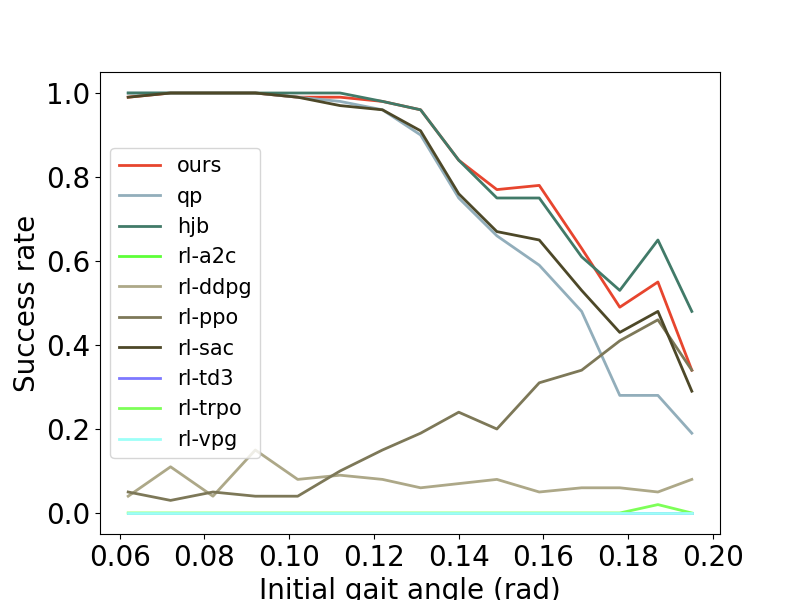}
\caption{Walker success rate comparison under different initial states}
\label{fig:supple-walker-init}
\end{figure}

For the bipedal walker experiment, we compare our approach with multiple RL-based approaches (A2C, DDPG, PPO, SAC, TD3, TRPO, VPG) and classic methods (QP, HJB) under different initial gait angles. As shown in~\figref{fig:supple-walker-init}, our approach can achieve similar-to-HJB performance, outperforming all the RL-baselines and the QP baseline. The largest improvement (comparing to RL methods) is from the "small initial angles". And our gain compared to QP-based methods is mostly from the "large initial angles", which might because the large deviation from the target gait angle makes the linearization more inaccurate, hence the QP-based method cannot achieve good performance.

\section{Visualization of learned RoA}
\label{appendix-roa-viz}
From \figref{fig:supple-car-roa-02} to \figref{fig:supple-walker-roa-04}, we show the visualization of the learned RoA in all three experiments under different configurations.

\begin{figure}[!htbp]
\floatconts{fig:supple-car-roa-02}
{\caption{Car experiment (friction $\mu=1.0$,  reference speed $v=5m/s$)}}
{
\subfigure[$x$-$y$ plane]{
\includegraphics[width=0.230\textwidth]{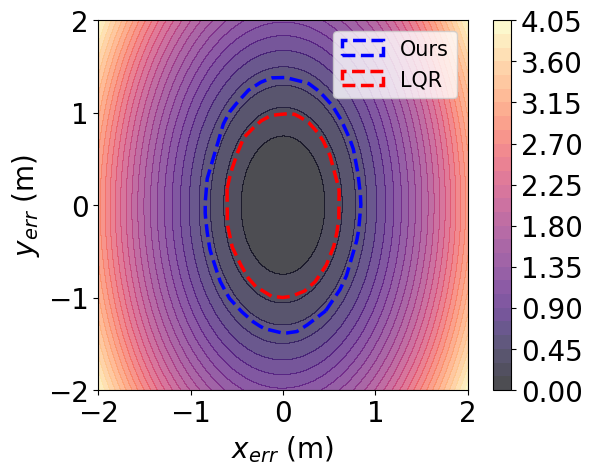}
}
\hfill
\subfigure[$v$-$\delta$ plane]{
\includegraphics[width=0.230\textwidth]{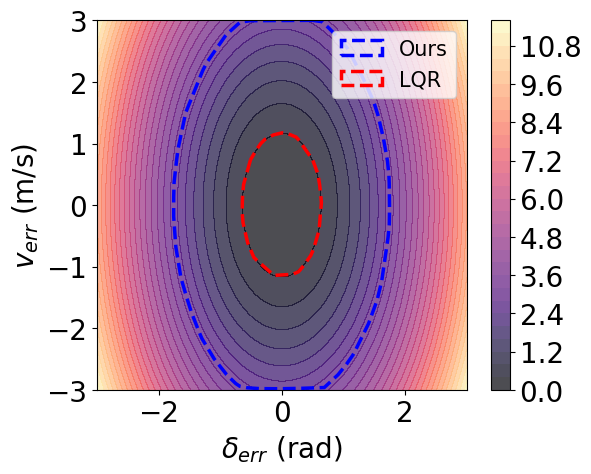}
}
\hfill
\subfigure[$\psi$-$\dot{\psi}$ plane]{
\includegraphics[width=0.230\textwidth]{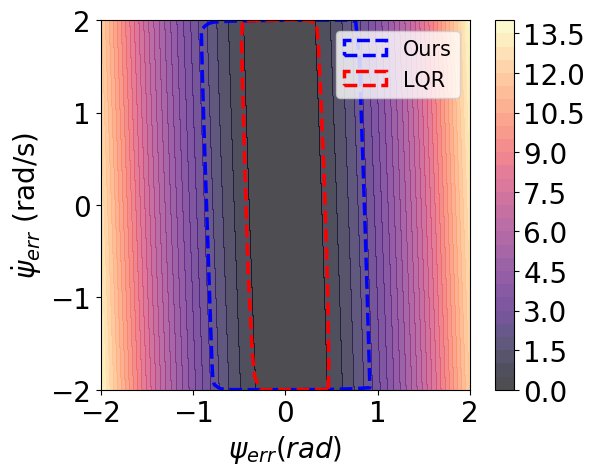}
}
\hfill
\subfigure[$\dot{\psi}$-$\beta$ plane]{
\includegraphics[width=0.230\textwidth]{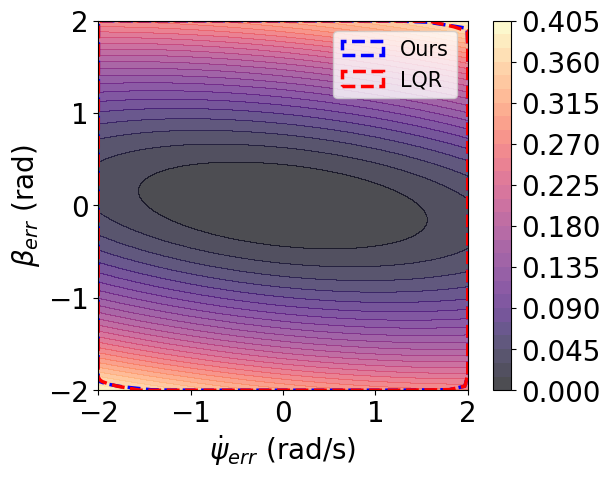}
}
\hfill
}
\end{figure}
\begin{figure}[!htbp]
\floatconts{fig:supple-car-roa-03}
{\caption{Car experiment (friction $\mu=0.1$,  reference speed $v=5m/s$)}}
{
\subfigure[$x$-$y$ plane]{
\includegraphics[width=0.230\textwidth]{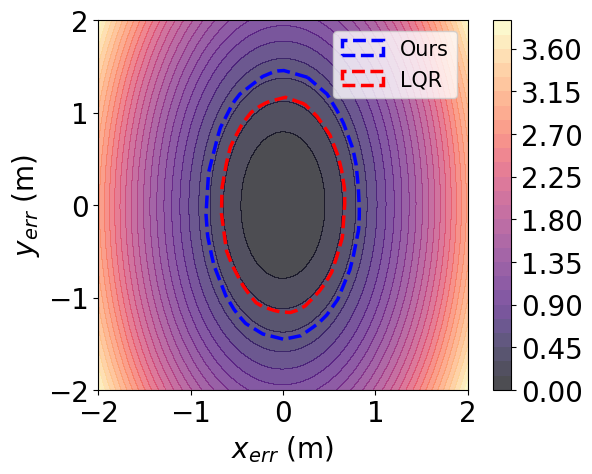}
}
\hfill
\subfigure[$v$-$\delta$ plane]{
\includegraphics[width=0.230\textwidth]{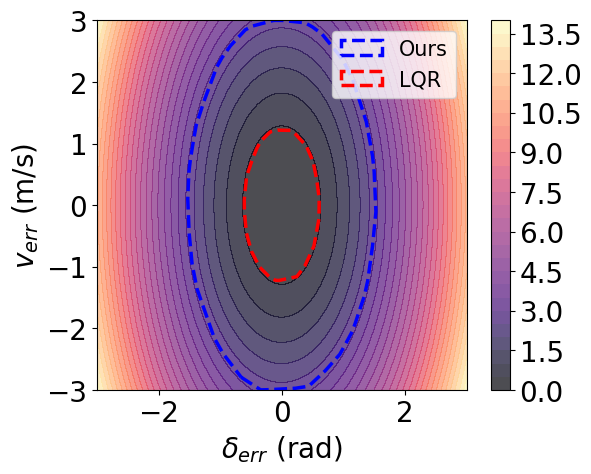}
}
\hfill
\subfigure[$\psi$-$\dot{\psi}$ plane]{
\includegraphics[width=0.230\textwidth]{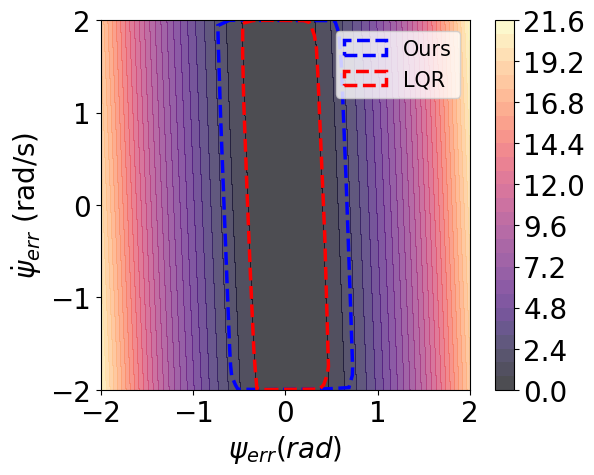}
}
\hfill
\subfigure[$\dot{\psi}$-$\beta$ plane]{
\includegraphics[width=0.230\textwidth]{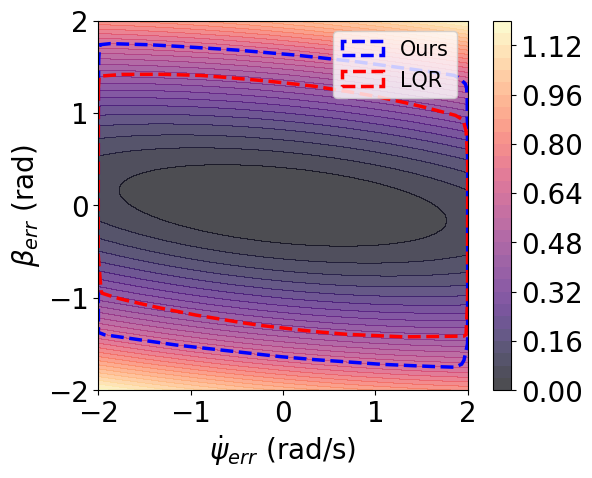}
}
\hfill
}
\end{figure}
\begin{figure}[!htbp]
\floatconts{fig:supple-car-roa-04}
{\caption{Car experiment (friction $\mu=1.0$,  reference speed $v=10m/s$)}}
{
\subfigure[$x$-$y$ plane]{
\includegraphics[width=0.230\textwidth]{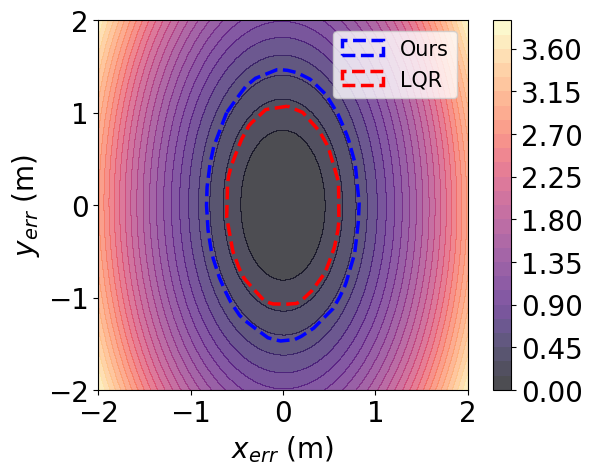}
}
\hfill
\subfigure[$v$-$\delta$ plane]{
\includegraphics[width=0.230\textwidth]{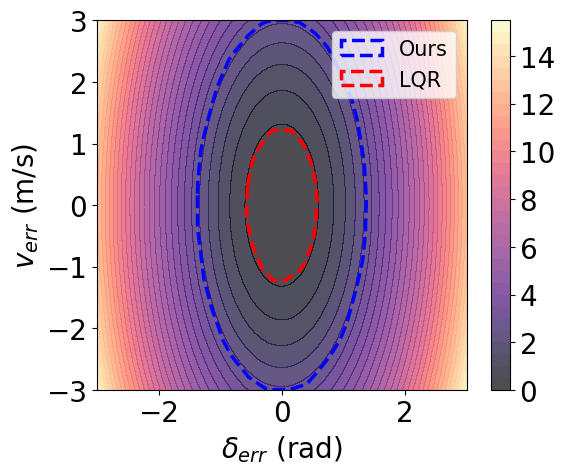}
}
\hfill
\subfigure[$\psi$-$\dot{\psi}$ plane]{
\includegraphics[width=0.230\textwidth]{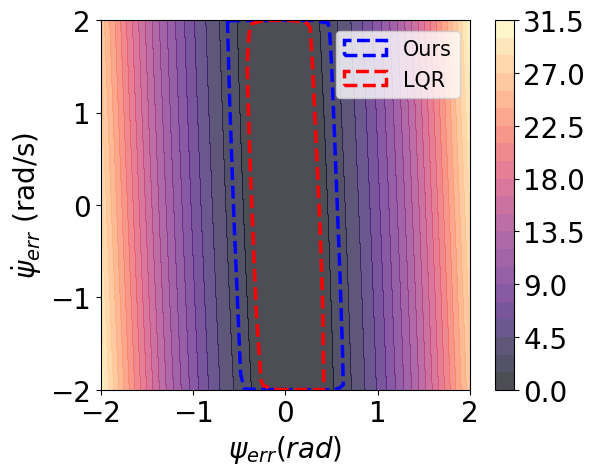}
}
\hfill
\subfigure[$\dot{\psi}$-$\beta$ plane]{
\includegraphics[width=0.230\textwidth]{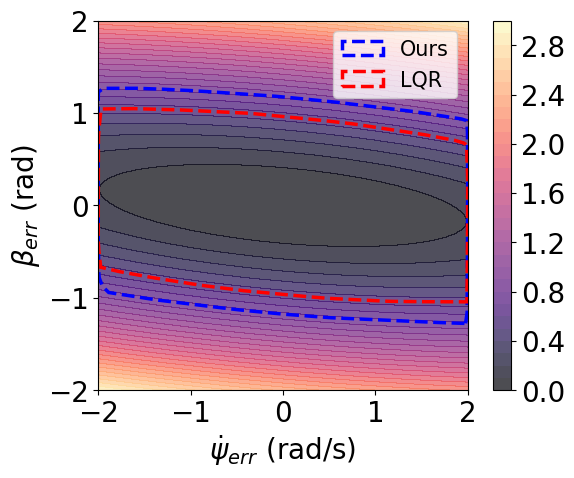}
}
\hfill
}
\end{figure}
\begin{figure}[!htbp]
\floatconts{fig:supple-car-roa-05}
{\caption{Car experiment (friction $\mu=0.1$,  reference speed $v=10m/s$)}}
{
\subfigure[$x$-$y$ plane]{
\includegraphics[width=0.230\textwidth]{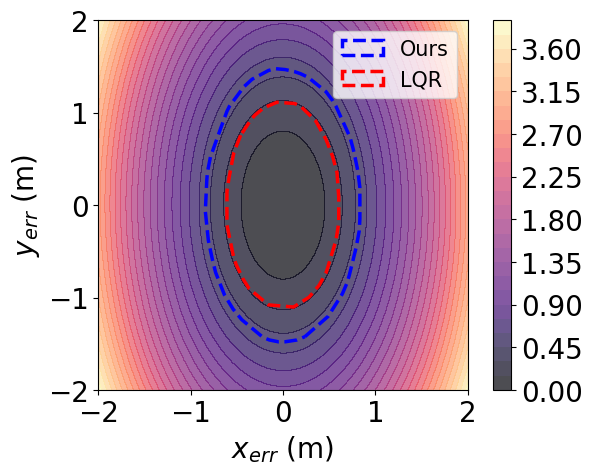}
}
\hfill
\subfigure[$v$-$\delta$ plane]{
\includegraphics[width=0.230\textwidth]{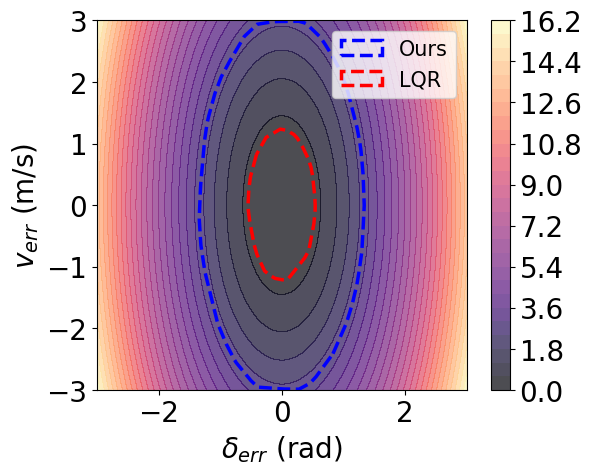}
}
\hfill
\subfigure[$\psi$-$\dot{\psi}$ plane]{
\includegraphics[width=0.230\textwidth]{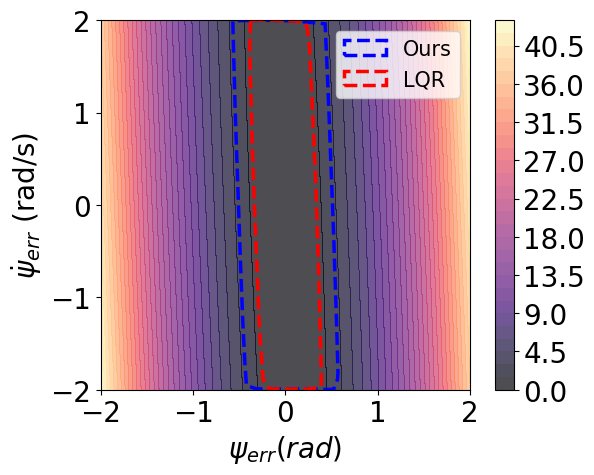}
}
\hfill
\subfigure[$\dot{\psi}$-$\beta$ plane]{
\includegraphics[width=0.230\textwidth]{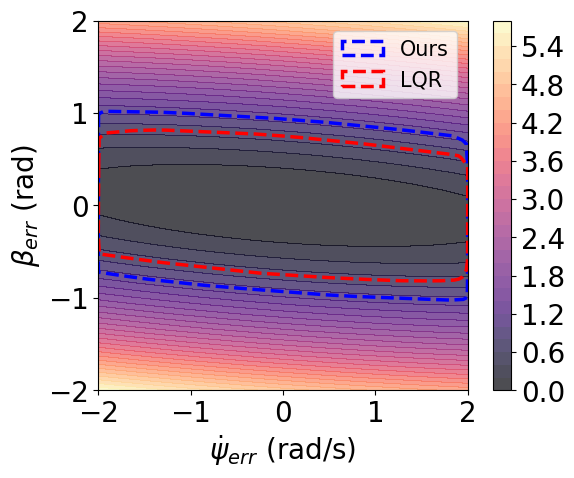}
}
\hfill
}
\end{figure}
\begin{figure}[!htbp]
\floatconts{fig:supple-car-roa-08}
{\caption{Car experiment (friction $\mu=1.0$,  reference speed $v=20m/s$)}}
{
\subfigure[$x$-$y$ plane]{
\includegraphics[width=0.230\textwidth]{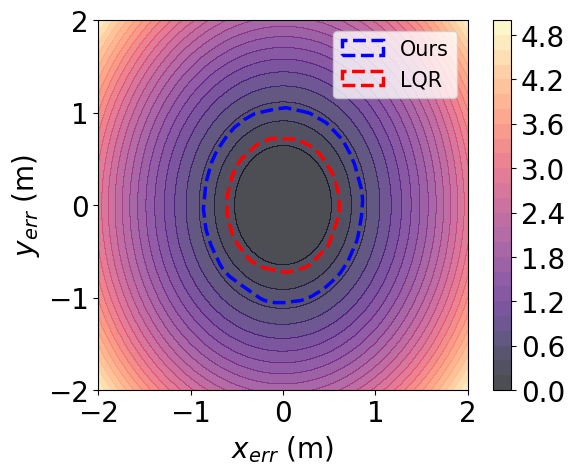}
}
\hfill
\subfigure[$v$-$\delta$ plane]{
\includegraphics[width=0.230\textwidth]{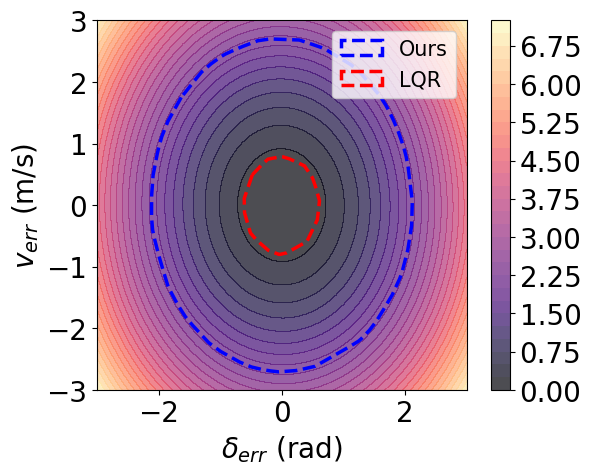}
}
\hfill
\subfigure[$\psi$-$\dot{\psi}$ plane]{
\includegraphics[width=0.230\textwidth]{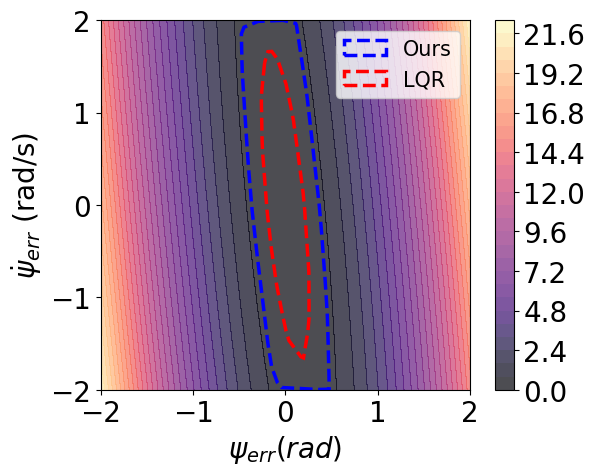}
}
\hfill
\subfigure[$\dot{\psi}$-$\beta$ plane]{
\includegraphics[width=0.230\textwidth]{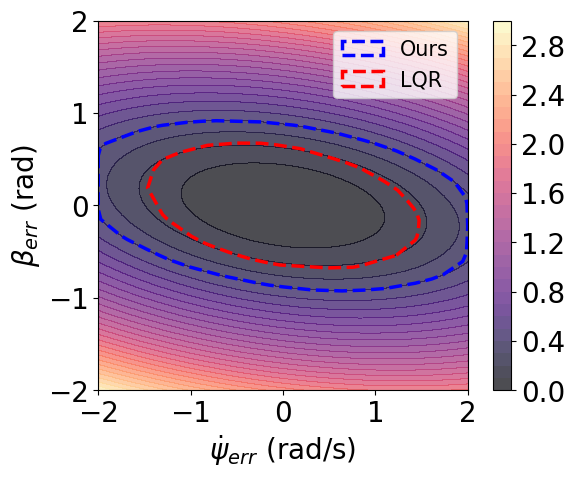}
}
\hfill
}
\end{figure}
\begin{figure}[!htbp]
\floatconts{fig:supple-car-roa-09}
{\caption{Car experiment (friction $\mu=0.1$,  reference speed $v=20m/s$)}}
{
\subfigure[$x$-$y$ plane]{
\includegraphics[width=0.230\textwidth]{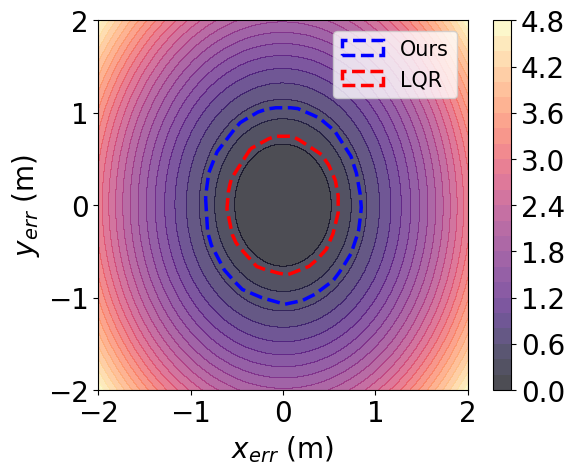}
}
\hfill
\subfigure[$v$-$\delta$ plane]{
\includegraphics[width=0.230\textwidth]{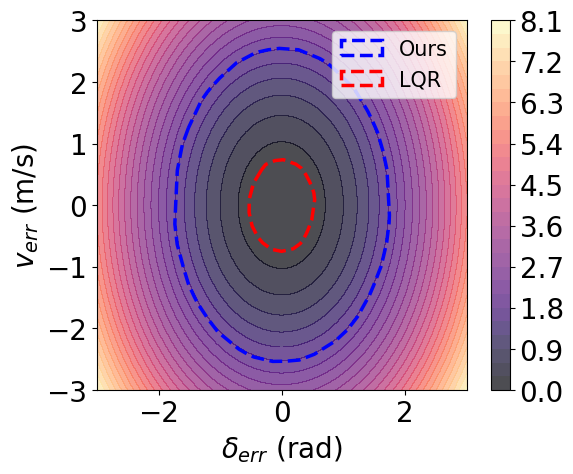}
}
\hfill
\subfigure[$\psi$-$\dot{\psi}$ plane]{
\includegraphics[width=0.230\textwidth]{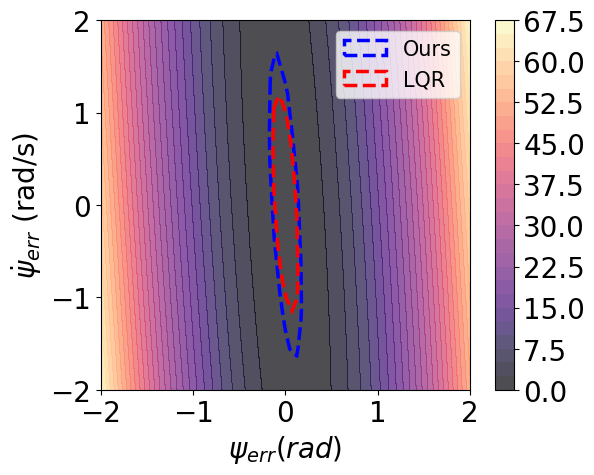}
}
\hfill
\subfigure[$\dot{\psi}$-$\beta$ plane]{
\includegraphics[width=0.230\textwidth]{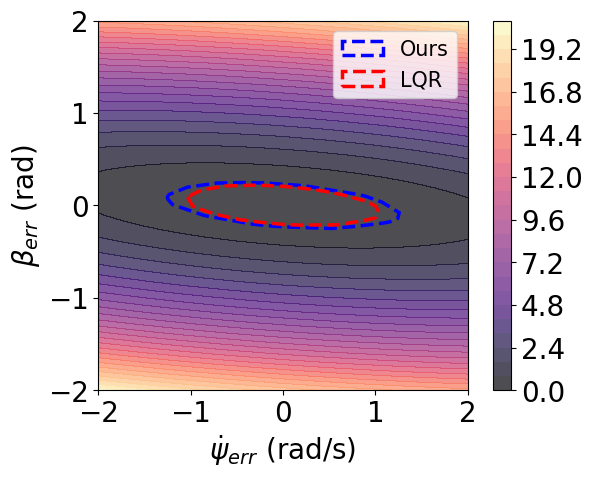}
}
\hfill
}
\end{figure}
\begin{figure}[!htbp]
\floatconts{fig:supple-car-roa-12}
{\caption{Car experiment (friction $\mu=1.0$,  reference speed $v=30m/s$)}}
{
\subfigure[$x$-$y$ plane]{
\includegraphics[width=0.230\textwidth]{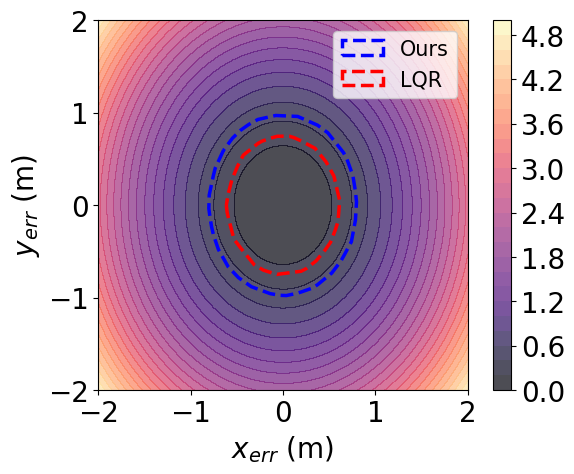}
}
\hfill
\subfigure[$v$-$\delta$ plane]{
\includegraphics[width=0.230\textwidth]{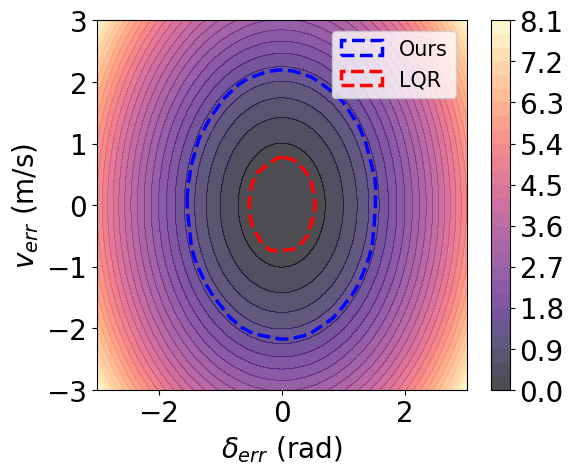}
}
\hfill
\subfigure[$\psi$-$\dot{\psi}$ plane]{
\includegraphics[width=0.230\textwidth]{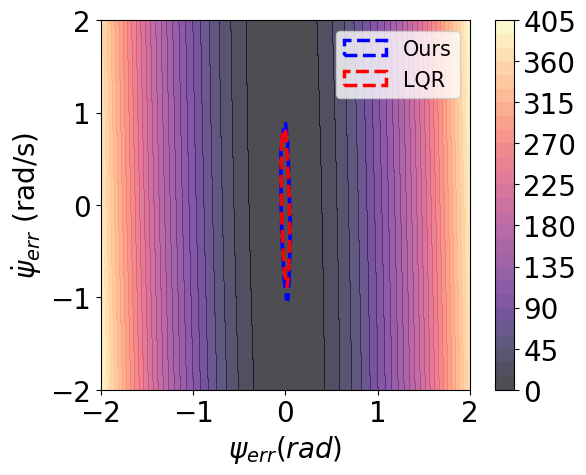}
}
\hfill
\subfigure[$\dot{\psi}$-$\beta$ plane]{
\includegraphics[width=0.230\textwidth]{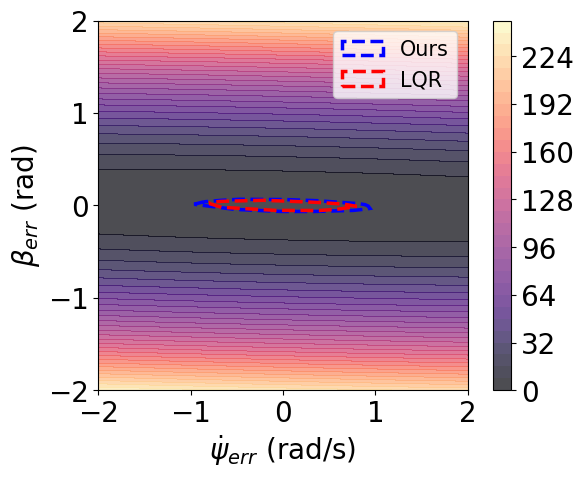}
}
\hfill
}
\end{figure}
\begin{figure}[!htbp]
\floatconts{fig:supple-car-roa-13}
{\caption{Car experiment (friction $\mu=0.1$,  reference speed $v=30m/s$)}}
{
\subfigure[$x$-$y$ plane]{
\includegraphics[width=0.230\textwidth]{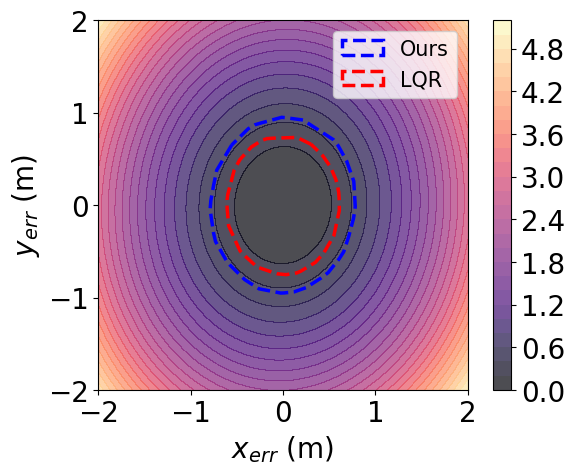}
}
\hfill
\subfigure[$v$-$\delta$ plane]{
\includegraphics[width=0.230\textwidth]{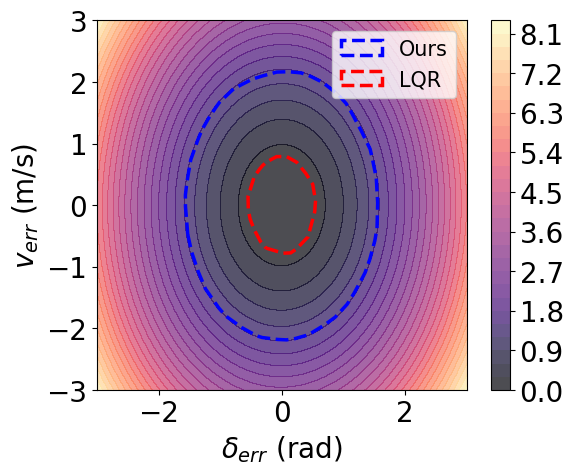}
}
\hfill
\subfigure[$\psi$-$\dot{\psi}$ plane]{
\includegraphics[width=0.230\textwidth]{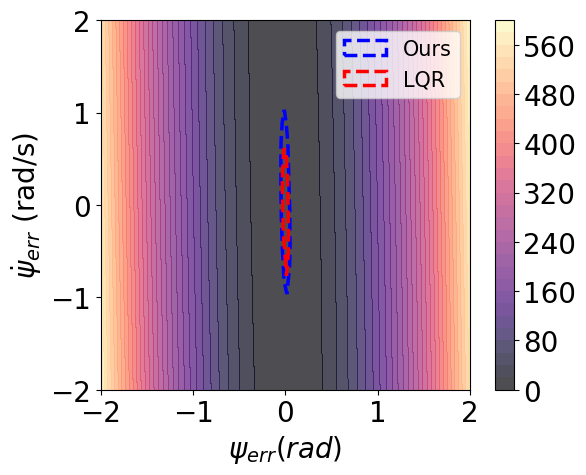}
}
\hfill
\subfigure[$\dot{\psi}$-$\beta$ plane]{
\includegraphics[width=0.230\textwidth]{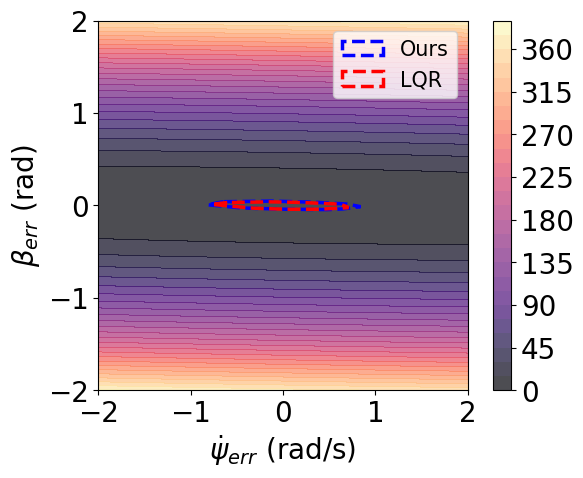}
}
\hfill
}
\end{figure}

\begin{figure}[!htbp]
\floatconts{fig:supple-pogo-roa-00}
{\caption{Pogobot experiment}}
{
\subfigure[apex velocity=1.0 m/s, apex height= 2.1 m]{
\includegraphics[width=0.230\textwidth]{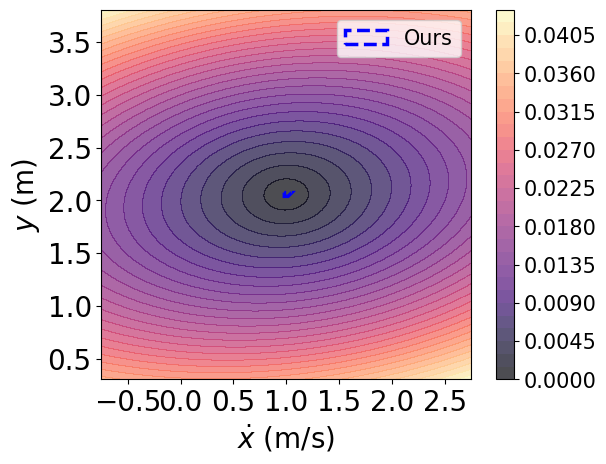}
}
\hfill
\subfigure[apex velocity=1.0 m/s, apex height= 2.3 m]{
\includegraphics[width=0.230\textwidth]{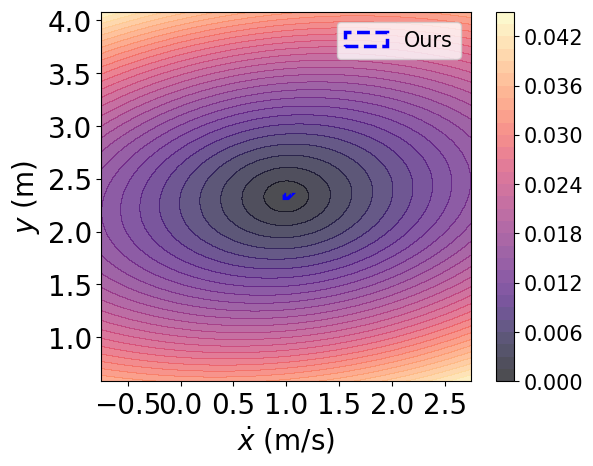}
}
\hfill
\subfigure[apex velocity=1.0 m/s, apex height= 2.6 m]{
\includegraphics[width=0.230\textwidth]{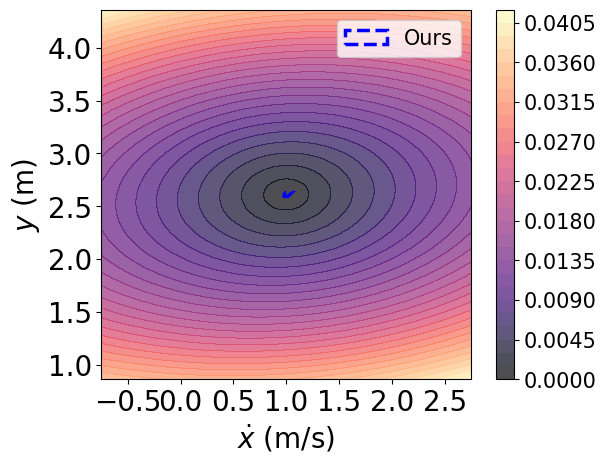}
}
\hfill
\subfigure[apex velocity=1.0 m/s, apex height= 2.9 m]{
\includegraphics[width=0.230\textwidth]{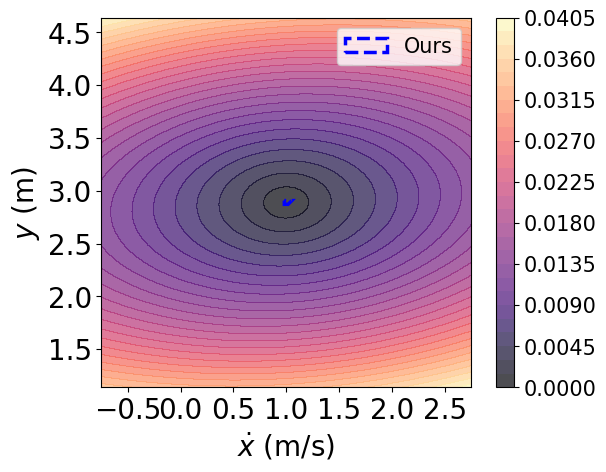}
}
\hfill
}
\end{figure}
\begin{figure}[!htbp]
\floatconts{fig:supple-pogo-roa-01}
{\caption{Pogobot experiment}}
{
\subfigure[apex velocity=1.0 m/s, apex height= 3.2 m]{
\includegraphics[width=0.230\textwidth]{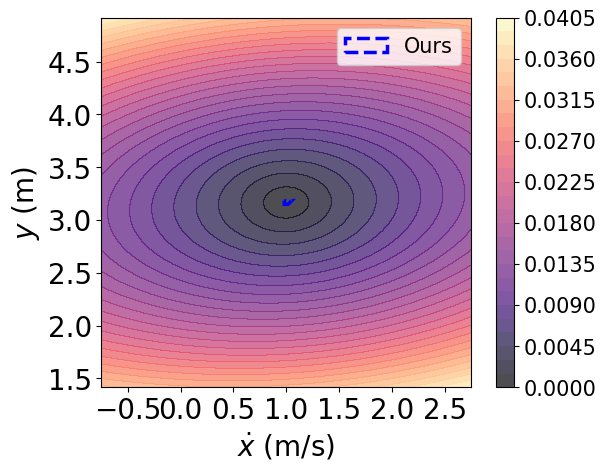}
}
\hfill
\subfigure[apex velocity=1.3 m/s, apex height= 2.1 m]{
\includegraphics[width=0.230\textwidth]{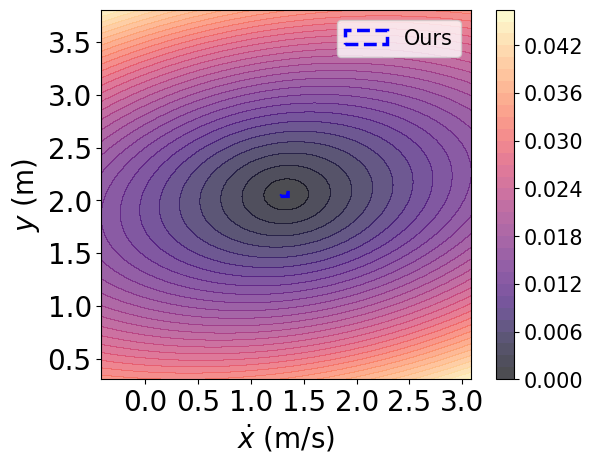}
}
\hfill
\subfigure[apex velocity=1.3 m/s, apex height= 2.3 m]{
\includegraphics[width=0.230\textwidth]{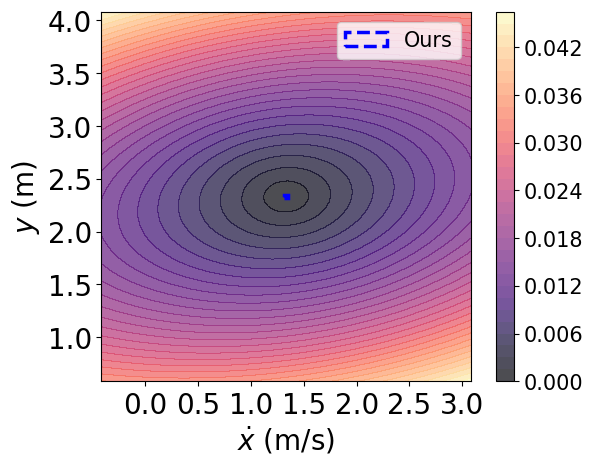}
}
\hfill
\subfigure[apex velocity=1.3 m/s, apex height= 2.6 m]{
\includegraphics[width=0.230\textwidth]{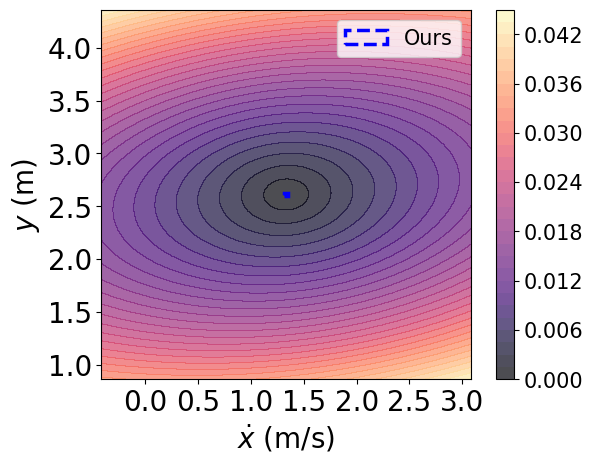}
}
\hfill
}
\end{figure}
\begin{figure}[!htbp]
\floatconts{fig:supple-pogo-roa-02}
{\caption{Pogobot experiment}}
{
\subfigure[apex velocity=1.3 m/s, apex height= 2.9 m]{
\includegraphics[width=0.230\textwidth]{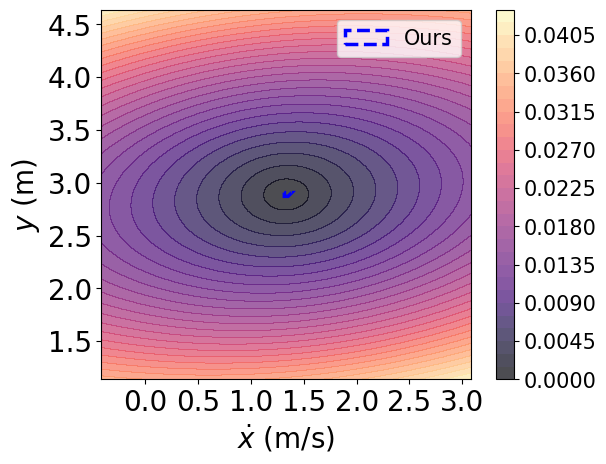}
}
\hfill
\subfigure[apex velocity=1.3 m/s, apex height= 3.2 m]{
\includegraphics[width=0.230\textwidth]{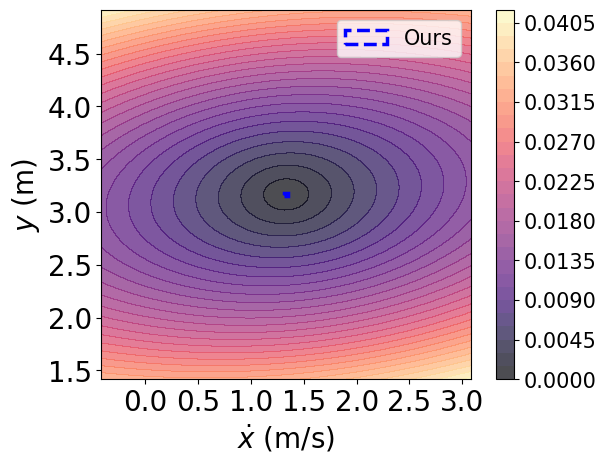}
}
\hfill
\subfigure[apex velocity=1.7 m/s, apex height= 2.1 m]{
\includegraphics[width=0.230\textwidth]{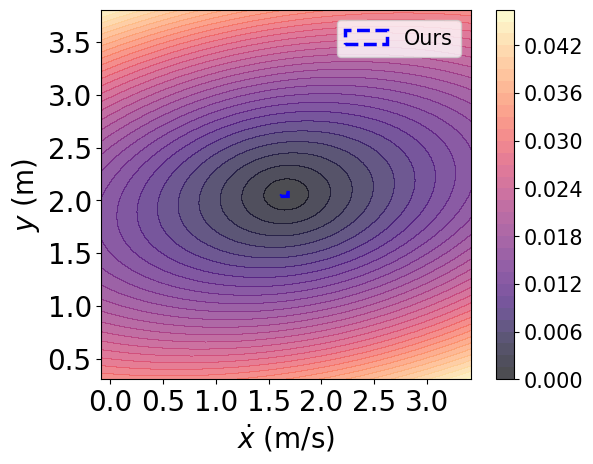}
}
\hfill
\subfigure[apex velocity=1.7 m/s, apex height= 2.3 m]{
\includegraphics[width=0.230\textwidth]{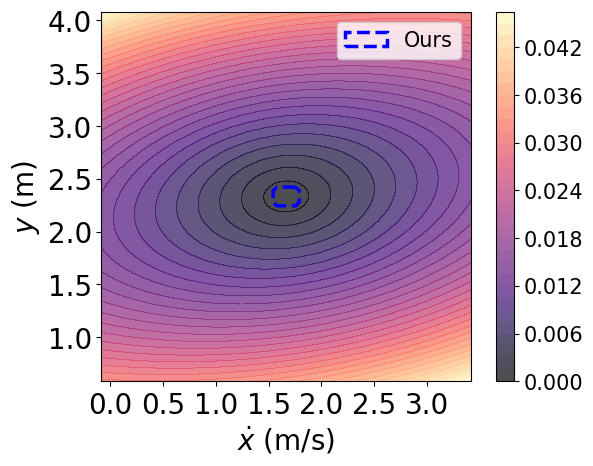}
}
\hfill
}
\end{figure}
\begin{figure}[!htbp]
\floatconts{fig:supple-pogo-roa-03}
{\caption{Pogobot experiment}}
{
\subfigure[apex velocity=1.7 m/s, apex height= 2.6 m]{
\includegraphics[width=0.230\textwidth]{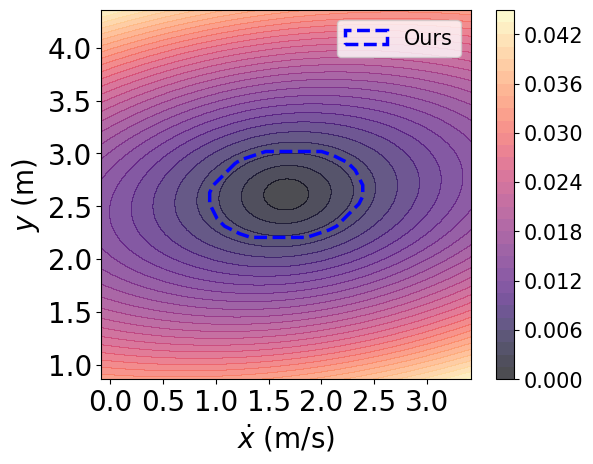}
}
\hfill
\subfigure[apex velocity=1.7 m/s, apex height= 2.9 m]{
\includegraphics[width=0.230\textwidth]{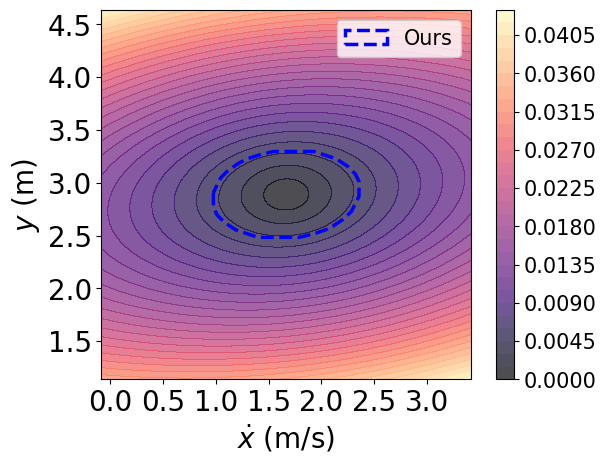}
}
\hfill
\subfigure[apex velocity=1.7 m/s, apex height= 3.2 m]{
\includegraphics[width=0.230\textwidth]{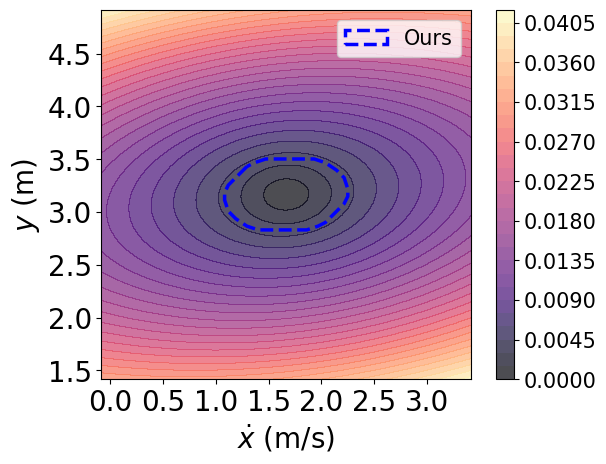}
}
\hfill
\subfigure[apex velocity=2.0 m/s, apex height= 2.1 m]{
\includegraphics[width=0.230\textwidth]{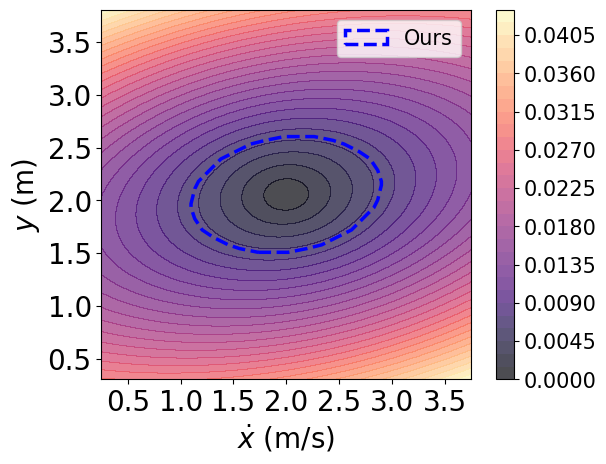}
}
\hfill
}
\end{figure}
\begin{figure}[!htbp]
\floatconts{fig:supple-pogo-roa-04}
{\caption{Pogobot experiment}}
{
\subfigure[apex velocity=2.0 m/s, apex height= 2.3 m]{
\includegraphics[width=0.230\textwidth]{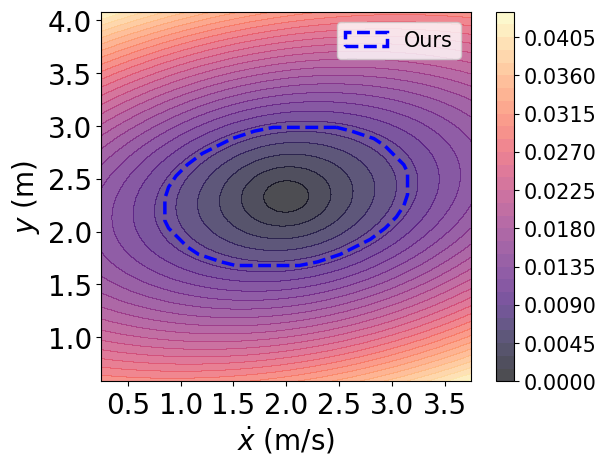}
}
\hfill
\subfigure[apex velocity=2.0 m/s, apex height= 2.6 m]{
\includegraphics[width=0.230\textwidth]{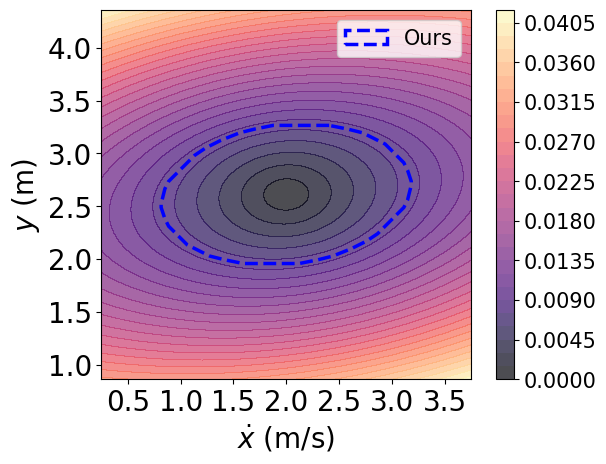}
}
\hfill
\subfigure[apex velocity=2.0 m/s, apex height= 2.9 m]{
\includegraphics[width=0.230\textwidth]{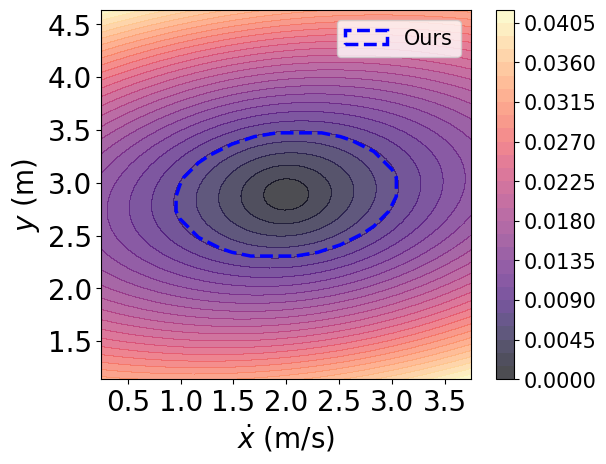}
}
\hfill
\subfigure[apex velocity=2.0 m/s, apex height= 3.2 m]{
\includegraphics[width=0.230\textwidth]{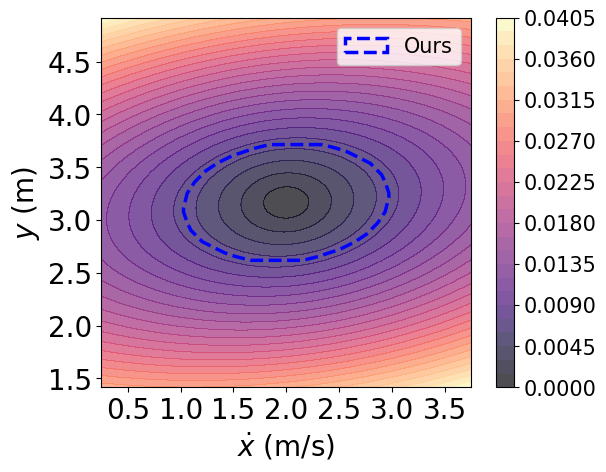}
}
\hfill
}
\end{figure}
\begin{figure}[!htbp]
\floatconts{fig:supple-pogo-roa-05}
{\caption{Pogobot experiment}}
{
\subfigure[apex velocity=2.3 m/s, apex height= 2.1 m]{
\includegraphics[width=0.230\textwidth]{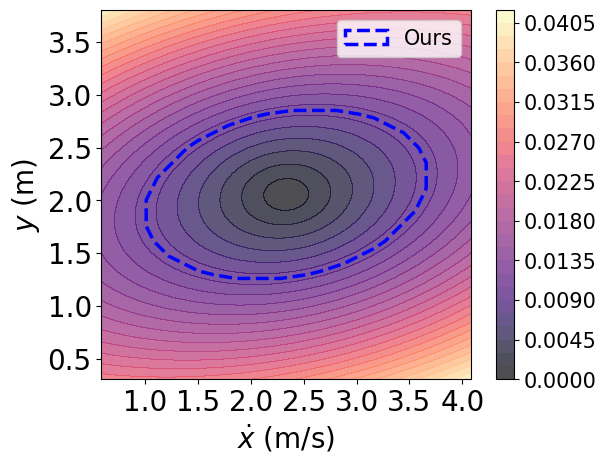}
}
\hfill
\subfigure[apex velocity=2.3 m/s, apex height= 2.3 m]{
\includegraphics[width=0.230\textwidth]{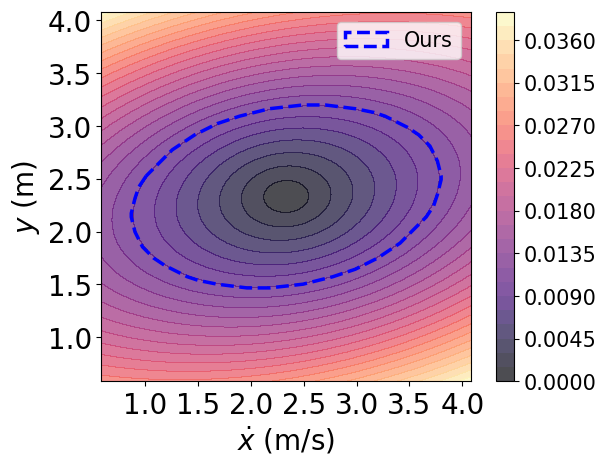}
}
\hfill
\subfigure[apex velocity=2.3 m/s, apex height= 2.6 m]{
\includegraphics[width=0.230\textwidth]{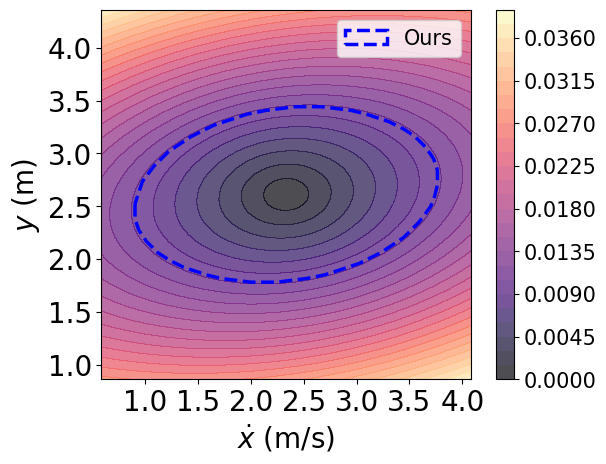}
}
\hfill
\subfigure[apex velocity=2.3 m/s, apex height= 2.9 m]{
\includegraphics[width=0.230\textwidth]{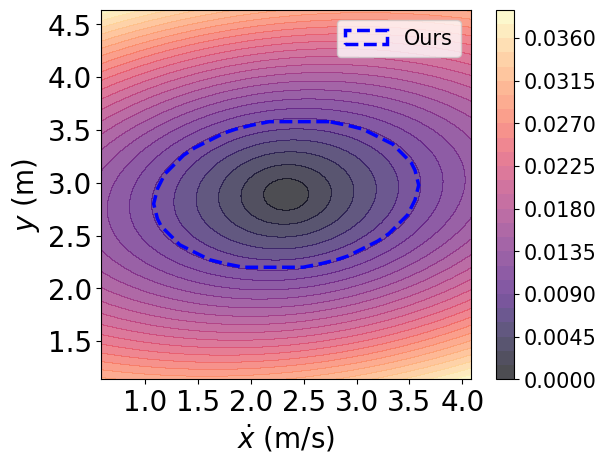}
}
\hfill
}
\end{figure}
\begin{figure}[!htbp]
\floatconts{fig:supple-pogo-roa-06}
{\caption{Pogobot experiment}}
{
\subfigure[apex velocity=2.3 m/s, apex height= 3.2 m]{
\includegraphics[width=0.230\textwidth]{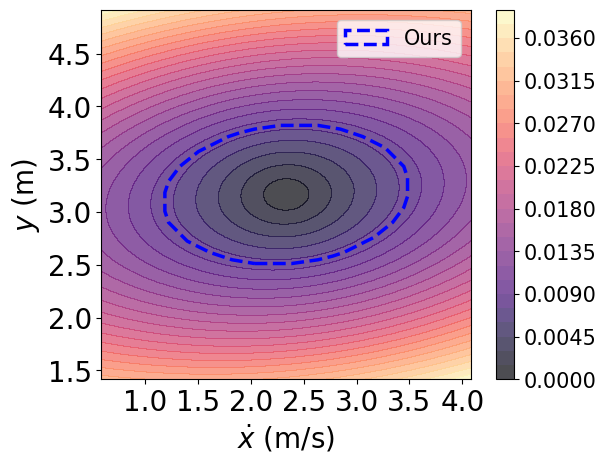}
}
\hfill
\subfigure[apex velocity=2.7 m/s, apex height= 2.1 m]{
\includegraphics[width=0.230\textwidth]{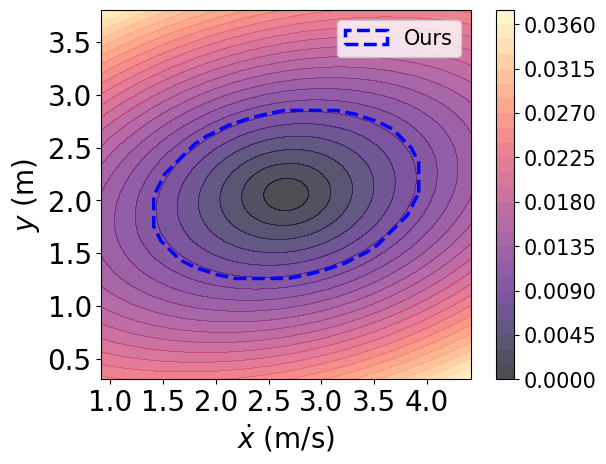}
}
\hfill
\subfigure[apex velocity=2.7 m/s, apex height= 2.3 m]{
\includegraphics[width=0.230\textwidth]{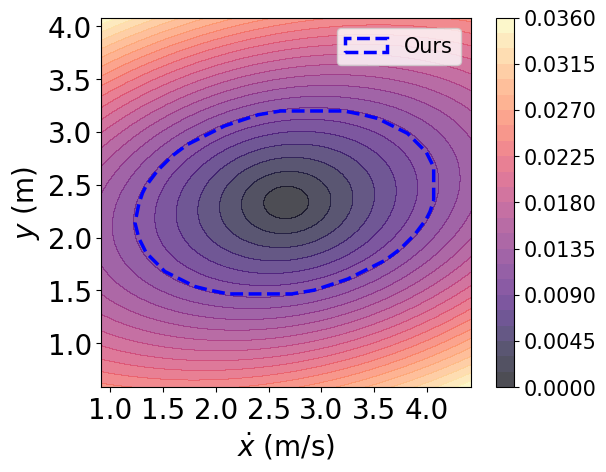}
}
\hfill
\subfigure[apex velocity=2.7 m/s, apex height= 2.6 m]{
\includegraphics[width=0.230\textwidth]{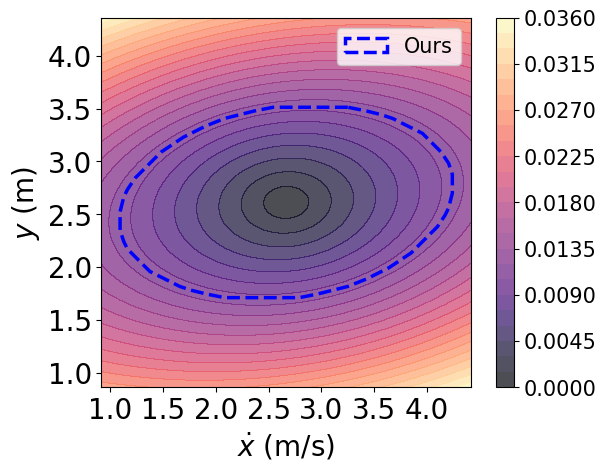}
}
\hfill
}
\end{figure}
\begin{figure}[!htbp]
\floatconts{fig:supple-pogo-roa-07}
{\caption{Pogobot experiment}}
{
\subfigure[apex velocity=2.7 m/s, apex height= 2.9 m]{
\includegraphics[width=0.230\textwidth]{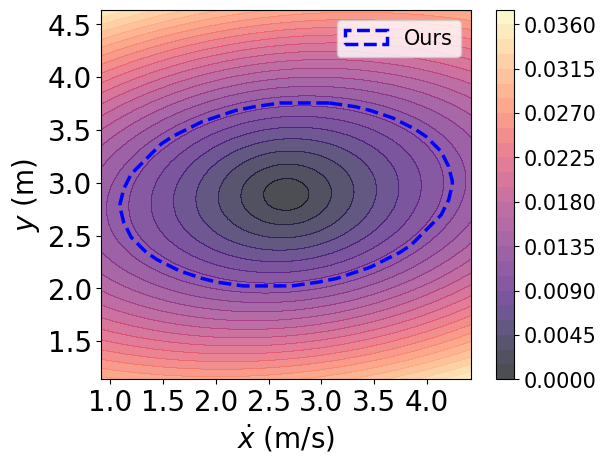}
}
\hfill
\subfigure[apex velocity=2.7 m/s, apex height= 3.2 m]{
\includegraphics[width=0.230\textwidth]{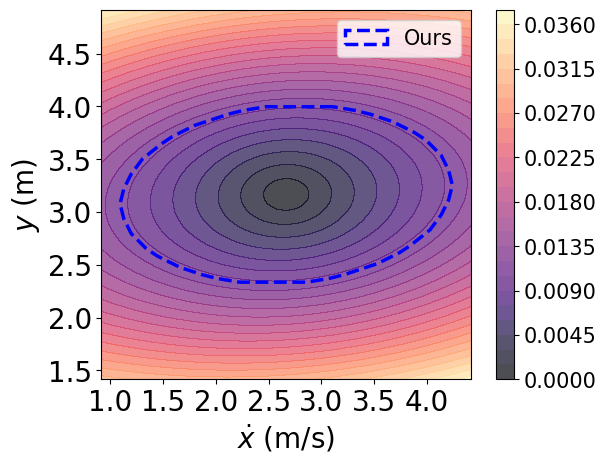}
}
\hfill
\subfigure[apex velocity=3.0 m/s, apex height= 2.1 m]{
\includegraphics[width=0.230\textwidth]{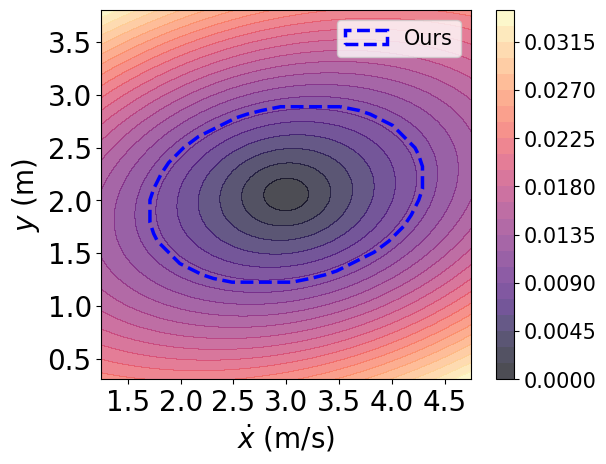}
}
\hfill
\subfigure[apex velocity=3.0 m/s, apex height= 2.3 m]{
\includegraphics[width=0.230\textwidth]{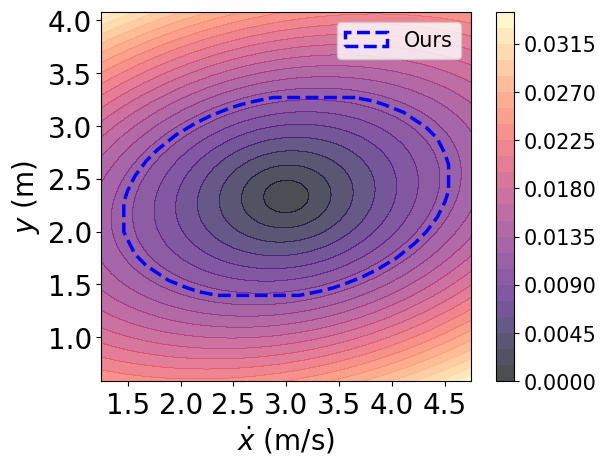}
}
\hfill
}
\end{figure}
\begin{figure}[!htbp]
\floatconts{fig:supple-pogo-roa-08}
{\caption{Pogobot experiment}}
{
\subfigure[apex velocity=3.0 m/s, apex height= 2.6 m]{
\includegraphics[width=0.230\textwidth]{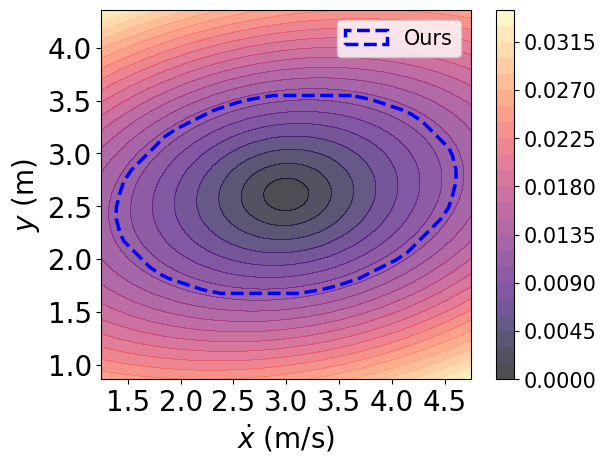}
}
\hfill
\subfigure[apex velocity=3.0 m/s, apex height= 2.9 m]{
\includegraphics[width=0.230\textwidth]{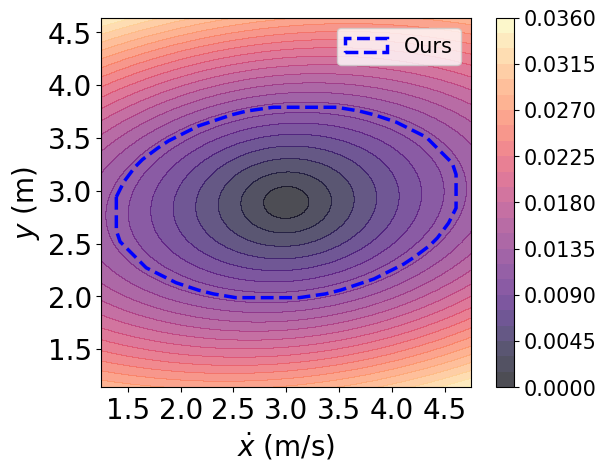}
}
\hfill
\subfigure[apex velocity=3.0 m/s, apex height= 3.2 m]{
\includegraphics[width=0.230\textwidth]{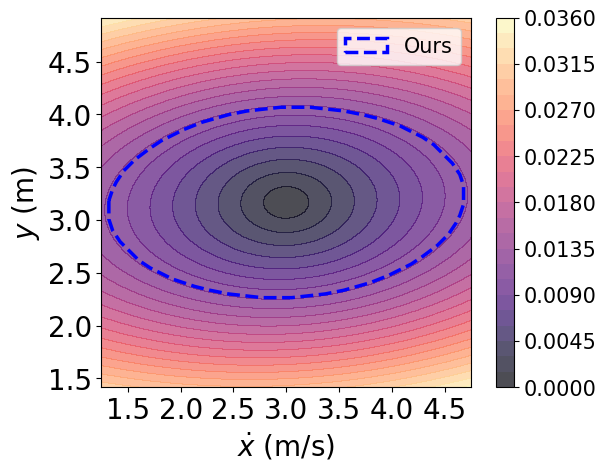}
}
\hfill
\subfigure[apex velocity=3.3 m/s, apex height= 2.1 m]{
\includegraphics[width=0.230\textwidth]{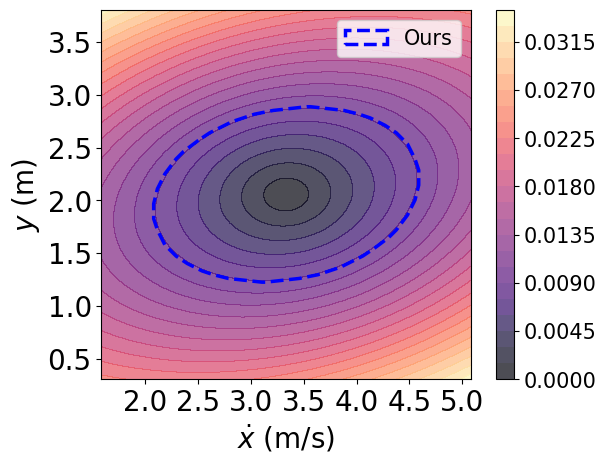}
}
\hfill
}
\end{figure}
\begin{figure}[!htbp]
\floatconts{fig:supple-pogo-roa-09}
{\caption{Pogobot experiment}}
{
\subfigure[apex velocity=3.3 m/s, apex height= 2.3 m]{
\includegraphics[width=0.230\textwidth]{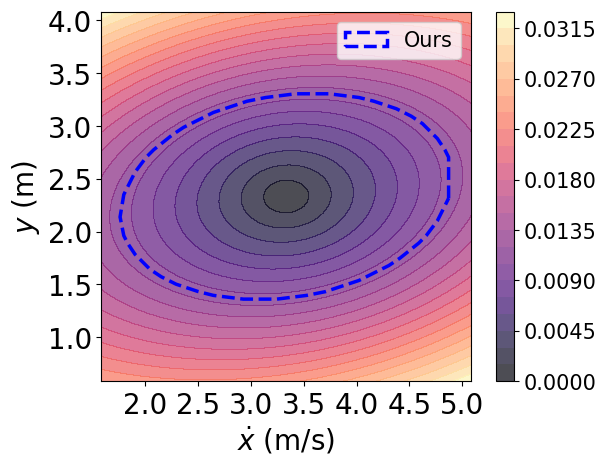}
}
\hfill
\subfigure[apex velocity=3.3 m/s, apex height= 2.6 m]{
\includegraphics[width=0.230\textwidth]{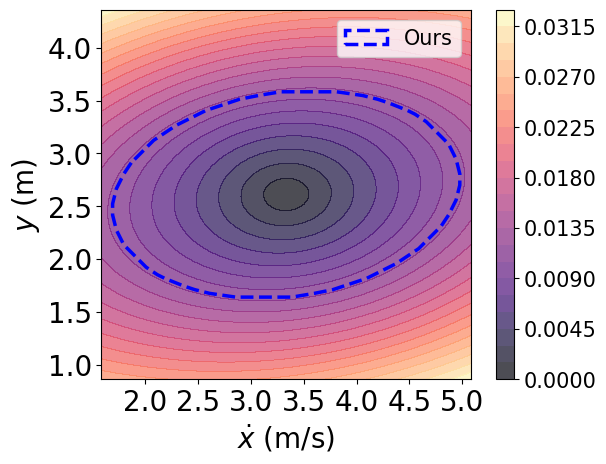}
}
\hfill
\subfigure[apex velocity=3.3 m/s, apex height= 2.9 m]{
\includegraphics[width=0.230\textwidth]{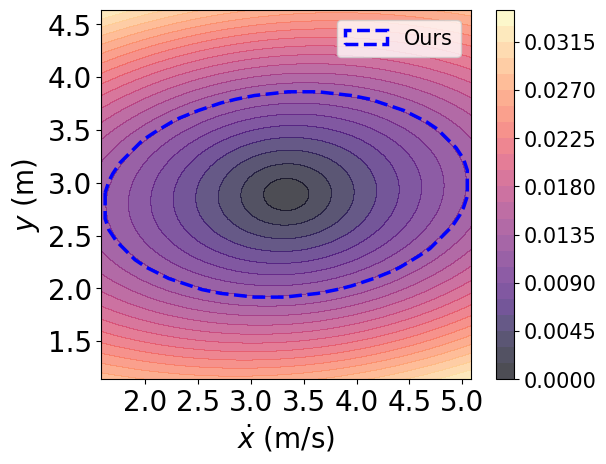}
}
\hfill
\subfigure[apex velocity=3.3 m/s, apex height= 3.2 m]{
\includegraphics[width=0.230\textwidth]{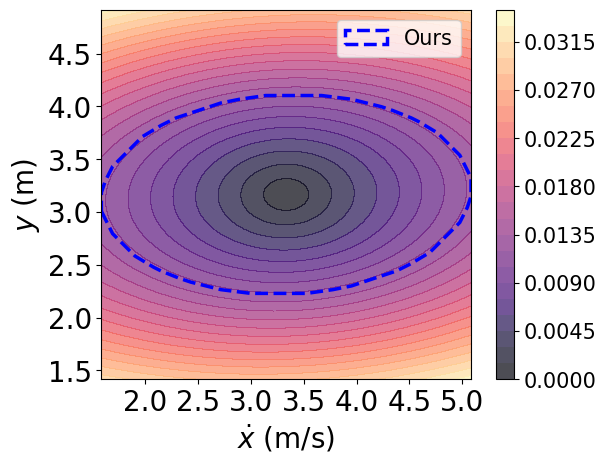}
}
\hfill
}
\end{figure}
\begin{figure}[!htbp]
\floatconts{fig:supple-pogo-roa-10}
{\caption{Pogobot experiment}}
{
\subfigure[apex velocity=3.7 m/s, apex height= 2.1 m]{
\includegraphics[width=0.230\textwidth]{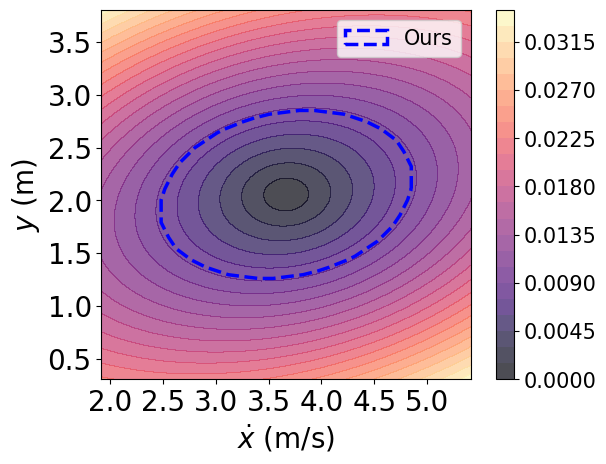}
}
\hfill
\subfigure[apex velocity=3.7 m/s, apex height= 2.3 m]{
\includegraphics[width=0.230\textwidth]{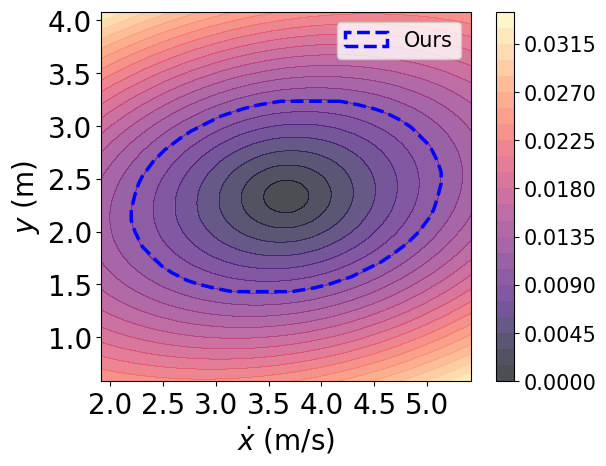}
}
\hfill
\subfigure[apex velocity=3.7 m/s, apex height= 2.6 m]{
\includegraphics[width=0.230\textwidth]{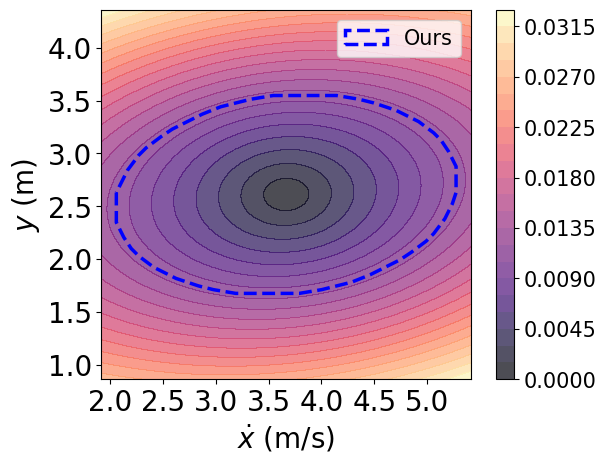}
}
\hfill
\subfigure[apex velocity=3.7 m/s, apex height= 2.9 m]{
\includegraphics[width=0.230\textwidth]{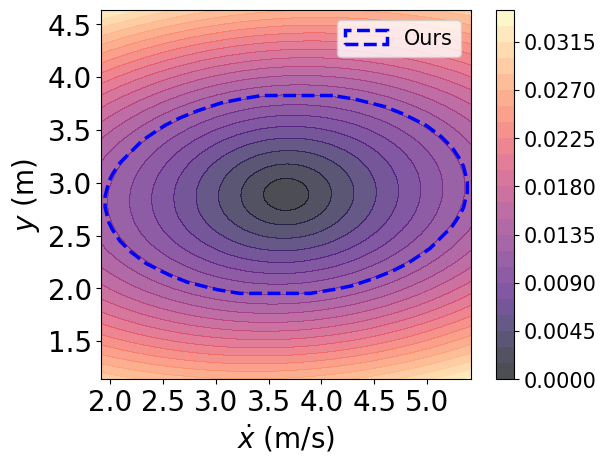}
}
\hfill
}
\end{figure}
\begin{figure}[!htbp]
\floatconts{fig:supple-pogo-roa-11}
{\caption{Pogobot experiment}}
{
\subfigure[apex velocity=3.7 m/s, apex height= 3.2 m]{
\includegraphics[width=0.230\textwidth]{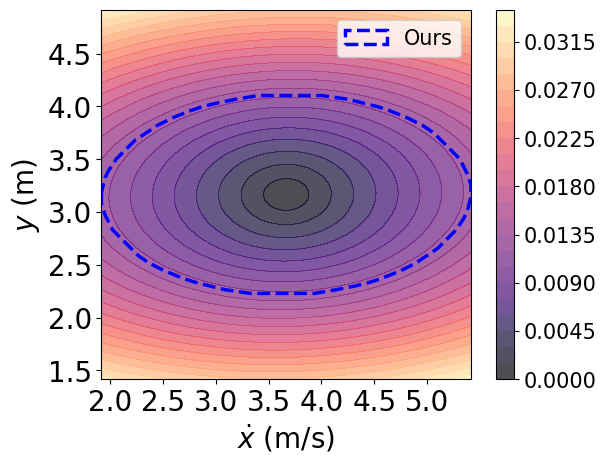}
}
\hfill
\subfigure[apex velocity=4.0 m/s, apex height= 2.1 m]{
\includegraphics[width=0.230\textwidth]{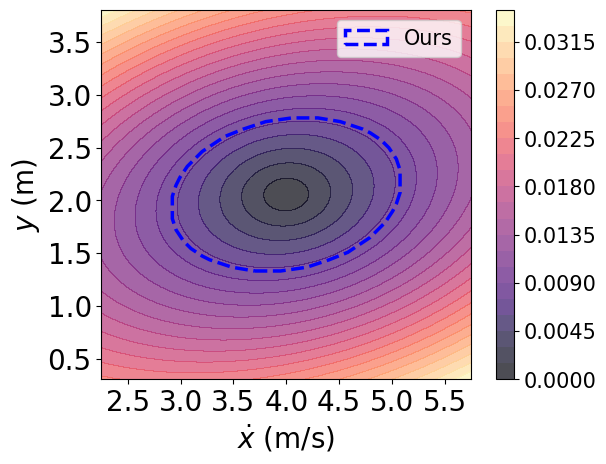}
}
\hfill
\subfigure[apex velocity=4.0 m/s, apex height= 2.3 m]{
\includegraphics[width=0.230\textwidth]{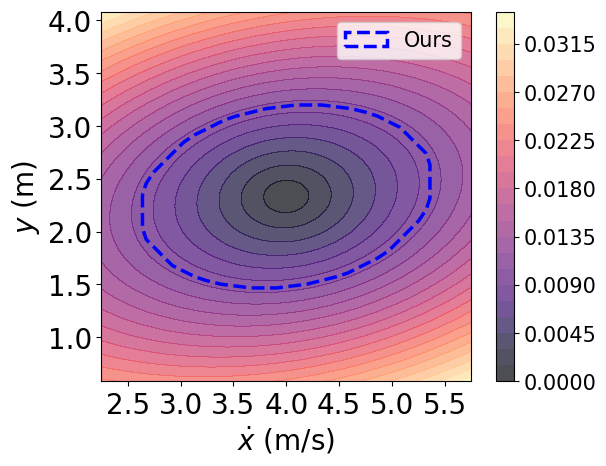}
}
\hfill
\subfigure[apex velocity=4.0 m/s, apex height= 2.6 m]{
\includegraphics[width=0.230\textwidth]{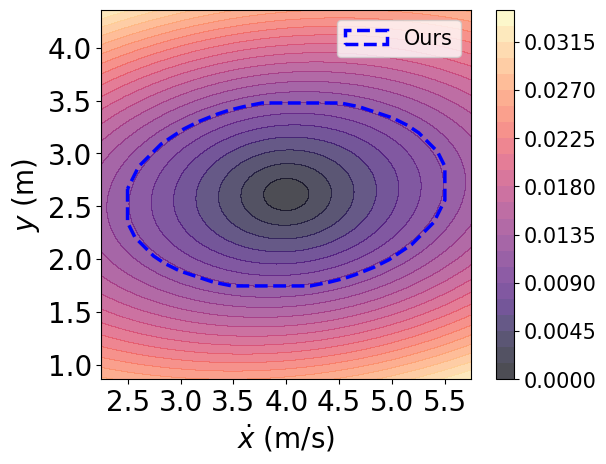}
}
\hfill
}
\end{figure}

\begin{figure}[!htbp]
\floatconts{fig:supple-walker-roa-00}
{\caption{Bipedal walker experiment (reference gait $q_1^{ref}$=0.05rad)}}
{
\subfigure[$x$-$y$ plane]{
\includegraphics[width=0.230\textwidth]{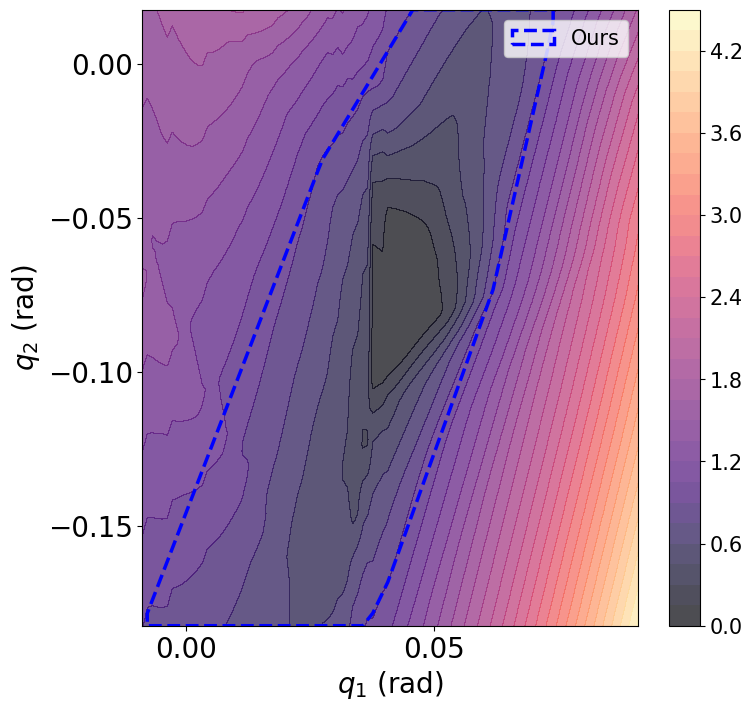}
}
\hfill
\subfigure[$q_2$-$\dot{q}_1$ plane]{
\includegraphics[width=0.230\textwidth]{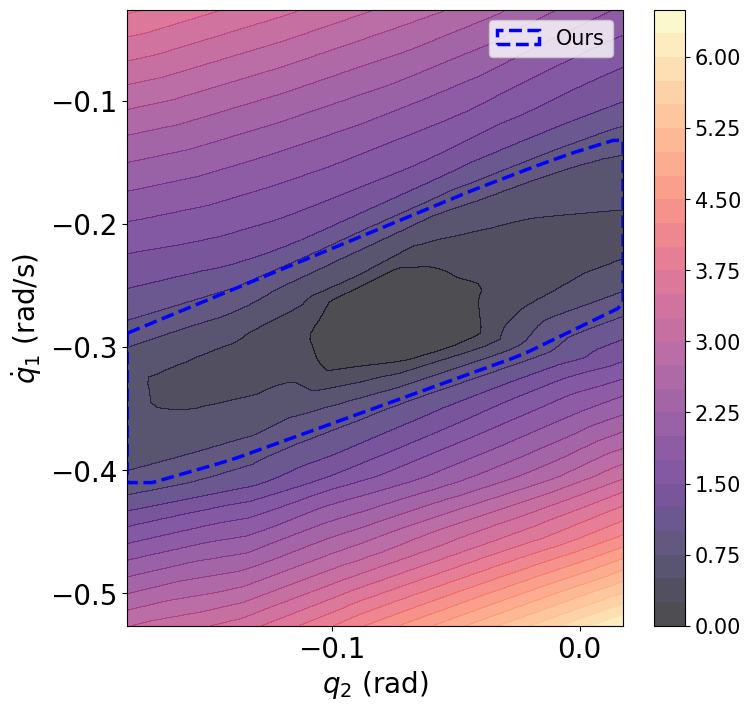}
}
\hfill
\subfigure[$\dot{q}_1$-$\dot{q}_2$ plane]{
\includegraphics[width=0.230\textwidth]{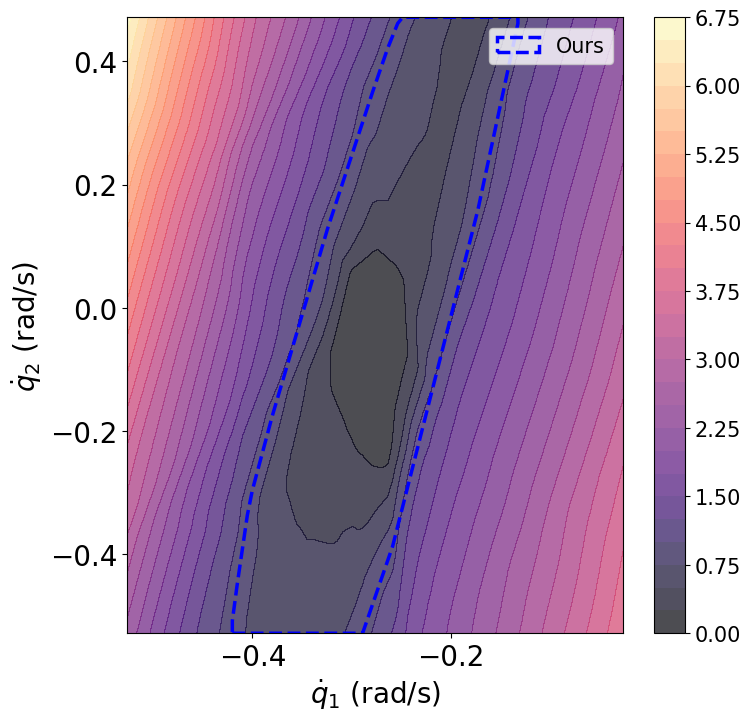}
}
\hfill
\subfigure[$q_2$-$\dot{q}_2$ plane]{
\includegraphics[width=0.230\textwidth]{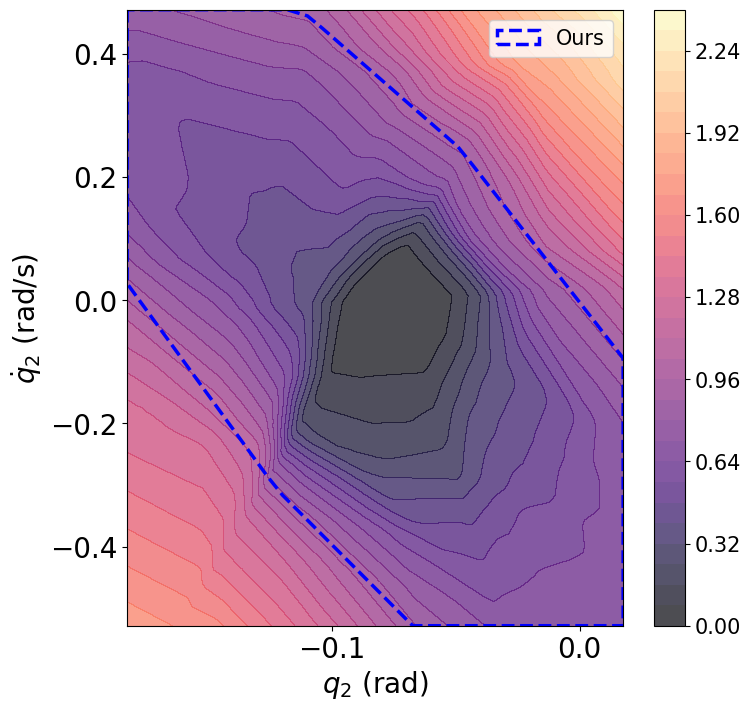}
}
\hfill
}
\end{figure}
\begin{figure}[!htbp]
\floatconts{fig:supple-walker-roa-01}
{\caption{Bipedal walker experiment (reference gait $q_1^{ref}$=0.08rad)}}
{
\subfigure[$x$-$y$ plane]{
\includegraphics[width=0.230\textwidth]{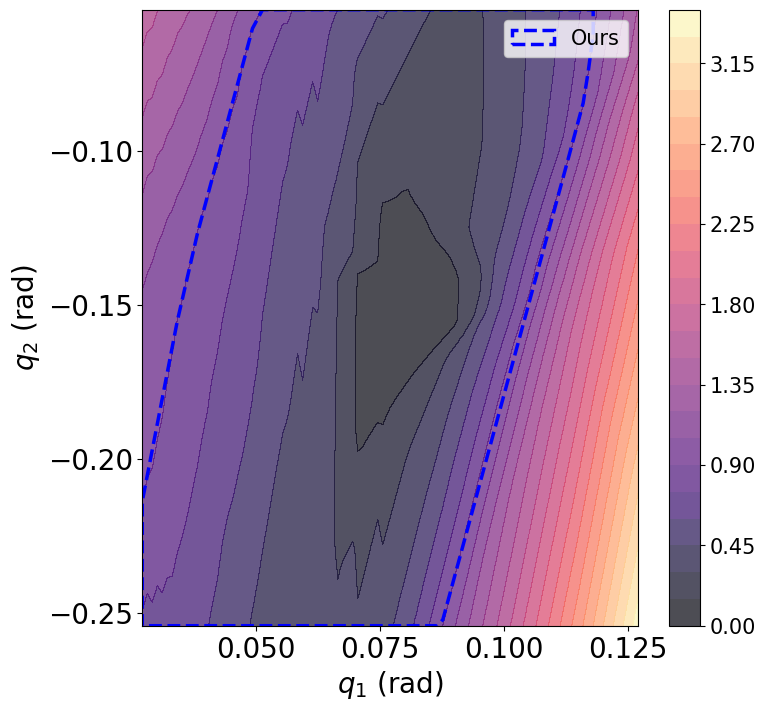}
}
\hfill
\subfigure[$q_2$-$\dot{q}_1$ plane]{
\includegraphics[width=0.230\textwidth]{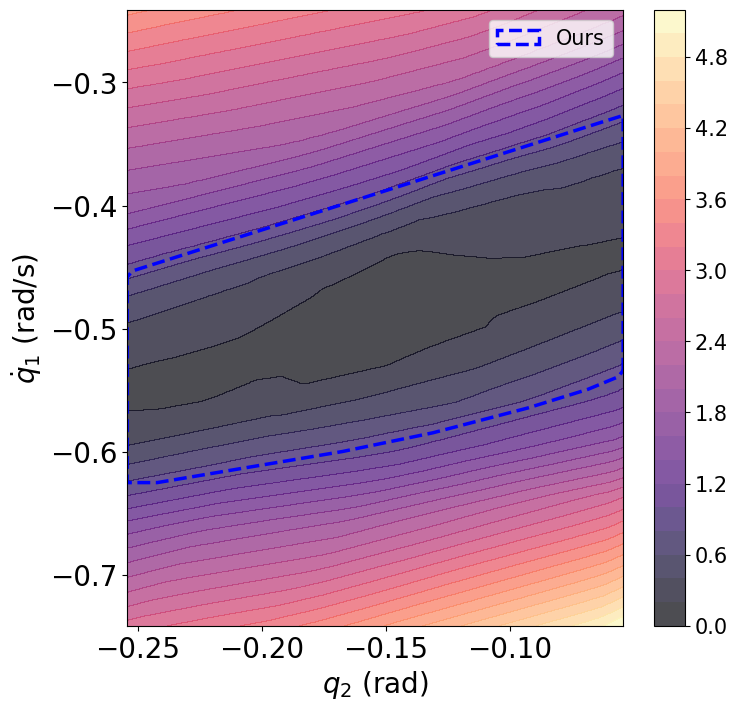}
}
\hfill
\subfigure[$\dot{q}_1$-$\dot{q}_2$ plane]{
\includegraphics[width=0.230\textwidth]{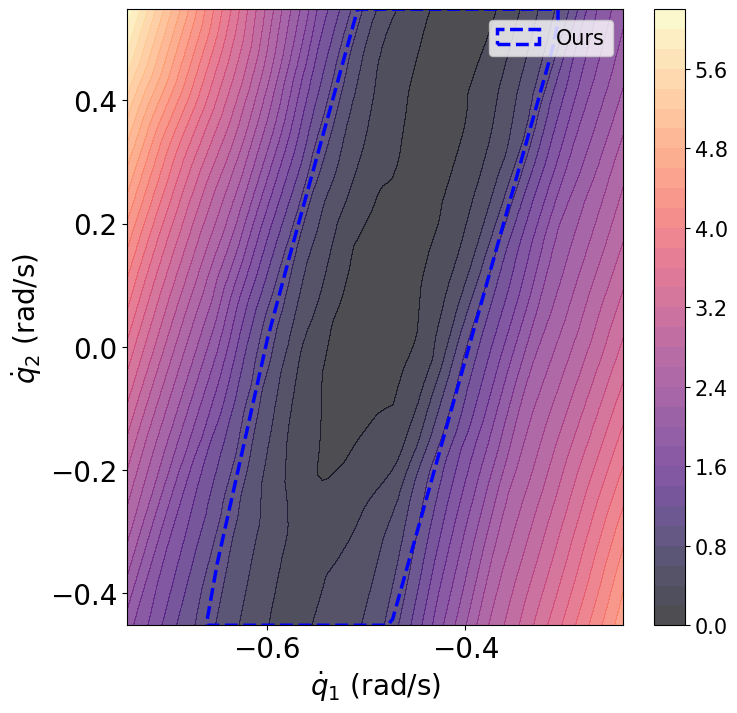}
}
\hfill
\subfigure[$q_2$-$\dot{q}_2$ plane]{
\includegraphics[width=0.230\textwidth]{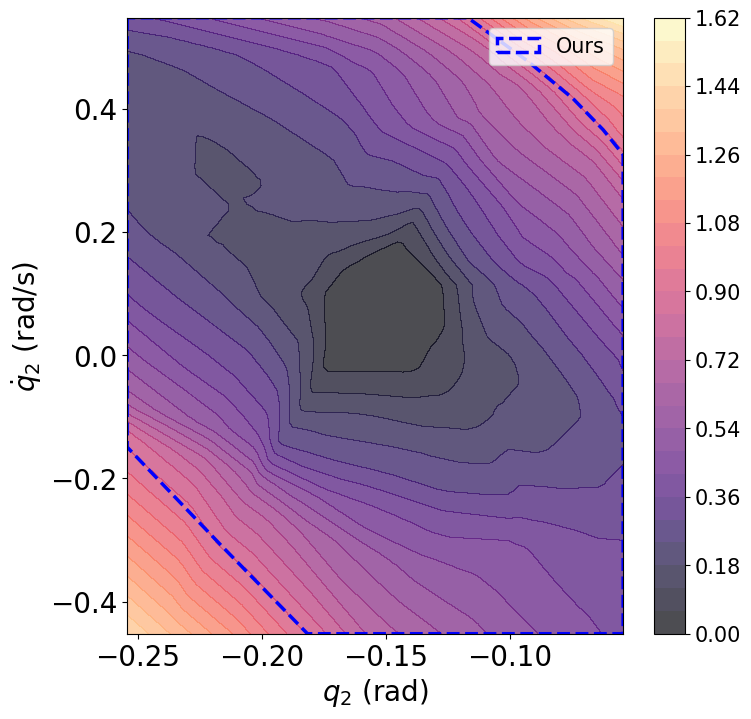}
}
\hfill
}
\end{figure}
\begin{figure}[!htbp]
\floatconts{fig:supple-walker-roa-02}
{\caption{Bipedal walker experiment (reference gait $q_1^{ref}$=0.10rad)}}
{
\subfigure[$x$-$y$ plane]{
\includegraphics[width=0.230\textwidth]{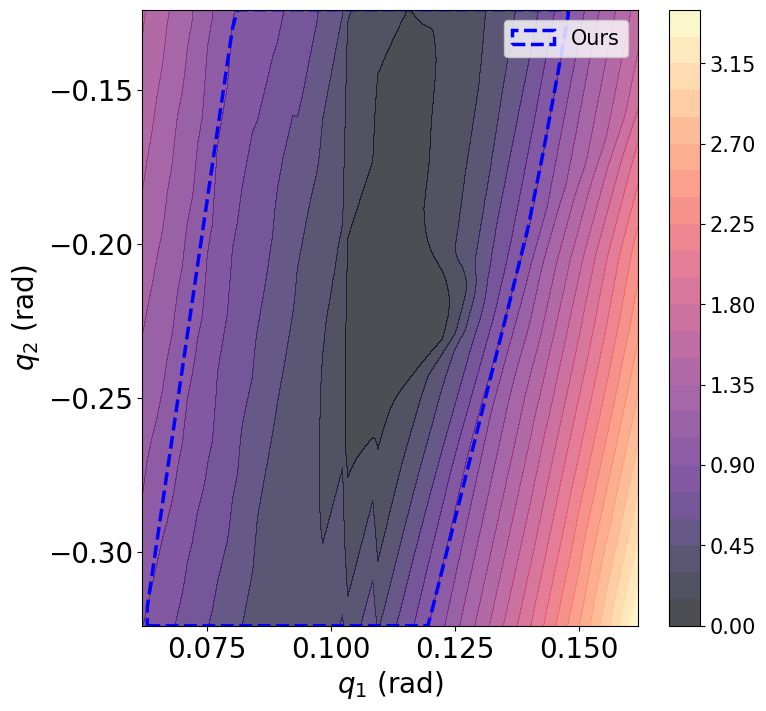}
}
\hfill
\subfigure[$q_2$-$\dot{q}_1$ plane]{
\includegraphics[width=0.230\textwidth]{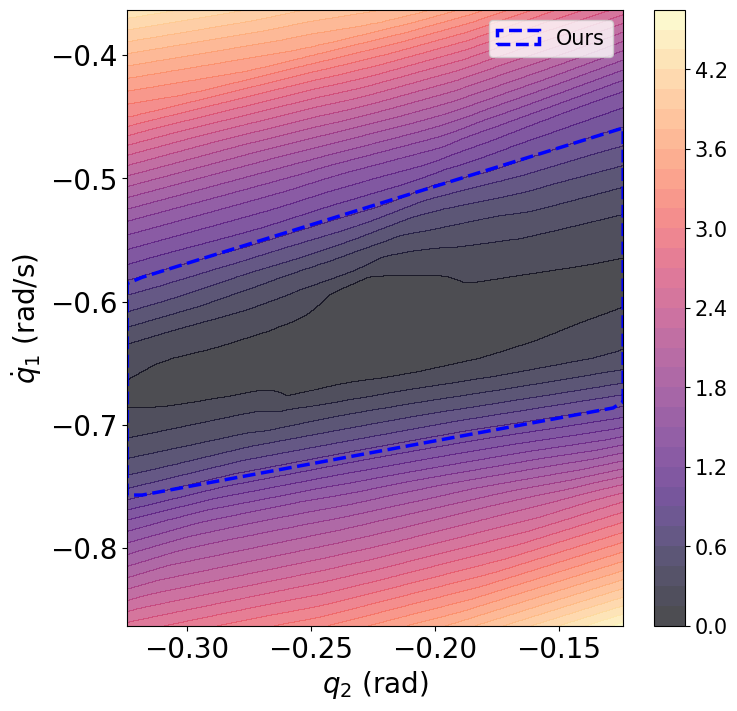}
}
\hfill
\subfigure[$\dot{q}_1$-$\dot{q}_2$ plane]{
\includegraphics[width=0.230\textwidth]{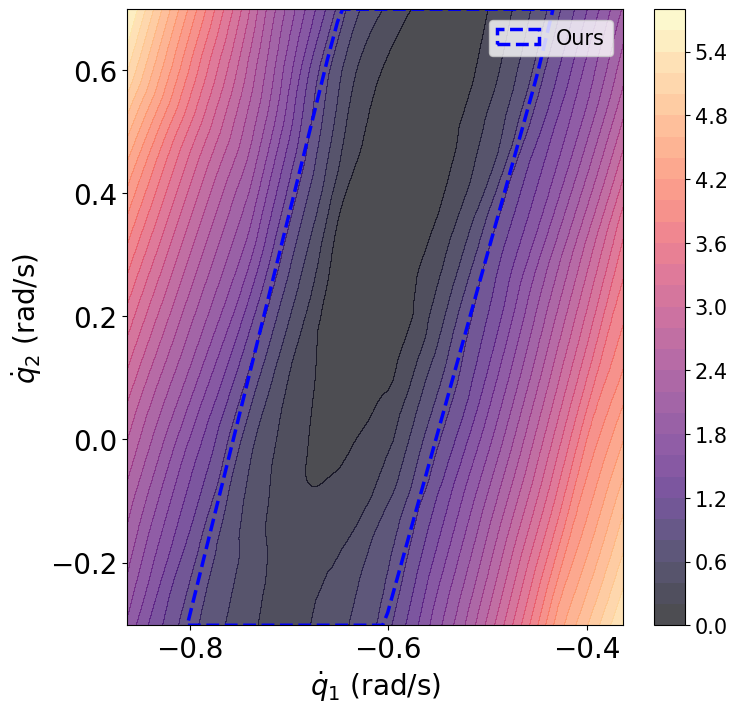}
}
\hfill
\subfigure[$q_2$-$\dot{q}_2$ plane]{
\includegraphics[width=0.230\textwidth]{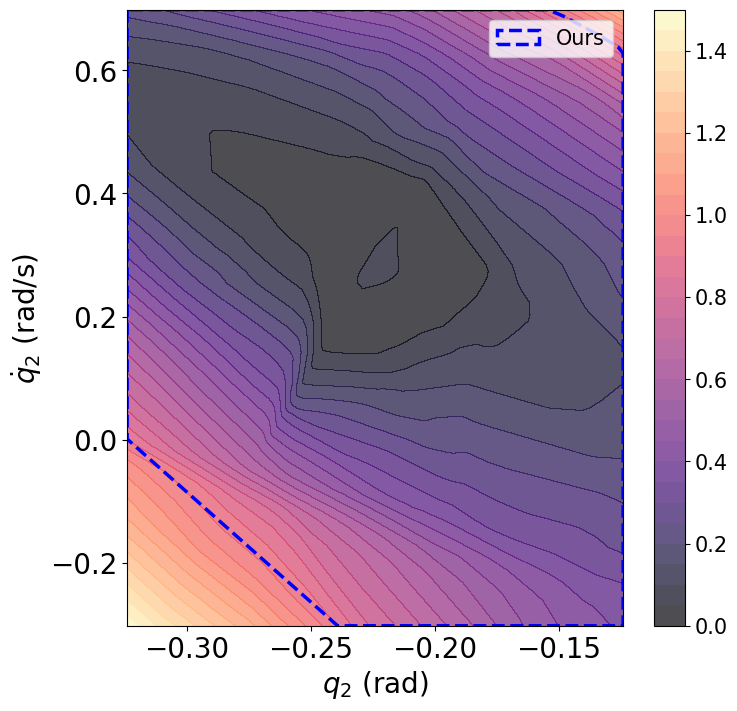}
}
\hfill
}
\end{figure}
\begin{figure}[!htbp]
\floatconts{fig:supple-walker-roa-03}
{\caption{Bipedal walker experiment (reference gait $q_1^{ref}$=0.13rad)}}
{
\subfigure[$x$-$y$ plane]{
\includegraphics[width=0.230\textwidth]{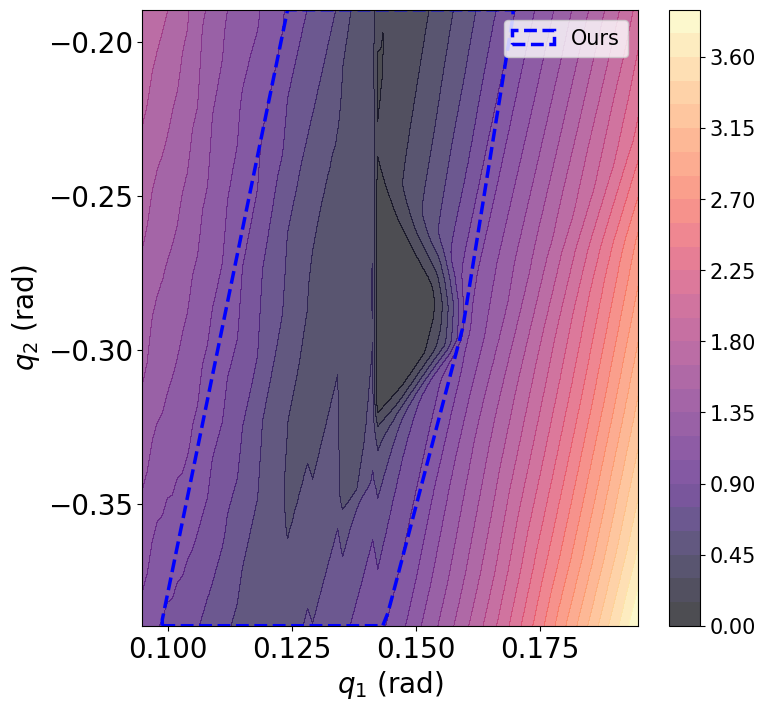}
}
\hfill
\subfigure[$q_2$-$\dot{q}_1$ plane]{
\includegraphics[width=0.230\textwidth]{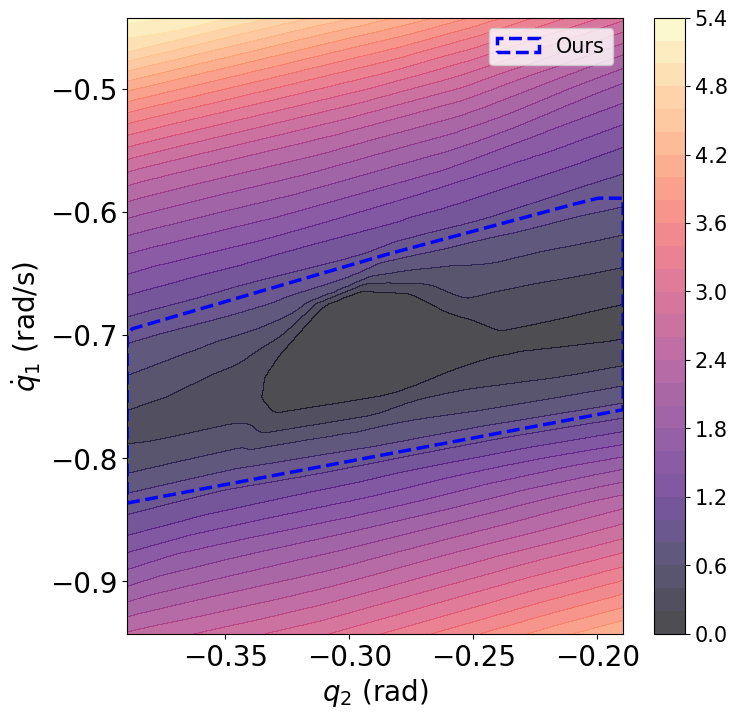}
}
\hfill
\subfigure[$\dot{q}_1$-$\dot{q}_2$ plane]{
\includegraphics[width=0.230\textwidth]{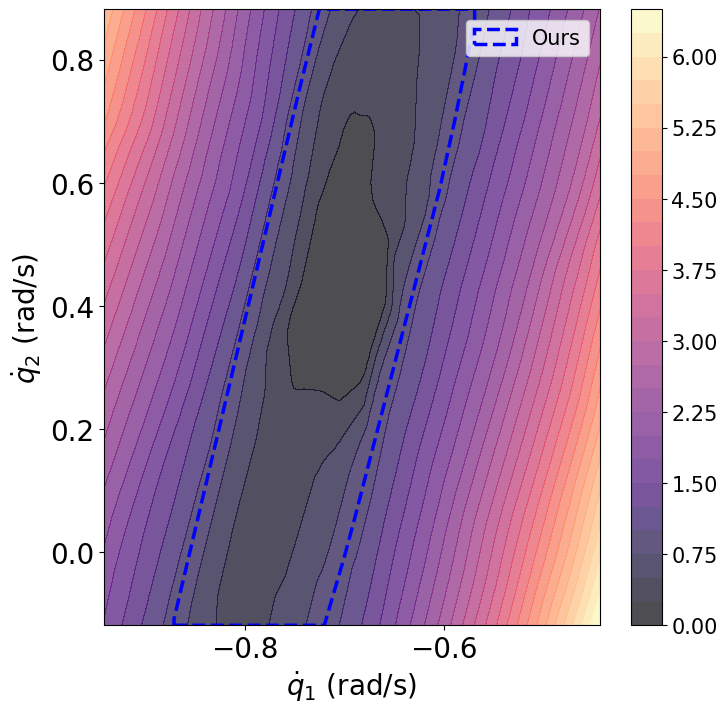}
}
\hfill
\subfigure[$q_2$-$\dot{q}_2$ plane]{
\includegraphics[width=0.230\textwidth]{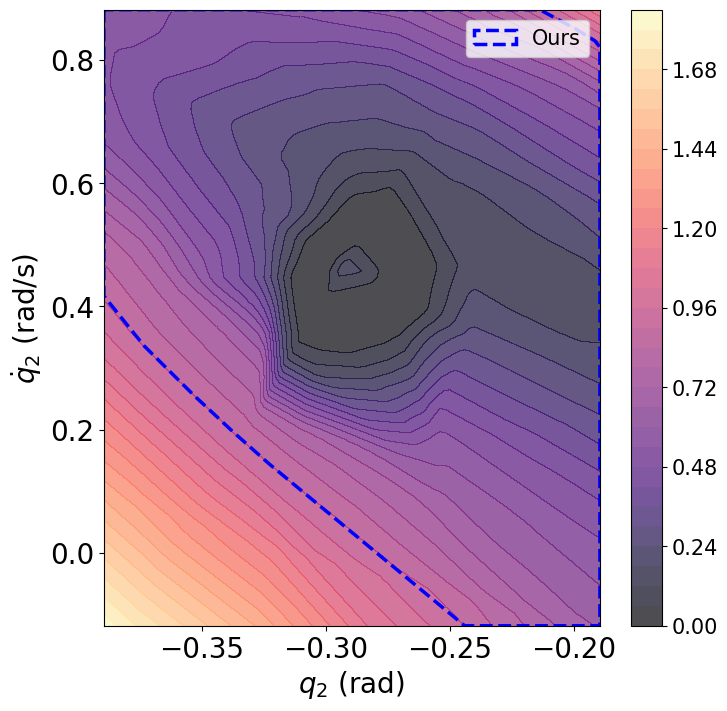}
}
\hfill
}
\end{figure}
\begin{figure}[!htbp]
\floatconts{fig:supple-walker-roa-04}
{\caption{Bipedal walker experiment (reference gait $q_1^{ref}$=0.18rad)}}
{
\subfigure[$x$-$y$ plane]{
\includegraphics[width=0.230\textwidth]{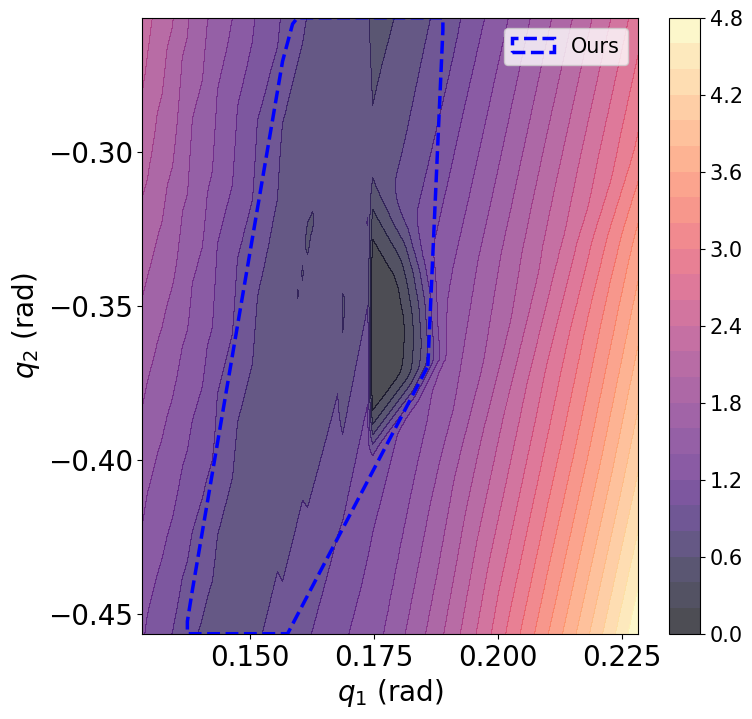}
}
\hfill
\subfigure[$q_2$-$\dot{q}_1$ plane]{
\includegraphics[width=0.230\textwidth]{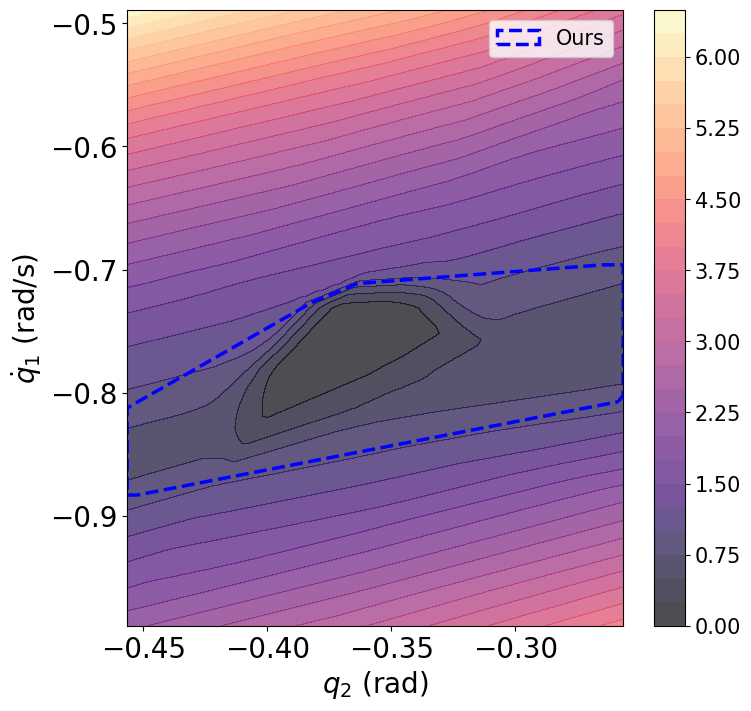}
}
\hfill
\subfigure[$\dot{q}_1$-$\dot{q}_2$ plane]{
\includegraphics[width=0.230\textwidth]{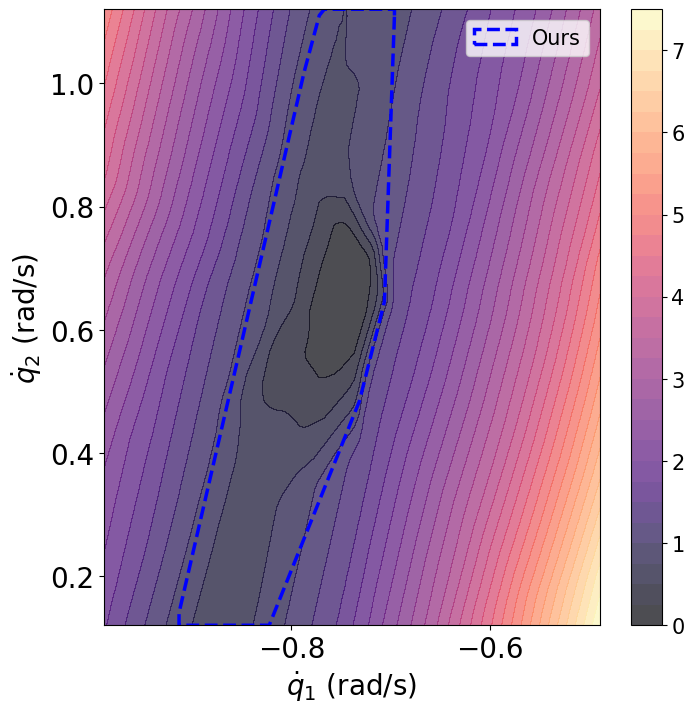}
}
\hfill
\subfigure[$q_2$-$\dot{q}_2$ plane]{
\includegraphics[width=0.230\textwidth]{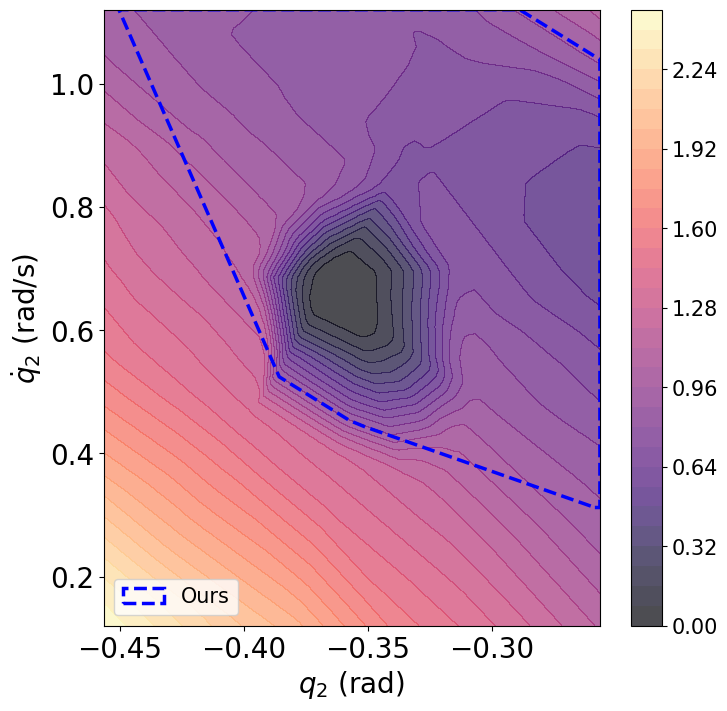}
}
\hfill
}
\end{figure}

\clearpage

\section{Visualization for the simulations}
\label{appendix-sim-viz}

From \figref{fig:supple-car-sim-04} to \figref{fig:supple-cgw-simx-430}, we visualize simulation results for all three experiments under different configurations. 

\begin{figure}[!htbp]
\floatconts{fig:supple-car-sim-04}
{\caption{Car simulation comparisons}}
{
\subfigure[Trial 00]{
\includegraphics[width=0.180\textwidth]{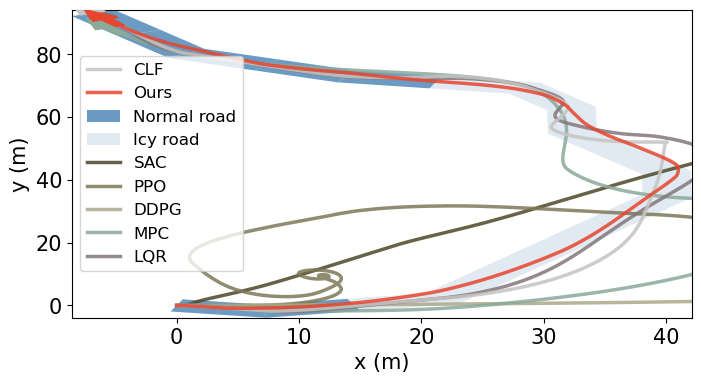}
}
\hfill
\subfigure[Trial 01]{
\includegraphics[width=0.180\textwidth]{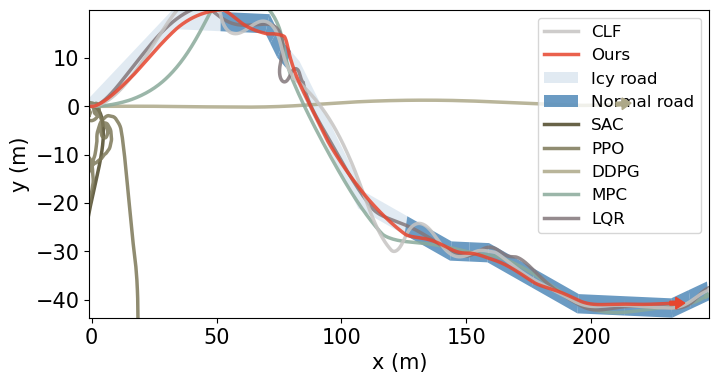}
}
\hfill
\subfigure[Trial 02]{
\includegraphics[width=0.180\textwidth]{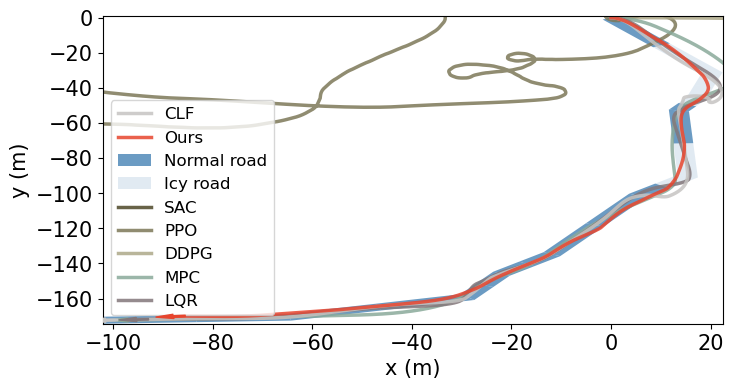}
}
\hfill
\subfigure[Trial 03]{
\includegraphics[width=0.180\textwidth]{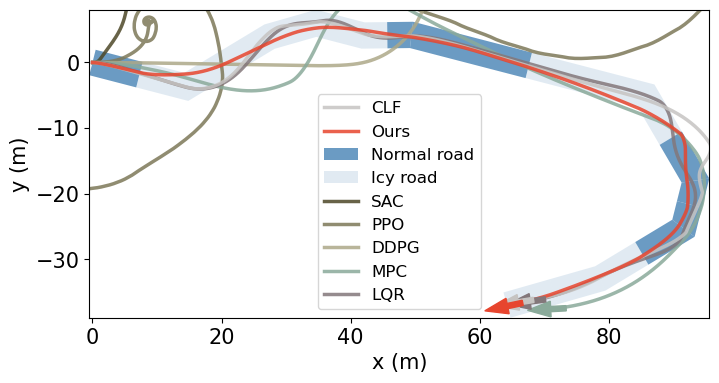}
}
\hfill
\subfigure[Trial 04]{
\includegraphics[width=0.180\textwidth]{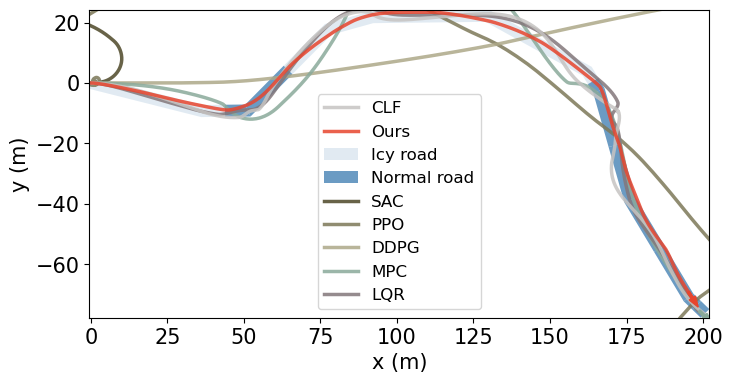}
}
\hfill
}
\end{figure}
\begin{figure}[!htbp]
\floatconts{fig:supple-car-sim-09}
{\caption{Car simulation comparisons}}
{
\subfigure[Trial 05]{
\includegraphics[width=0.180\textwidth]{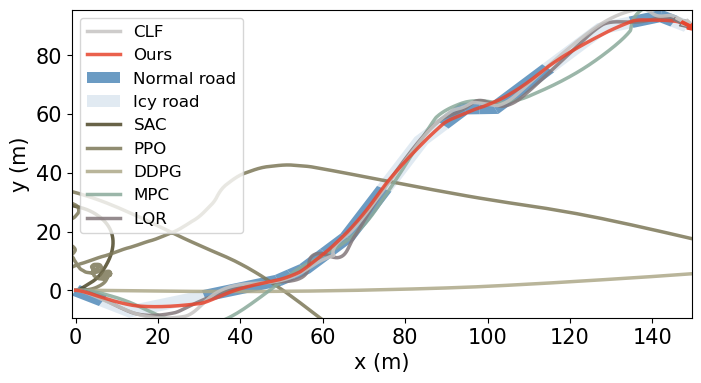}
}
\hfill
\subfigure[Trial 06]{
\includegraphics[width=0.180\textwidth]{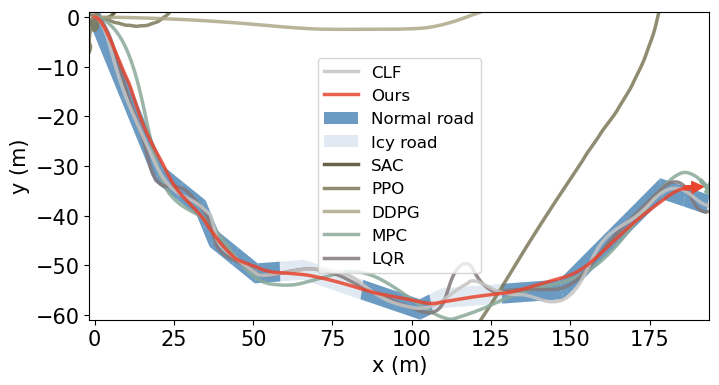}
}
\hfill
\subfigure[Trial 07]{
\includegraphics[width=0.180\textwidth]{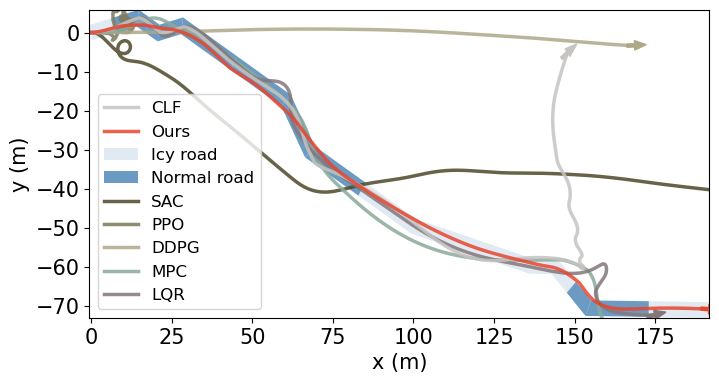}
}
\hfill
\subfigure[Trial 08]{
\includegraphics[width=0.180\textwidth]{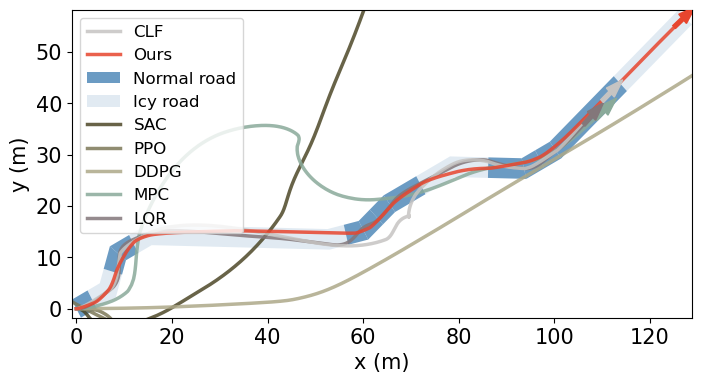}
}
\hfill
\subfigure[Trial 09]{
\includegraphics[width=0.180\textwidth]{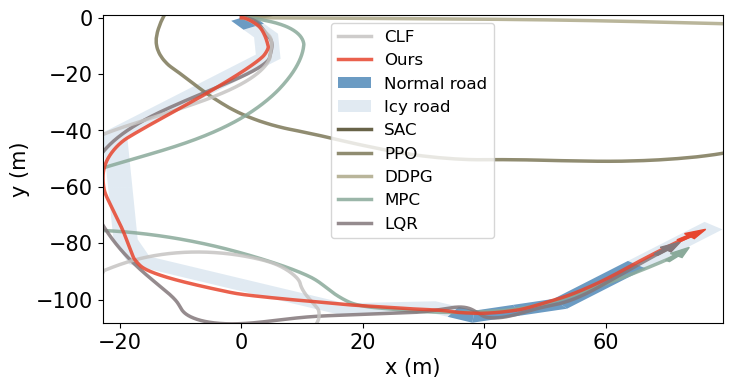}
}
\hfill
}
\end{figure}
\begin{figure}[!htbp]
\floatconts{fig:supple-car-sim-14}
{\caption{Car simulation comparisons}}
{
\subfigure[Trial 10]{
\includegraphics[width=0.180\textwidth]{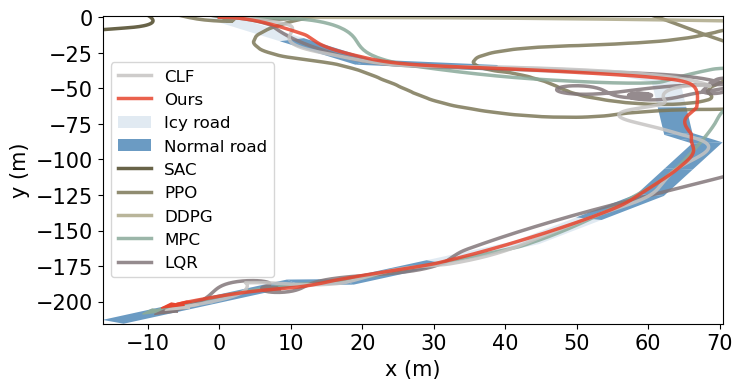}
}
\hfill
\subfigure[Trial 11]{
\includegraphics[width=0.180\textwidth]{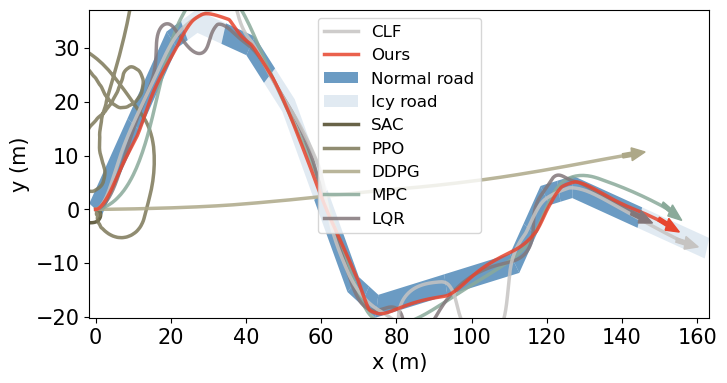}
}
\hfill
\subfigure[Trial 12]{
\includegraphics[width=0.180\textwidth]{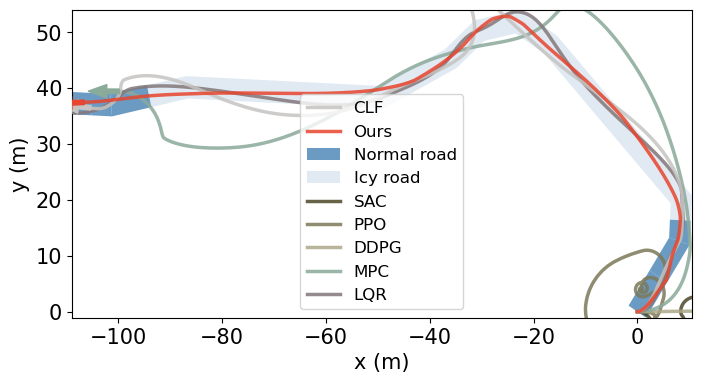}
}
\hfill
\subfigure[Trial 13]{
\includegraphics[width=0.180\textwidth]{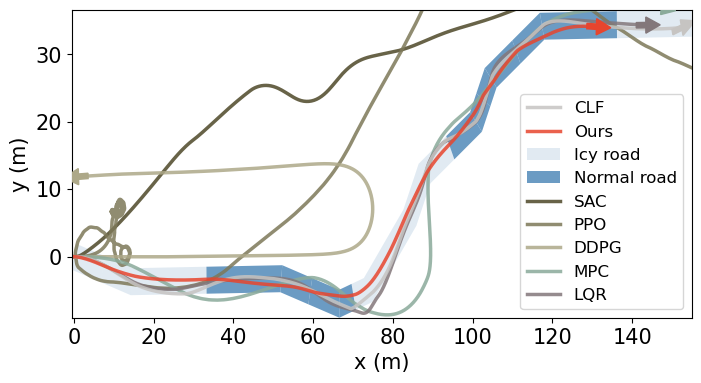}
}
\hfill
\subfigure[Trial 14]{
\includegraphics[width=0.180\textwidth]{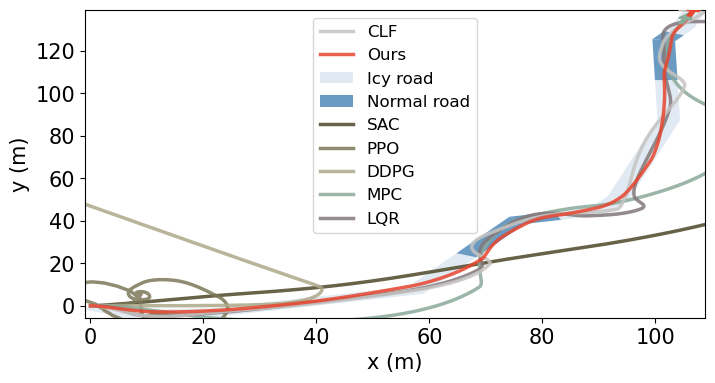}
}
\hfill
}
\end{figure}
\begin{figure}[!htbp]
\floatconts{fig:supple-car-sim-19}
{\caption{Car simulation comparisons}}
{
\subfigure[Trial 15]{
\includegraphics[width=0.180\textwidth]{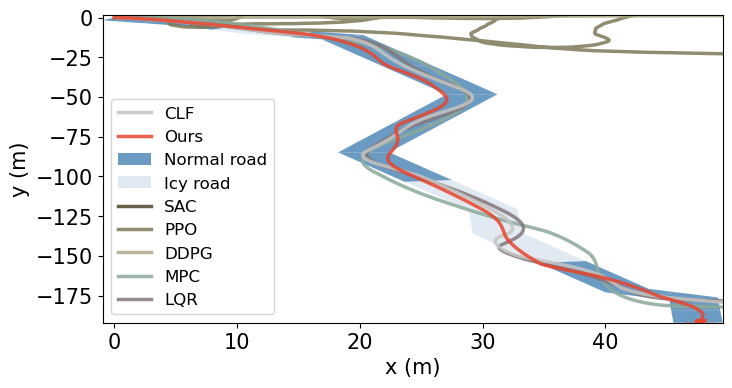}
}
\hfill
\subfigure[Trial 16]{
\includegraphics[width=0.180\textwidth]{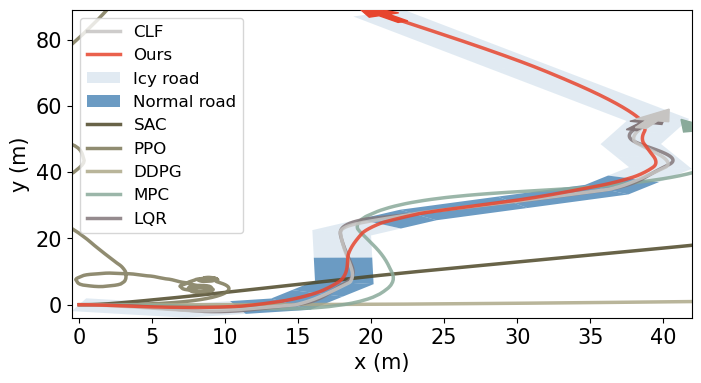}
}
\hfill
\subfigure[Trial 17]{
\includegraphics[width=0.180\textwidth]{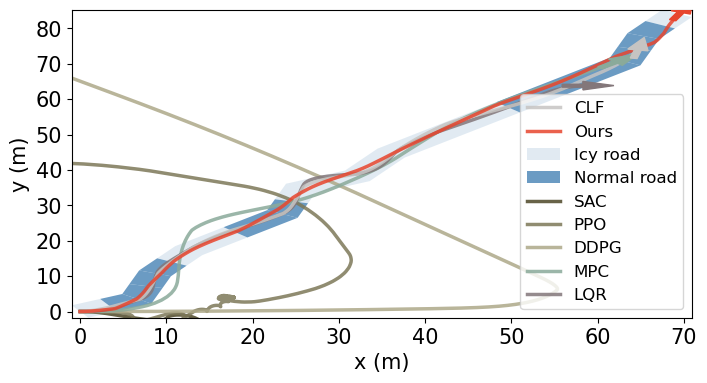}
}
\hfill
\subfigure[Trial 18]{
\includegraphics[width=0.180\textwidth]{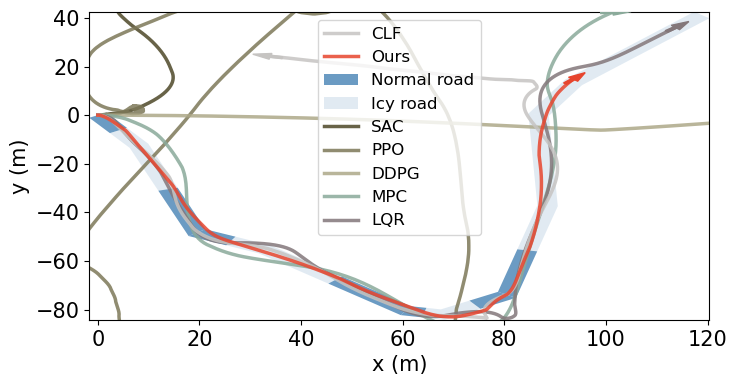}
}
\hfill
\subfigure[Trial 19]{
\includegraphics[width=0.180\textwidth]{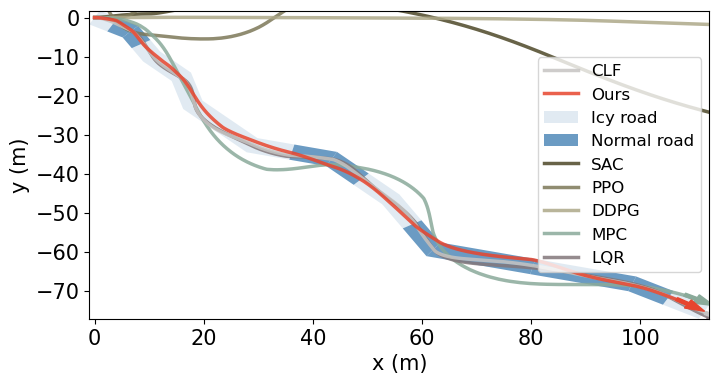}
}
\hfill
}
\end{figure}
\begin{figure}[!htbp]
\floatconts{fig:supple-car-sim-24}
{\caption{Car simulation comparisons}}
{
\subfigure[Trial 20]{
\includegraphics[width=0.180\textwidth]{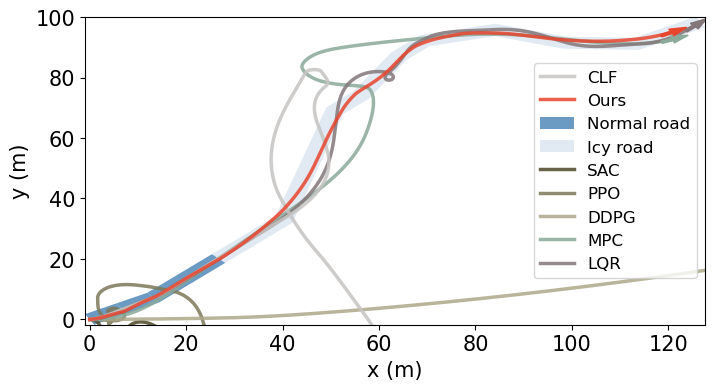}
}
\hfill
\subfigure[Trial 21]{
\includegraphics[width=0.180\textwidth]{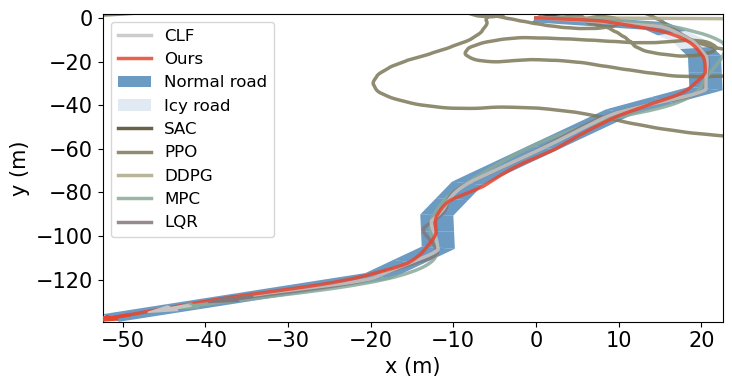}
}
\hfill
\subfigure[Trial 22]{
\includegraphics[width=0.180\textwidth]{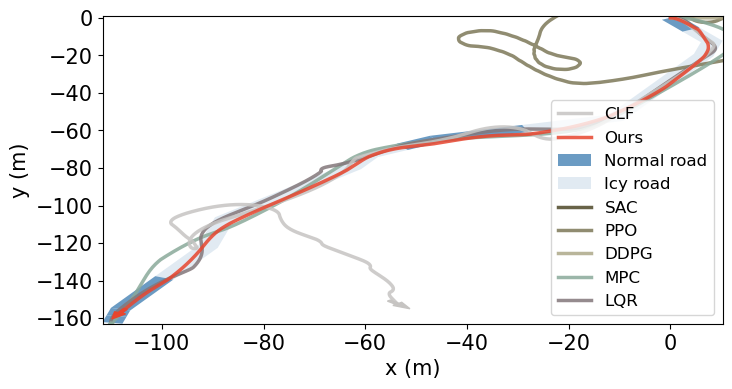}
}
\hfill
\subfigure[Trial 23]{
\includegraphics[width=0.180\textwidth]{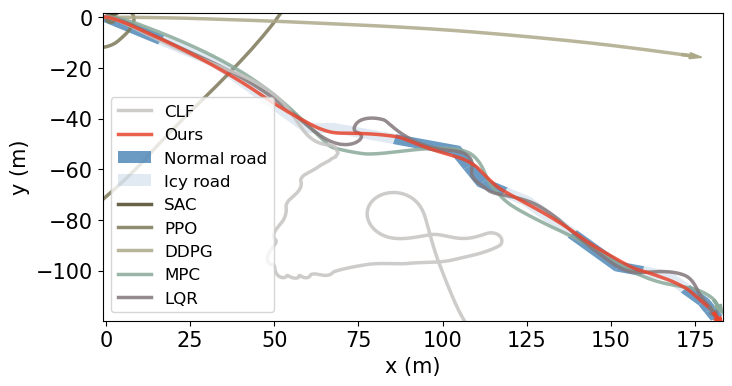}
}
\hfill
\subfigure[Trial 24]{
\includegraphics[width=0.180\textwidth]{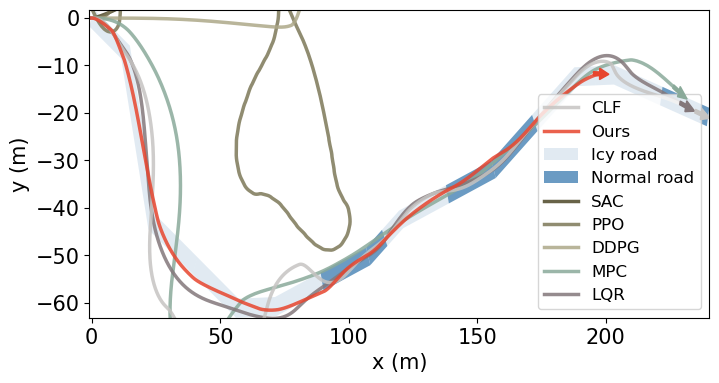}
}
\hfill
}
\end{figure}

\begin{figure}[!htbp]
\floatconts{fig:supple-pogo-sim-04}
{\caption{Pogobot simulation comparisons (trial 04)}}
{
\subfigure[RL-SAC]{
\includegraphics[width=0.180\textwidth]{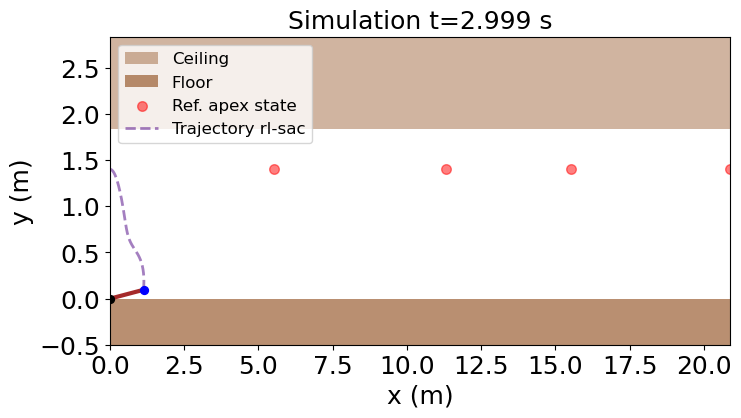}
}
\hfill
\subfigure[RL-PPO]{
\includegraphics[width=0.180\textwidth]{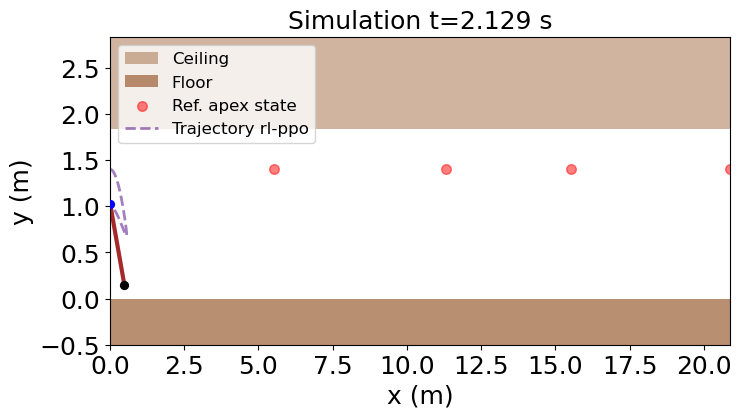}
}
\hfill
\subfigure[RL-DDPG]{
\includegraphics[width=0.180\textwidth]{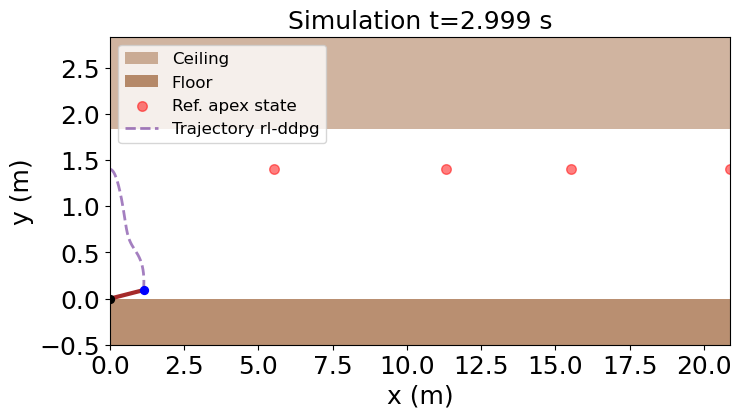}
}
\hfill
\subfigure[MPC]{
\includegraphics[width=0.180\textwidth]{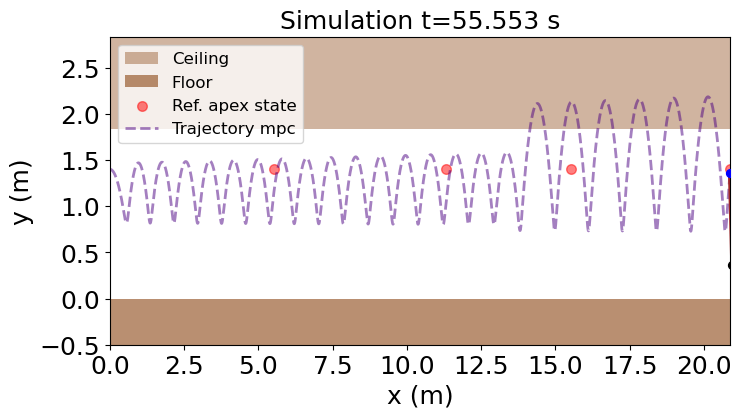}
}
\hfill
\subfigure[Ours]{
\includegraphics[width=0.180\textwidth]{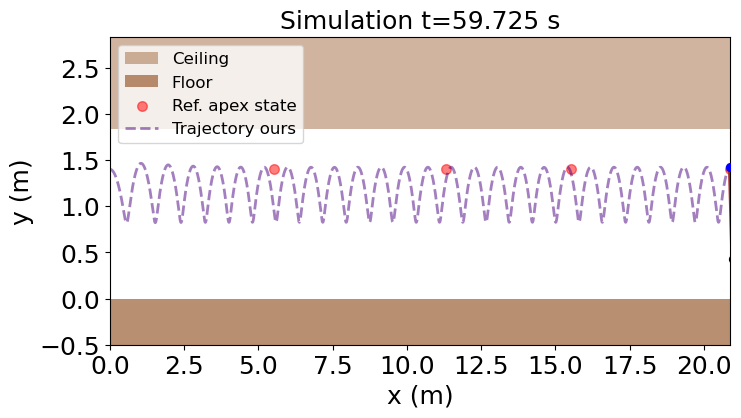}
}
\hfill
}
\end{figure}
\begin{figure}[!htbp]
\floatconts{fig:supple-pogo-sim-05}
{\caption{Pogobot simulation comparisons (trial 05)}}
{
\subfigure[RL-SAC]{
\includegraphics[width=0.180\textwidth]{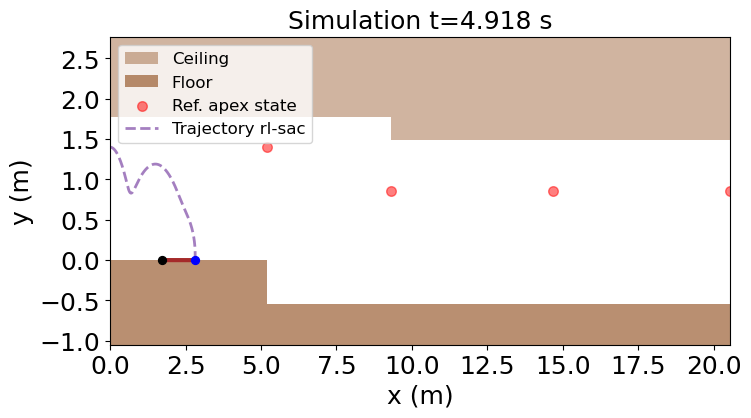}
}
\hfill
\subfigure[RL-PPO]{
\includegraphics[width=0.180\textwidth]{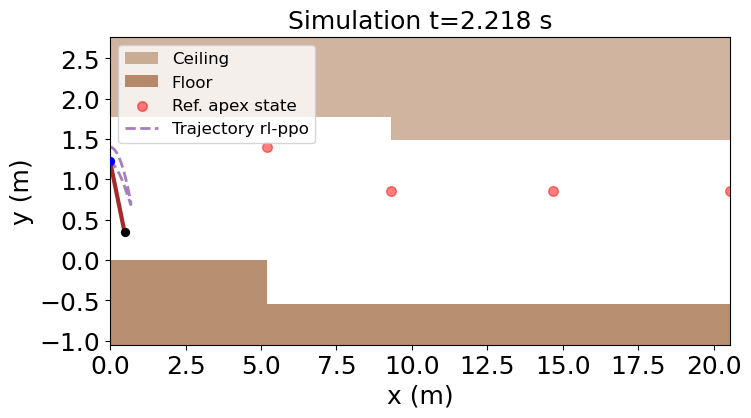}
}
\hfill
\subfigure[RL-DDPG]{
\includegraphics[width=0.180\textwidth]{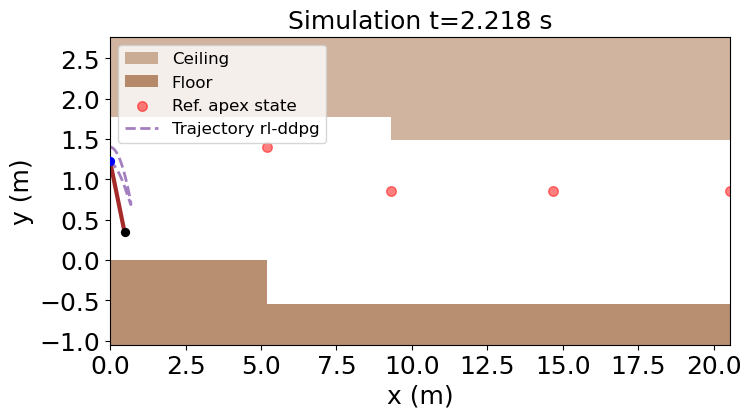}
}
\hfill
\subfigure[MPC]{
\includegraphics[width=0.180\textwidth]{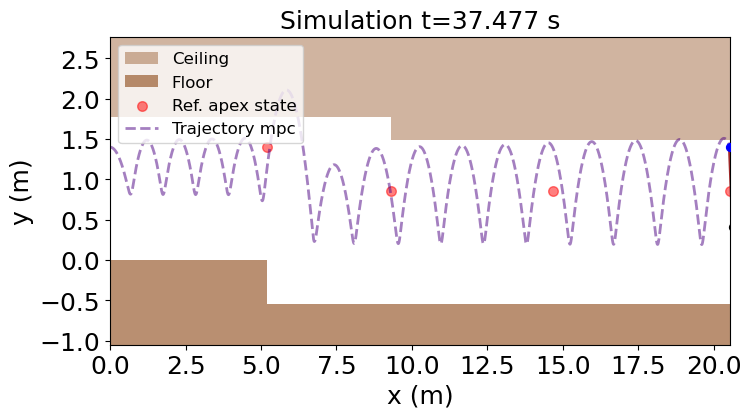}
}
\hfill
\subfigure[Ours]{
\includegraphics[width=0.180\textwidth]{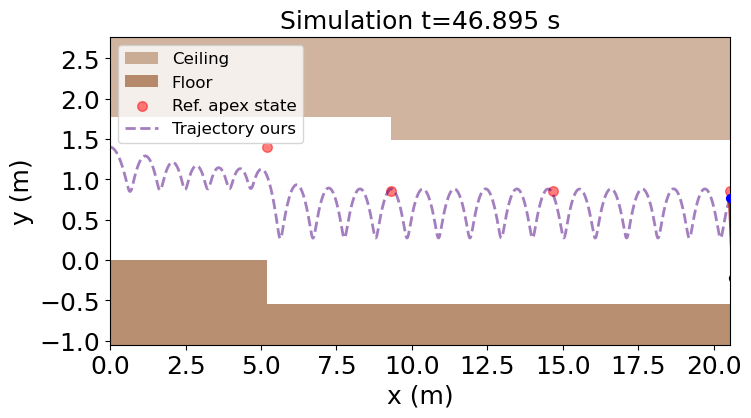}
}
\hfill
}
\end{figure}
\begin{figure}[!htbp]
\floatconts{fig:supple-pogo-sim-06}
{\caption{Pogobot simulation comparisons (trial 06)}}
{
\subfigure[RL-SAC]{
\includegraphics[width=0.180\textwidth]{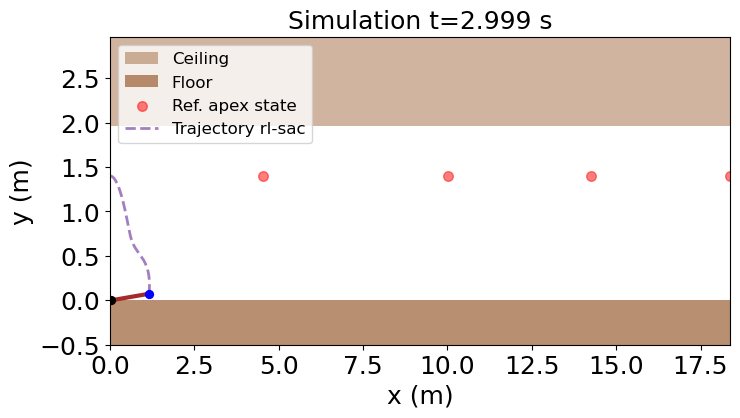}
}
\hfill
\subfigure[RL-PPO]{
\includegraphics[width=0.180\textwidth]{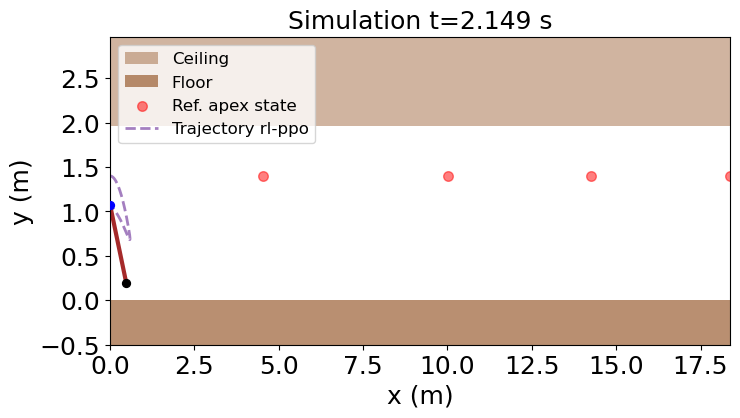}
}
\hfill
\subfigure[RL-DDPG]{
\includegraphics[width=0.180\textwidth]{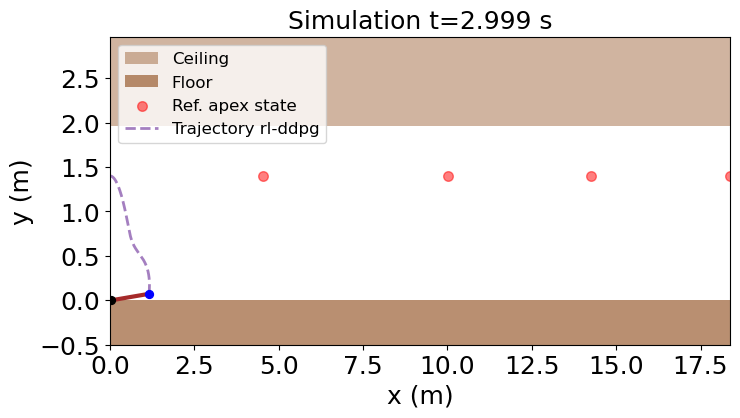}
}
\hfill
\subfigure[MPC]{
\includegraphics[width=0.180\textwidth]{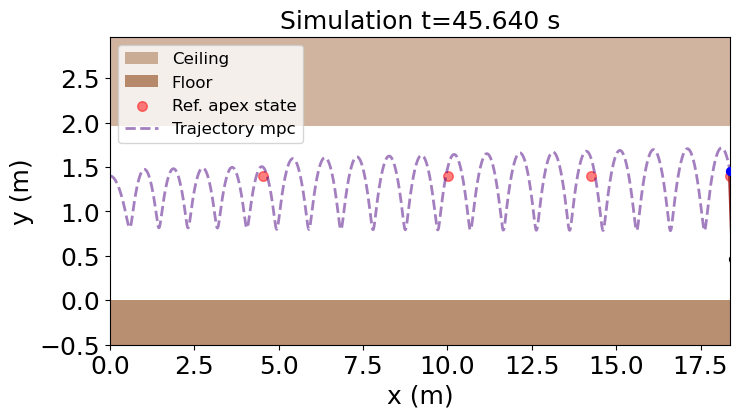}
}
\hfill
\subfigure[Ours]{
\includegraphics[width=0.180\textwidth]{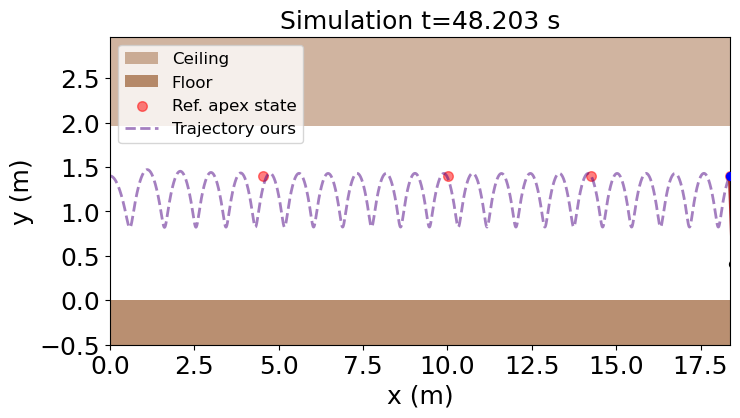}
}
\hfill
}
\end{figure}
\begin{figure}[!htbp]
\floatconts{fig:supple-pogo-sim-07}
{\caption{Pogobot simulation comparisons (trial 07)}}
{
\subfigure[RL-SAC]{
\includegraphics[width=0.180\textwidth]{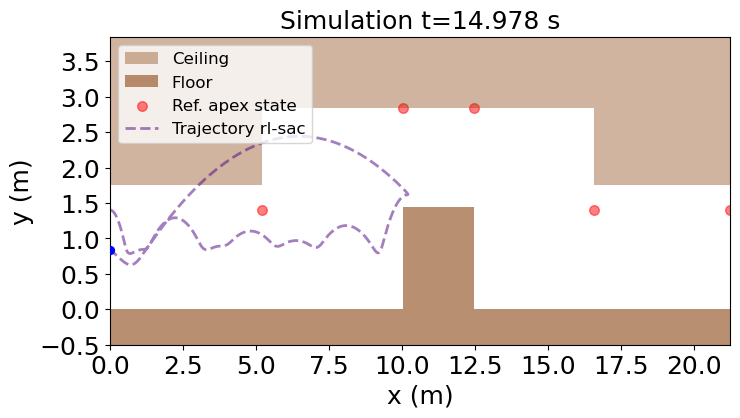}
}
\hfill
\subfigure[RL-PPO]{
\includegraphics[width=0.180\textwidth]{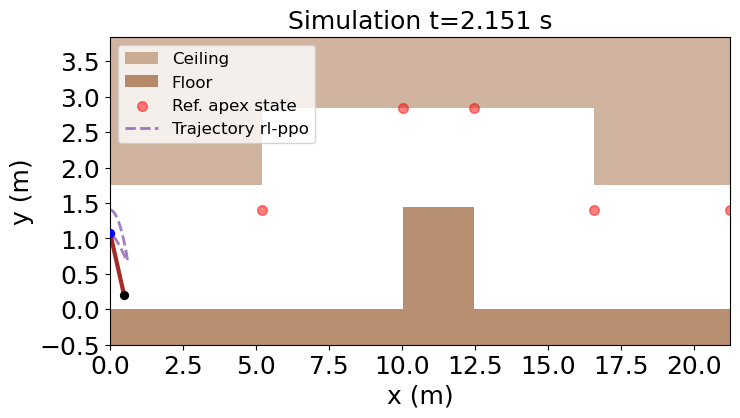}
}
\hfill
\subfigure[RL-DDPG]{
\includegraphics[width=0.180\textwidth]{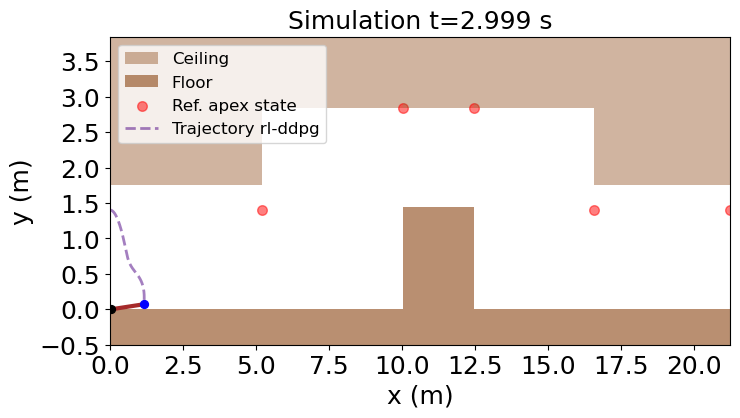}
}
\hfill
\subfigure[MPC]{
\includegraphics[width=0.180\textwidth]{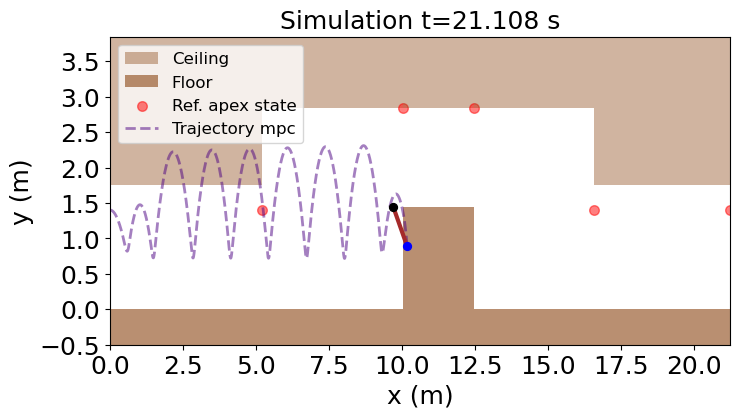}
}
\hfill
\subfigure[Ours]{
\includegraphics[width=0.180\textwidth]{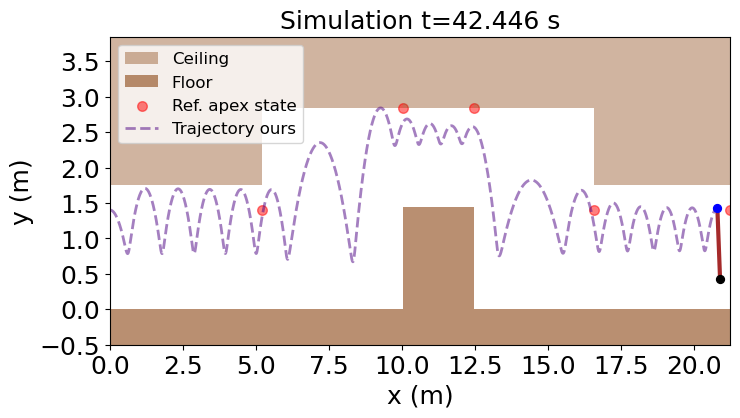}
}
\hfill
}
\end{figure}
\begin{figure}[!htbp]
\floatconts{fig:supple-pogo-sim-08}
{\caption{Pogobot simulation comparisons (trial 08)}}
{
\subfigure[RL-SAC]{
\includegraphics[width=0.180\textwidth]{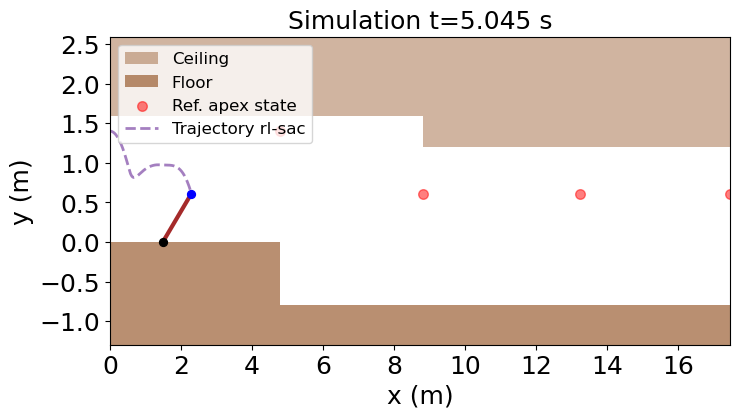}
}
\hfill
\subfigure[RL-PPO]{
\includegraphics[width=0.180\textwidth]{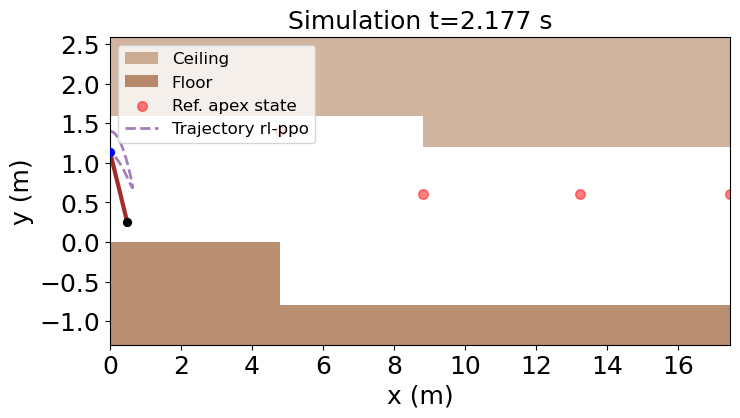}
}
\hfill
\subfigure[RL-DDPG]{
\includegraphics[width=0.180\textwidth]{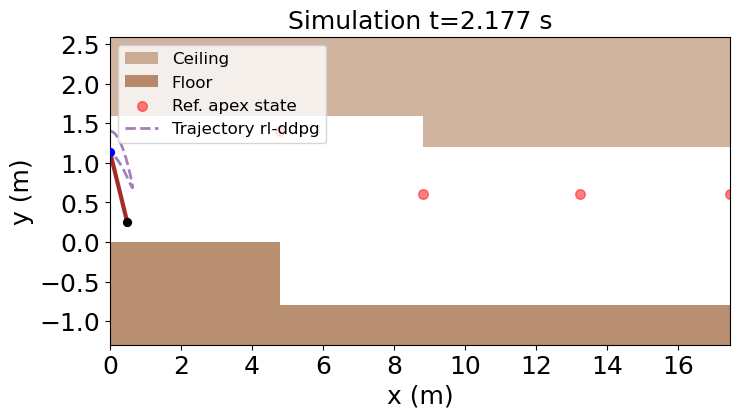}
}
\hfill
\subfigure[MPC]{
\includegraphics[width=0.180\textwidth]{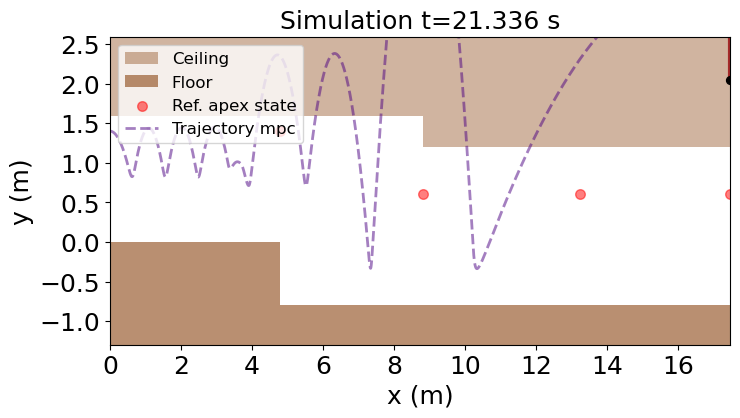}
}
\hfill
\subfigure[Ours]{
\includegraphics[width=0.180\textwidth]{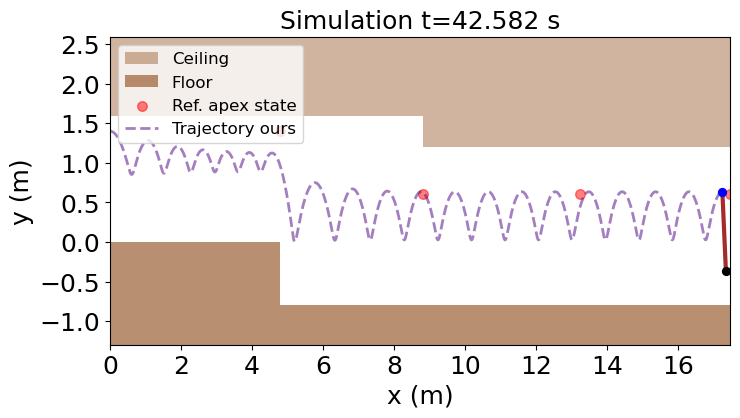}
}
\hfill
}
\end{figure}
\begin{figure}[!htbp]
\floatconts{fig:supple-pogo-sim-09}
{\caption{Pogobot simulation comparisons (trial 09)}}
{
\subfigure[RL-SAC]{
\includegraphics[width=0.180\textwidth]{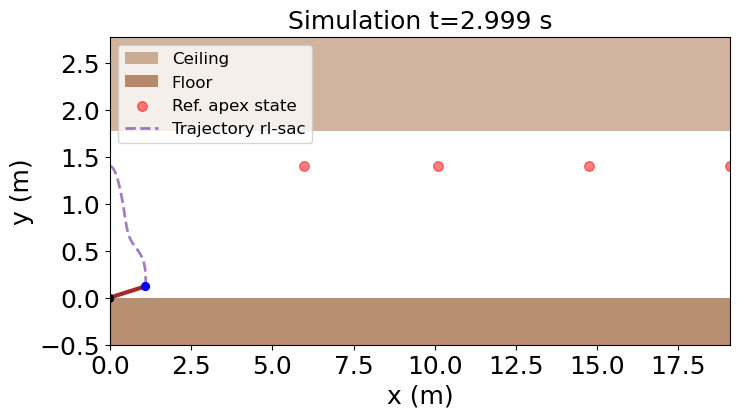}
}
\hfill
\subfigure[RL-PPO]{
\includegraphics[width=0.180\textwidth]{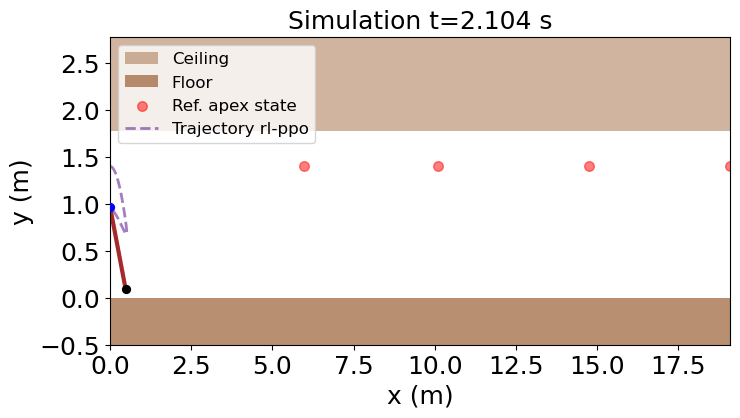}
}
\hfill
\subfigure[RL-DDPG]{
\includegraphics[width=0.180\textwidth]{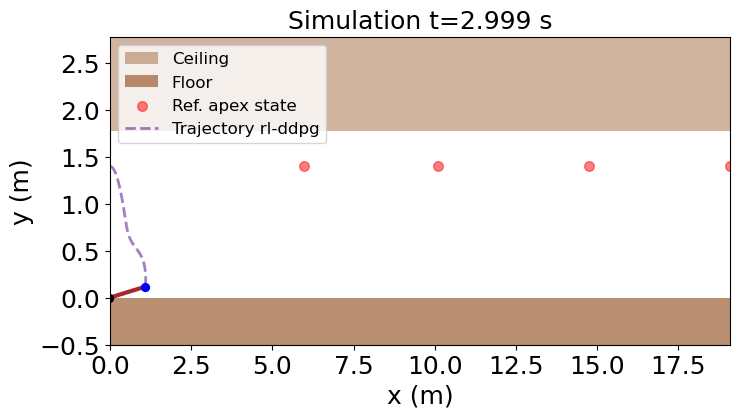}
}
\hfill
\subfigure[MPC]{
\includegraphics[width=0.180\textwidth]{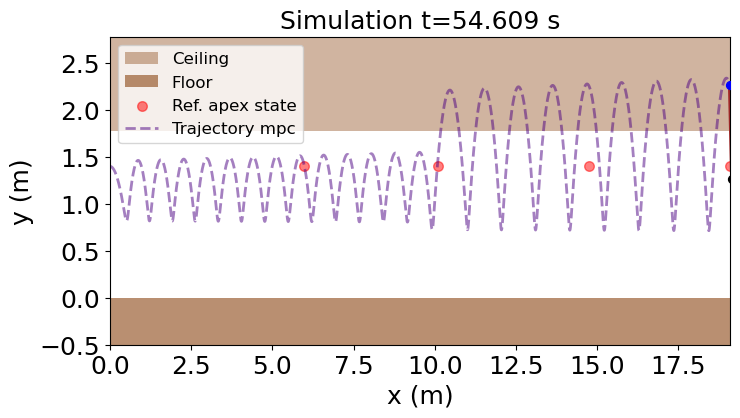}
}
\hfill
\subfigure[Ours]{
\includegraphics[width=0.180\textwidth]{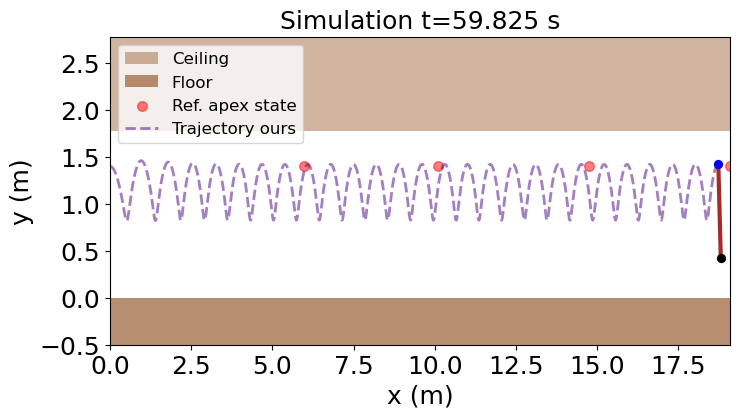}
}
\hfill
}
\end{figure}
\begin{figure}[!htbp]
\floatconts{fig:supple-pogo-sim-10}
{\caption{Pogobot simulation comparisons (trial 10)}}
{
\subfigure[RL-SAC]{
\includegraphics[width=0.180\textwidth]{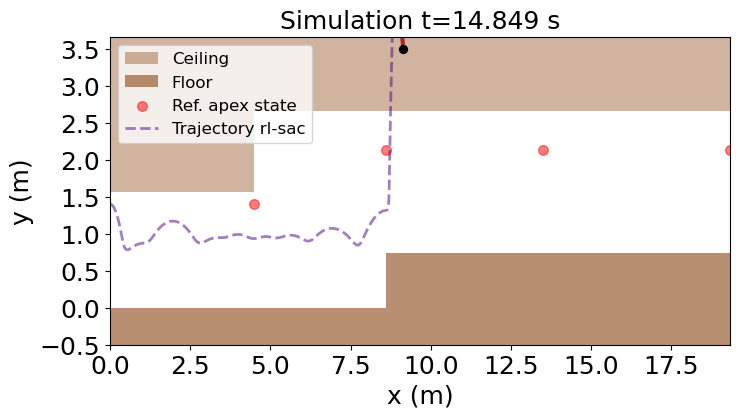}
}
\hfill
\subfigure[RL-PPO]{
\includegraphics[width=0.180\textwidth]{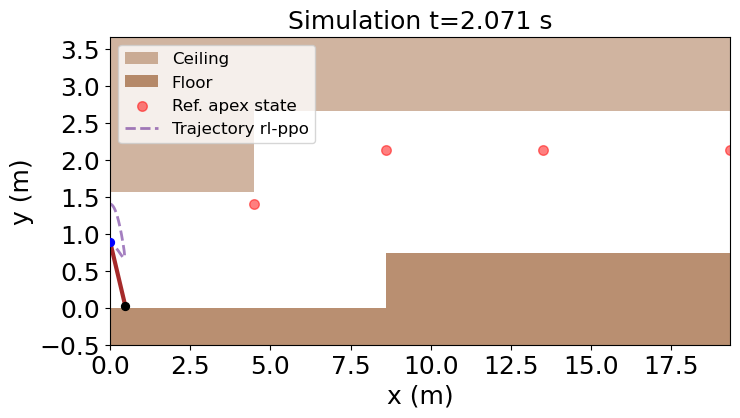}
}
\hfill
\subfigure[RL-DDPG]{
\includegraphics[width=0.180\textwidth]{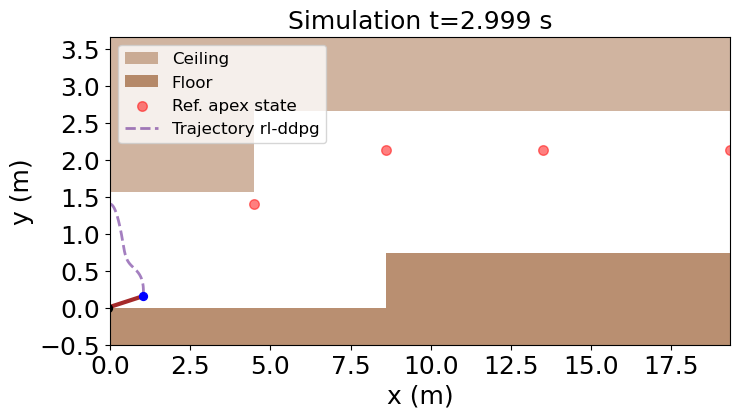}
}
\hfill
\subfigure[MPC]{
\includegraphics[width=0.180\textwidth]{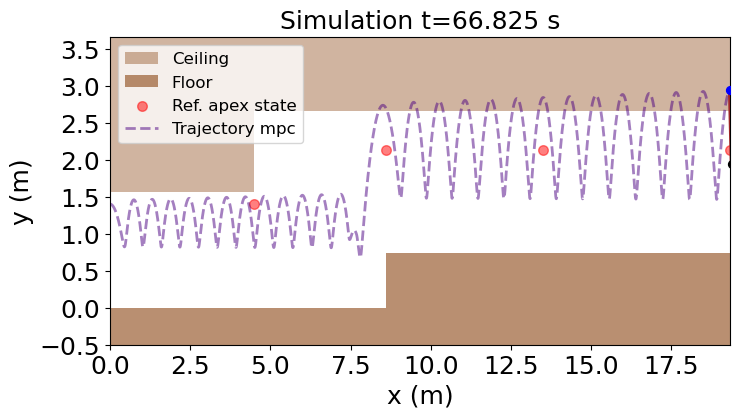}
}
\hfill
\subfigure[Ours]{
\includegraphics[width=0.180\textwidth]{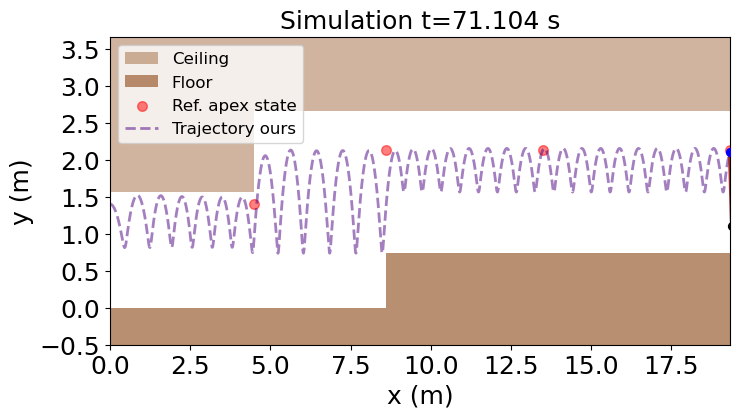}
}
\hfill
}
\end{figure}
\begin{figure}[!htbp]
\floatconts{fig:supple-pogo-sim-11}
{\caption{Pogobot simulation comparisons (trial 11)}}
{
\subfigure[RL-SAC]{
\includegraphics[width=0.180\textwidth]{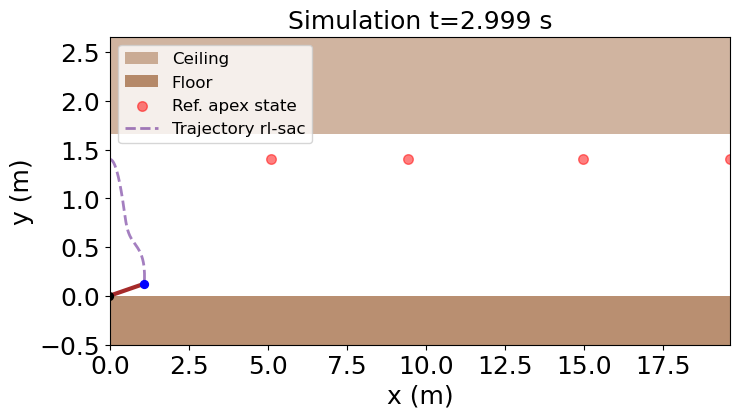}
}
\hfill
\subfigure[RL-PPO]{
\includegraphics[width=0.180\textwidth]{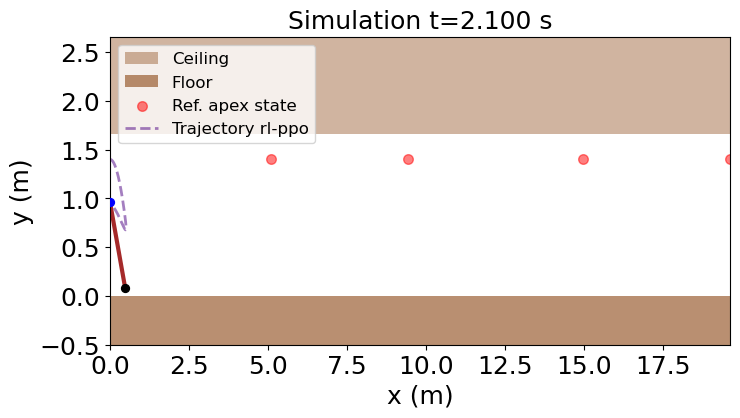}
}
\hfill
\subfigure[RL-DDPG]{
\includegraphics[width=0.180\textwidth]{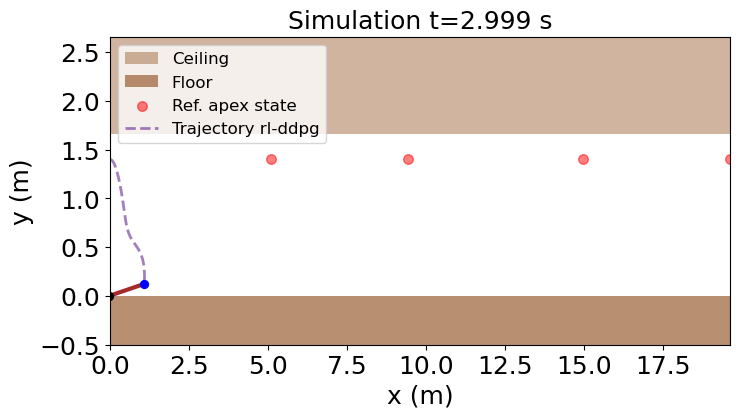}
}
\hfill
\subfigure[MPC]{
\includegraphics[width=0.180\textwidth]{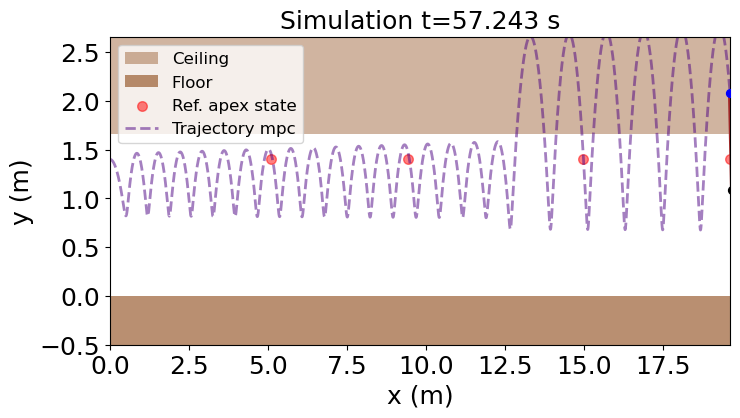}
}
\hfill
\subfigure[Ours]{
\includegraphics[width=0.180\textwidth]{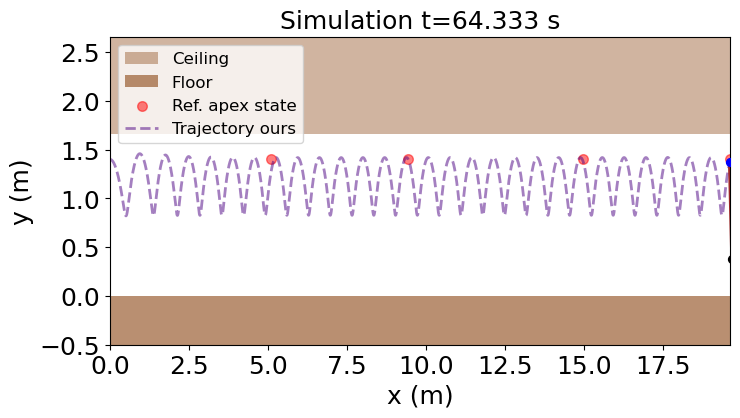}
}
\hfill
}
\end{figure}
\begin{figure}[!htbp]
\floatconts{fig:supple-pogo-sim-12}
{\caption{Pogobot simulation comparisons (trial 12)}}
{
\subfigure[RL-SAC]{
\includegraphics[width=0.180\textwidth]{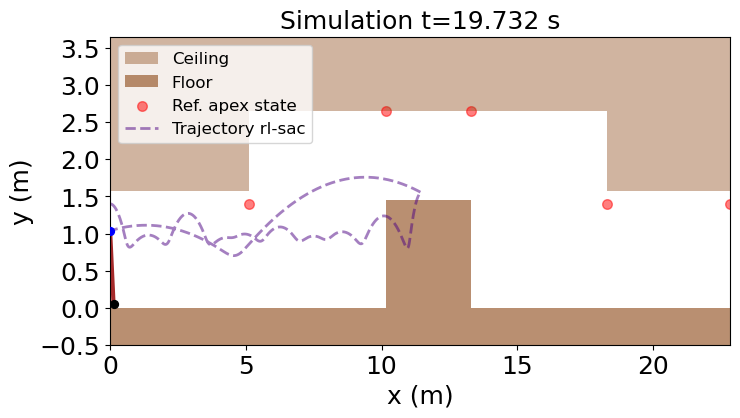}
}
\hfill
\subfigure[RL-PPO]{
\includegraphics[width=0.180\textwidth]{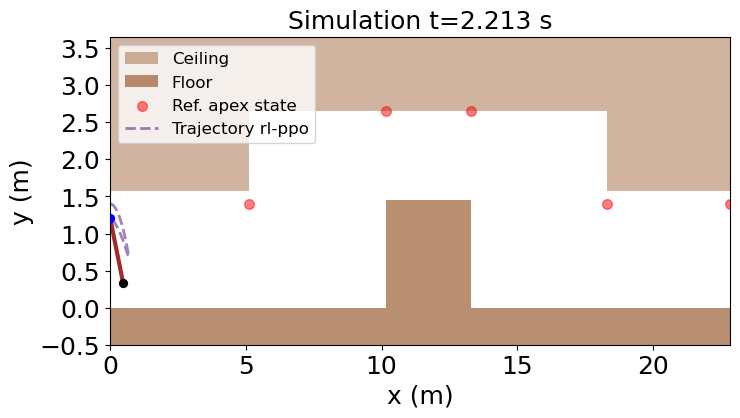}
}
\hfill
\subfigure[RL-DDPG]{
\includegraphics[width=0.180\textwidth]{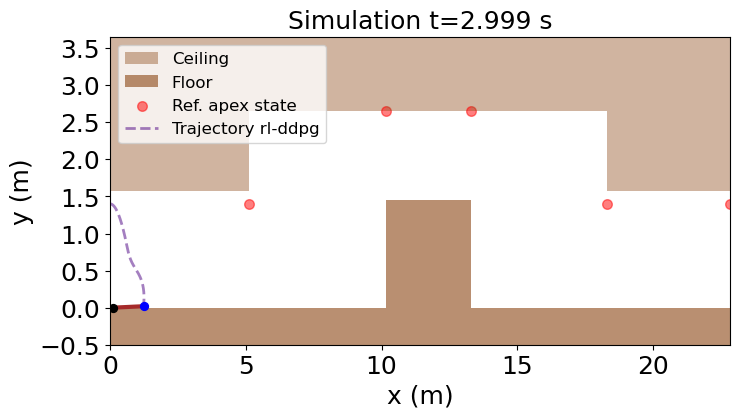}
}
\hfill
\subfigure[MPC]{
\includegraphics[width=0.180\textwidth]{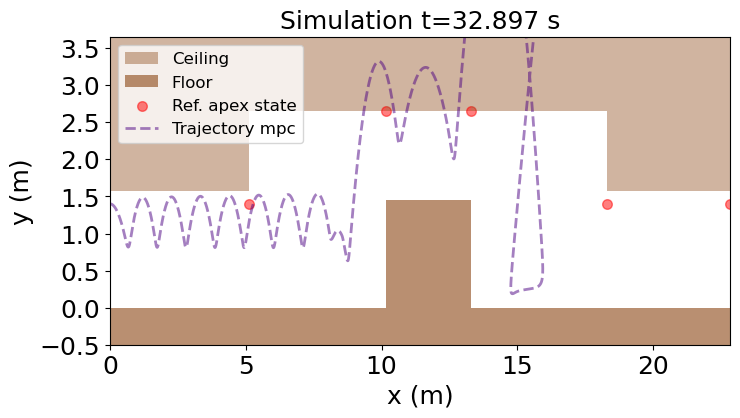}
}
\hfill
\subfigure[Ours]{
\includegraphics[width=0.180\textwidth]{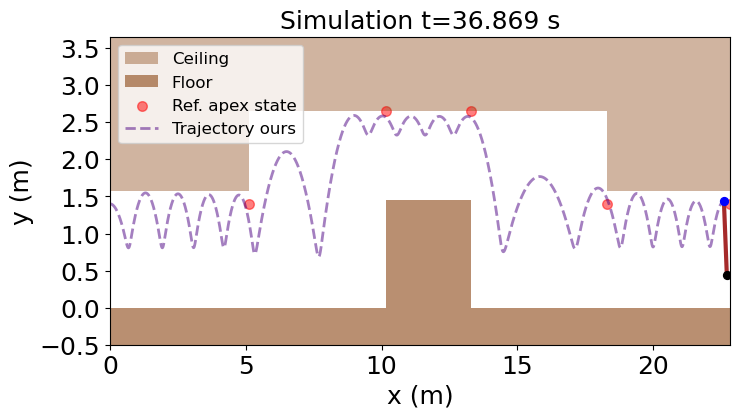}
}
\hfill
}
\end{figure}
\begin{figure}[!htbp]
\floatconts{fig:supple-pogo-sim-13}
{\caption{Pogobot simulation comparisons (trial 13)}}
{
\subfigure[RL-SAC]{
\includegraphics[width=0.180\textwidth]{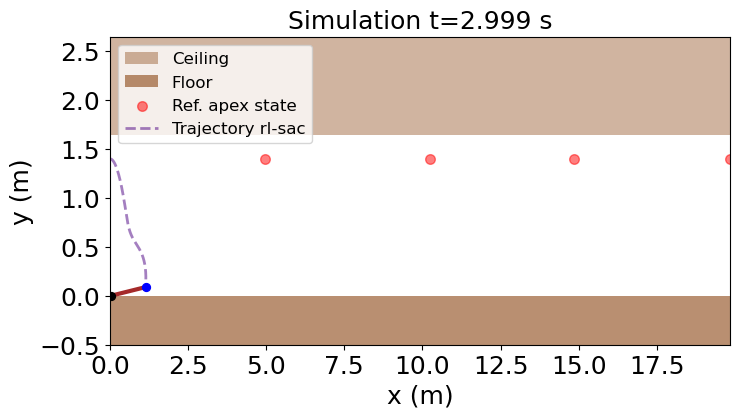}
}
\hfill
\subfigure[RL-PPO]{
\includegraphics[width=0.180\textwidth]{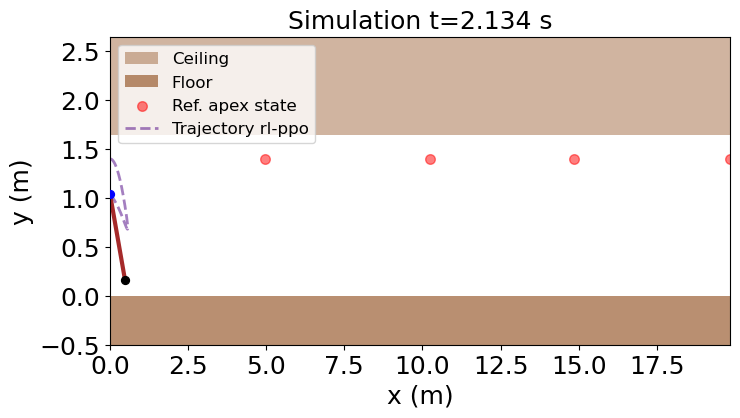}
}
\hfill
\subfigure[RL-DDPG]{
\includegraphics[width=0.180\textwidth]{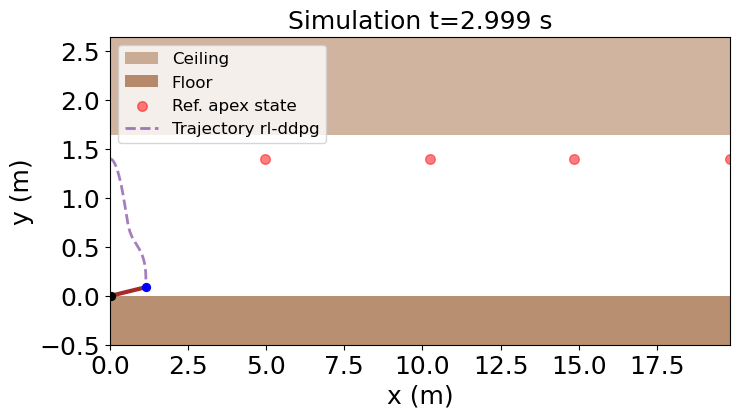}
}
\hfill
\subfigure[MPC]{
\includegraphics[width=0.180\textwidth]{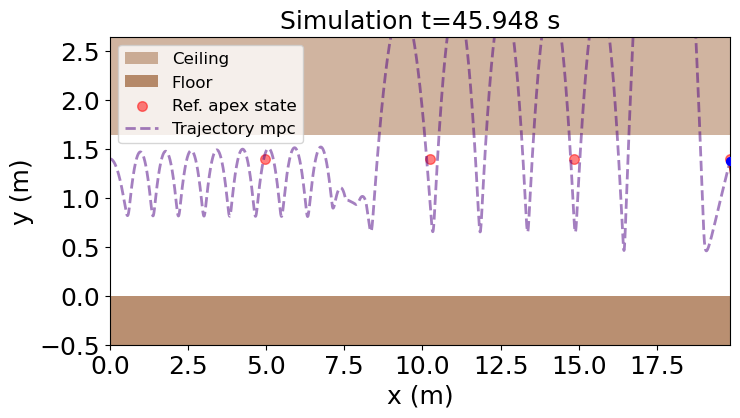}
}
\hfill
\subfigure[Ours]{
\includegraphics[width=0.180\textwidth]{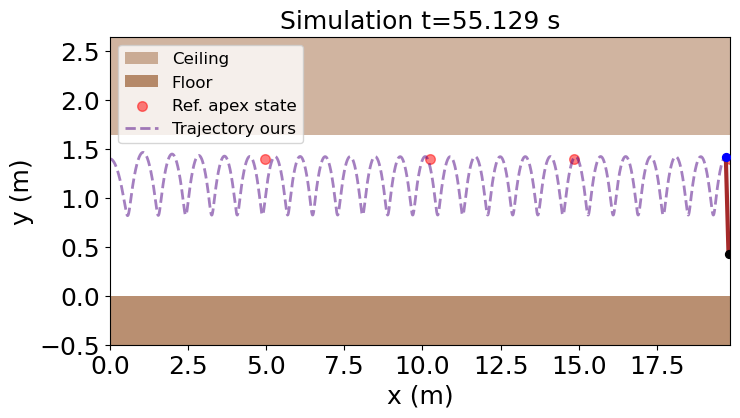}
}
\hfill
}
\end{figure}
\begin{figure}[!htbp]
\floatconts{fig:supple-pogo-sim-14}
{\caption{Pogobot simulation comparisons (trial 14)}}
{
\subfigure[RL-SAC]{
\includegraphics[width=0.180\textwidth]{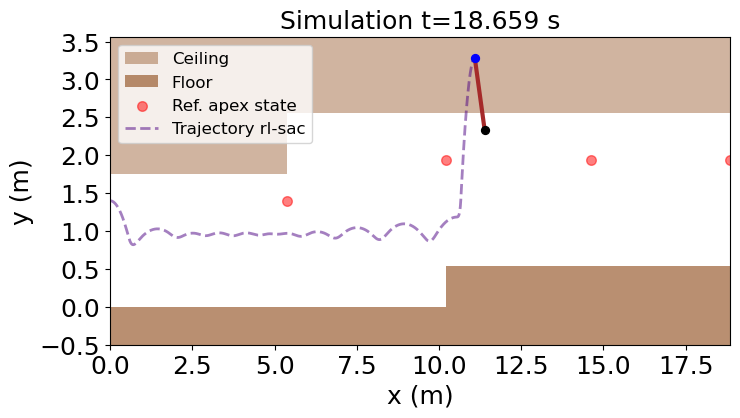}
}
\hfill
\subfigure[RL-PPO]{
\includegraphics[width=0.180\textwidth]{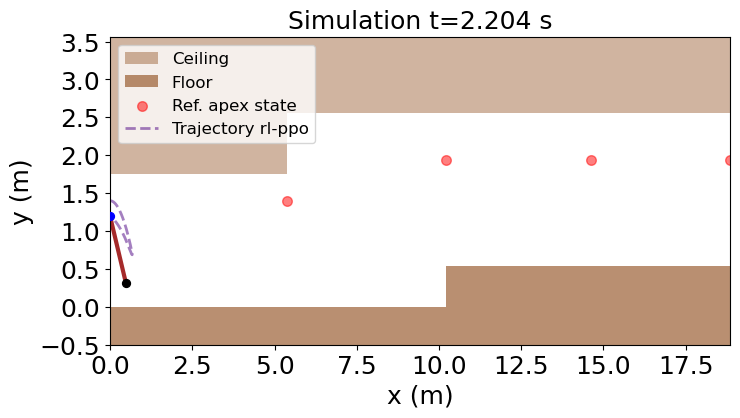}
}
\hfill
\subfigure[RL-DDPG]{
\includegraphics[width=0.180\textwidth]{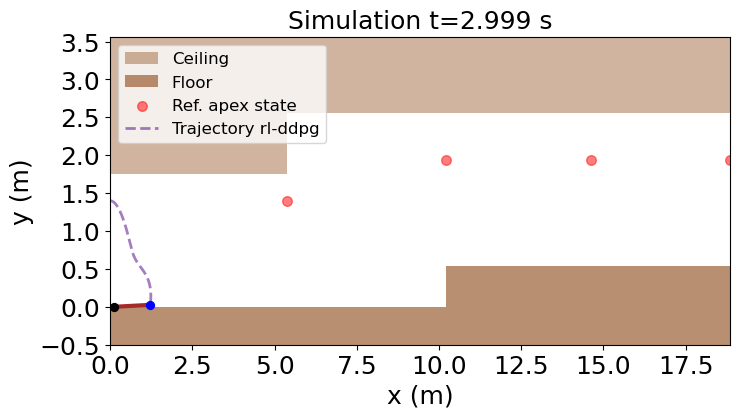}
}
\hfill
\subfigure[MPC]{
\includegraphics[width=0.180\textwidth]{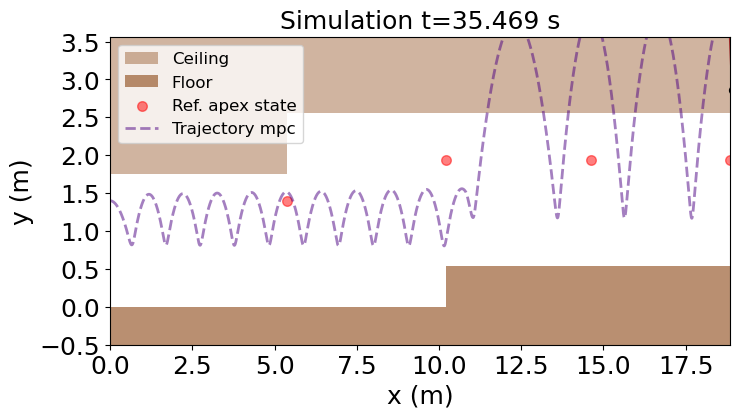}
}
\hfill
\subfigure[Ours]{
\includegraphics[width=0.180\textwidth]{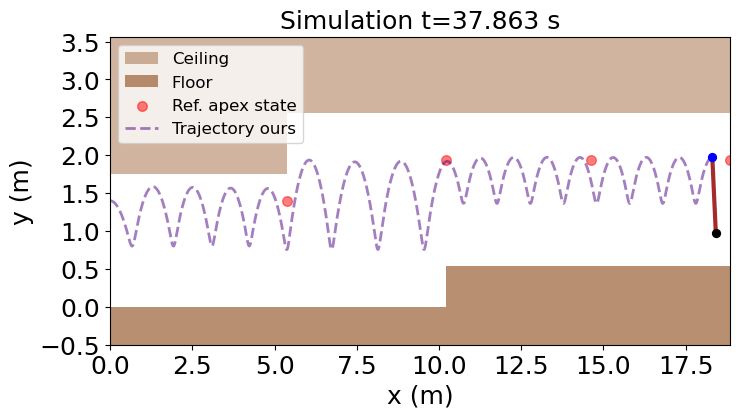}
}
\hfill
}
\end{figure}
\begin{figure}[!htbp]
\floatconts{fig:supple-pogo-sim-15}
{\caption{Pogobot simulation comparisons (trial 15)}}
{
\subfigure[RL-SAC]{
\includegraphics[width=0.180\textwidth]{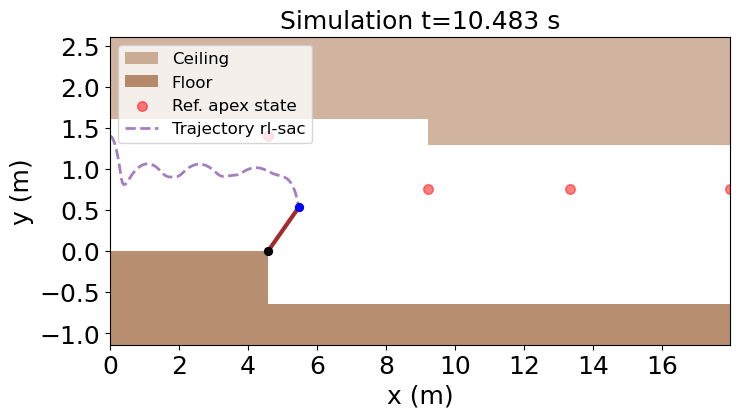}
}
\hfill
\subfigure[RL-PPO]{
\includegraphics[width=0.180\textwidth]{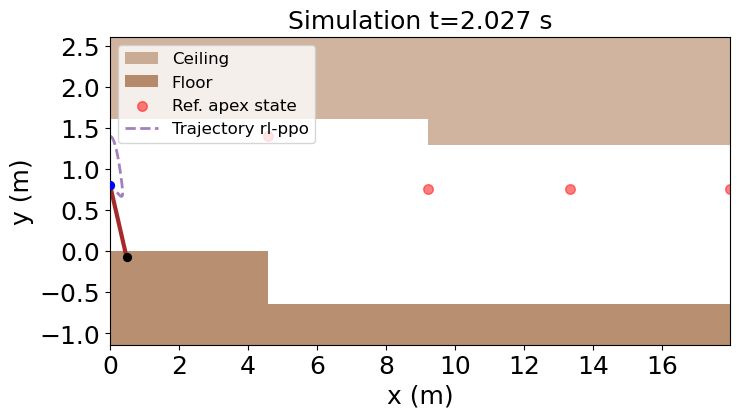}
}
\hfill
\subfigure[RL-DDPG]{
\includegraphics[width=0.180\textwidth]{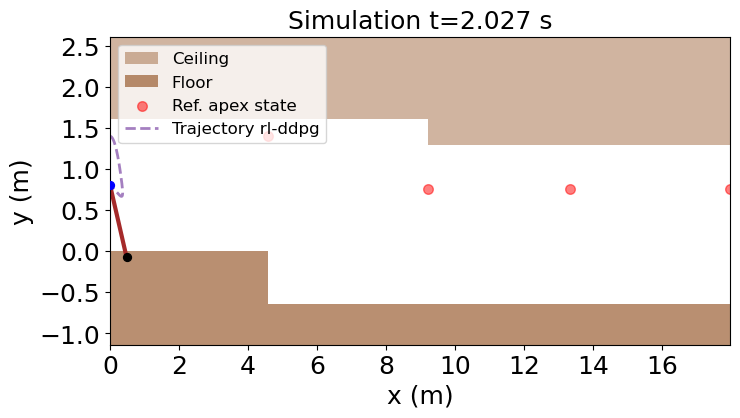}
}
\hfill
\subfigure[MPC]{
\includegraphics[width=0.180\textwidth]{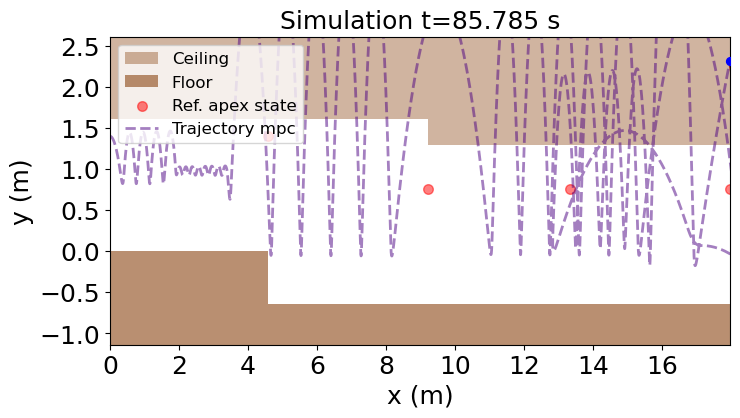}
}
\hfill
\subfigure[Ours]{
\includegraphics[width=0.180\textwidth]{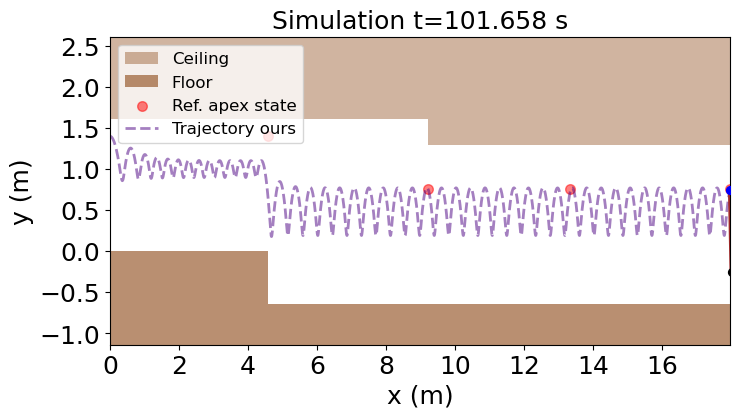}
}
\hfill
}
\end{figure}
\begin{figure}[!htbp]
\floatconts{fig:supple-pogo-sim-16}
{\caption{Pogobot simulation comparisons (trial 16)}}
{
\subfigure[RL-SAC]{
\includegraphics[width=0.180\textwidth]{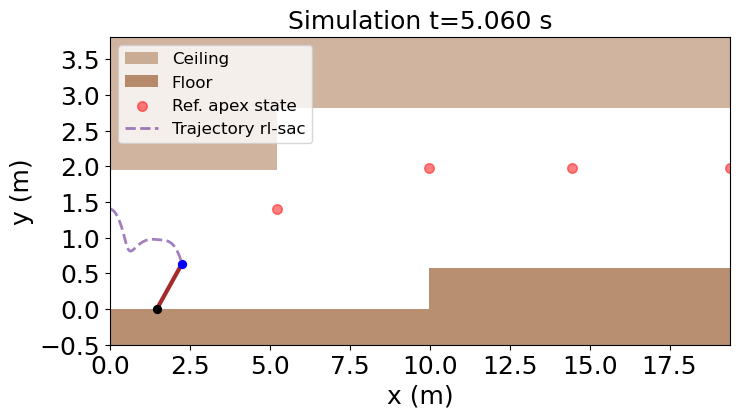}
}
\hfill
\subfigure[RL-PPO]{
\includegraphics[width=0.180\textwidth]{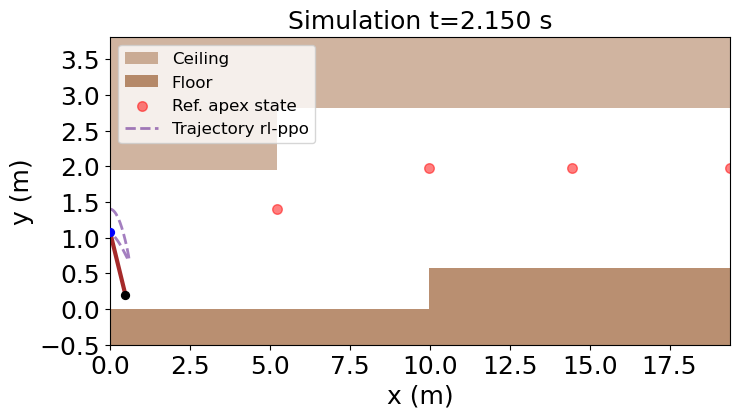}
}
\hfill
\subfigure[RL-DDPG]{
\includegraphics[width=0.180\textwidth]{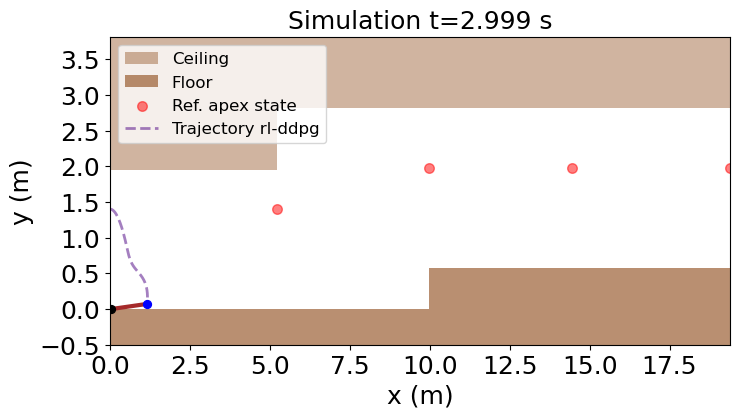}
}
\hfill
\subfigure[MPC]{
\includegraphics[width=0.180\textwidth]{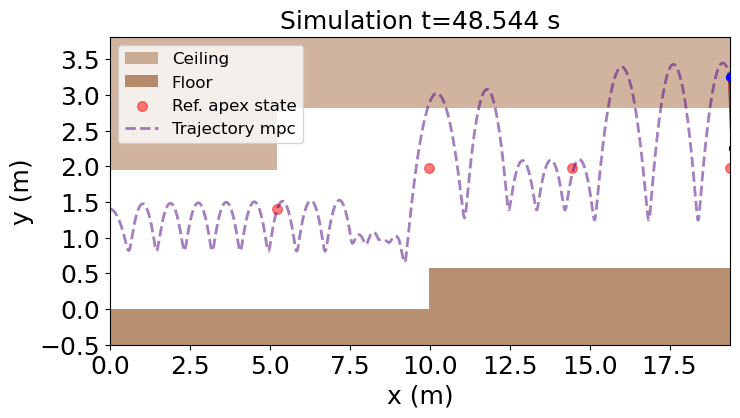}
}
\hfill
\subfigure[Ours]{
\includegraphics[width=0.180\textwidth]{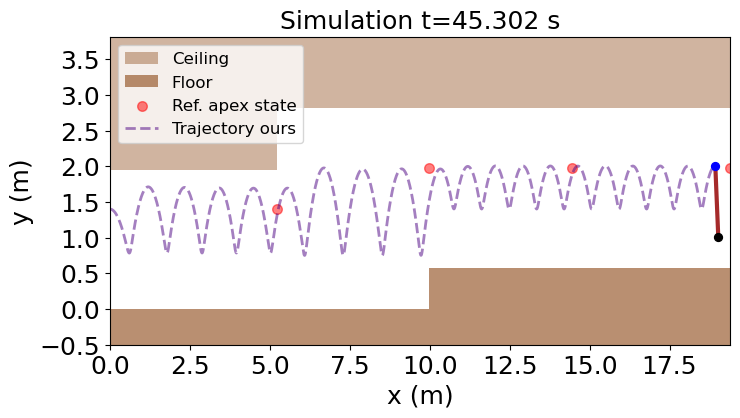}
}
\hfill
}
\end{figure}
\begin{figure}[!htbp]
\floatconts{fig:supple-pogo-sim-17}
{\caption{Pogobot simulation comparisons (trial 17)}}
{
\subfigure[RL-SAC]{
\includegraphics[width=0.180\textwidth]{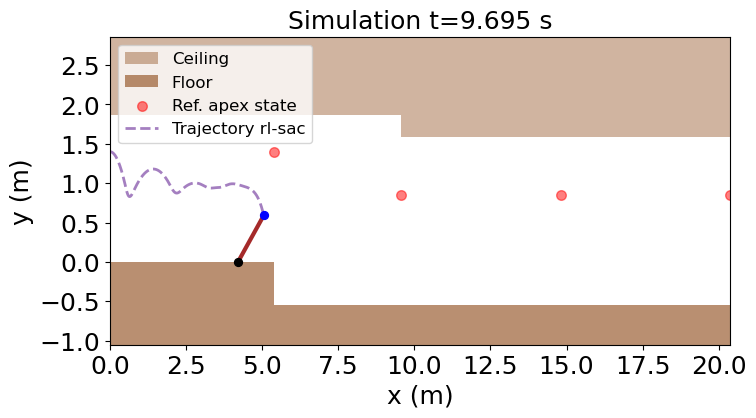}
}
\hfill
\subfigure[RL-PPO]{
\includegraphics[width=0.180\textwidth]{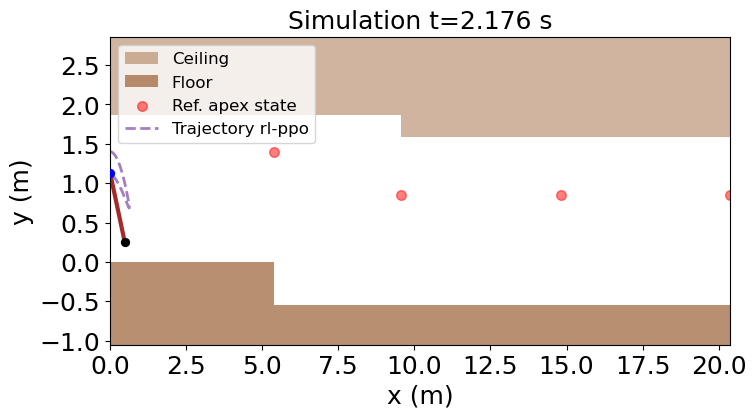}
}
\hfill
\subfigure[RL-DDPG]{
\includegraphics[width=0.180\textwidth]{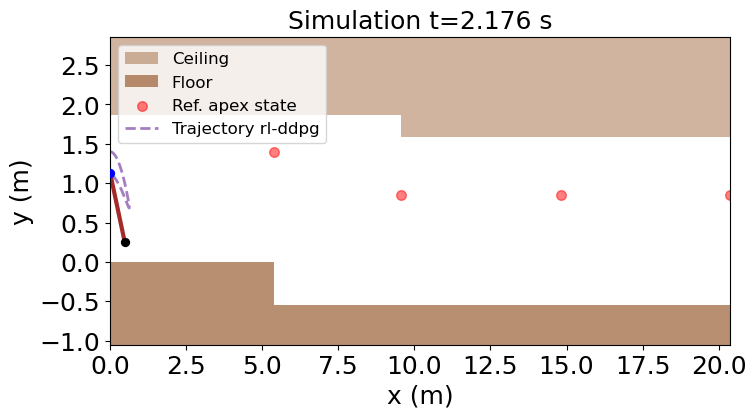}
}
\hfill
\subfigure[MPC]{
\includegraphics[width=0.180\textwidth]{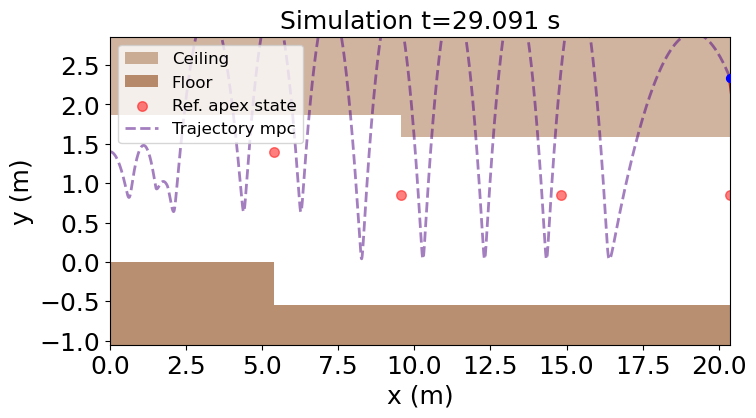}
}
\hfill
\subfigure[Ours]{
\includegraphics[width=0.180\textwidth]{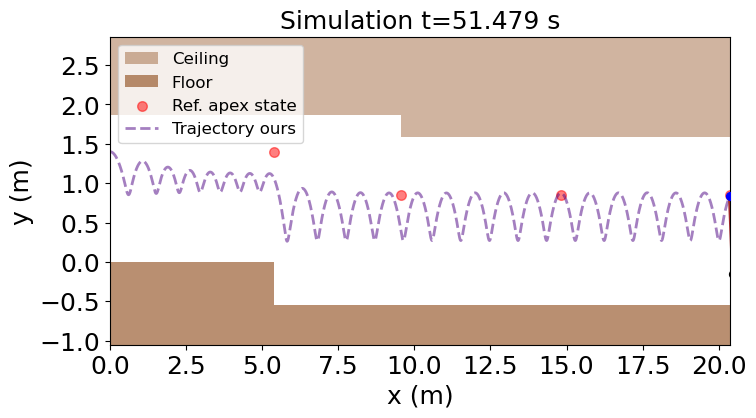}
}
\hfill
}
\end{figure}
\begin{figure}[!htbp]
\floatconts{fig:supple-pogo-sim-18}
{\caption{Pogobot simulation comparisons (trial 18)}}
{
\subfigure[RL-SAC]{
\includegraphics[width=0.180\textwidth]{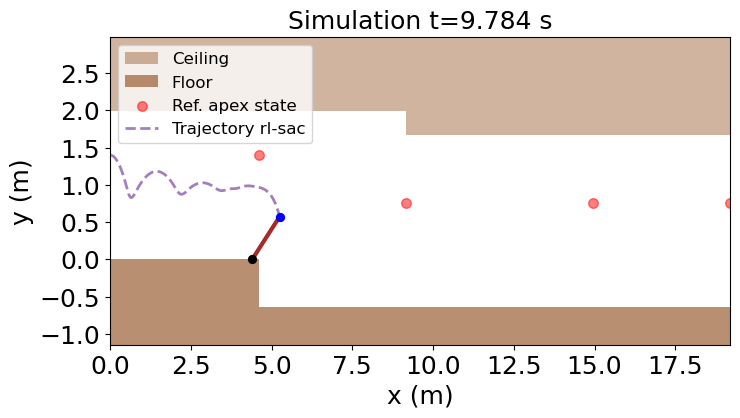}
}
\hfill
\subfigure[RL-PPO]{
\includegraphics[width=0.180\textwidth]{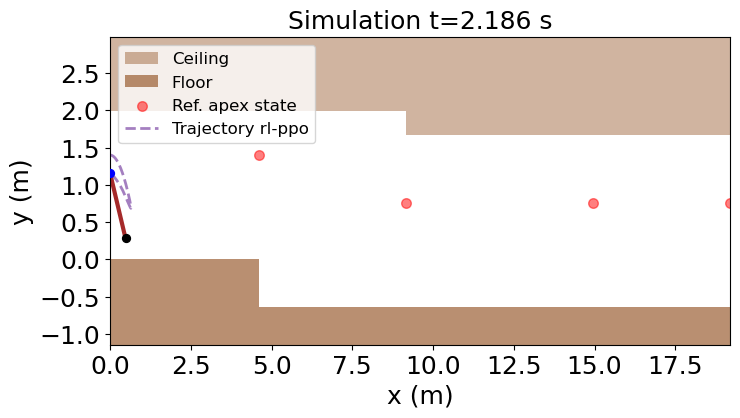}
}
\hfill
\subfigure[RL-DDPG]{
\includegraphics[width=0.180\textwidth]{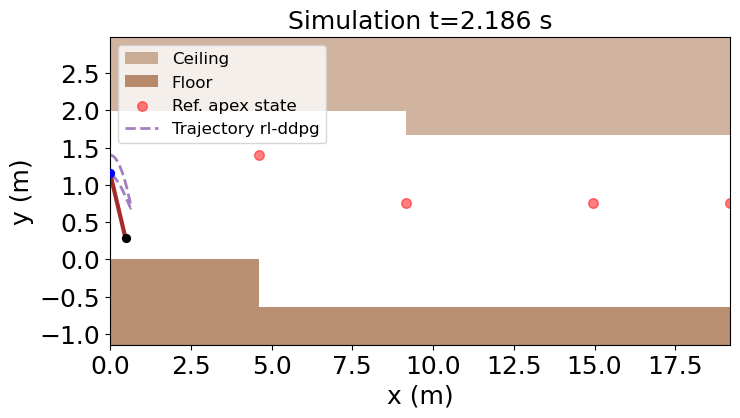}
}
\hfill
\subfigure[MPC]{
\includegraphics[width=0.180\textwidth]{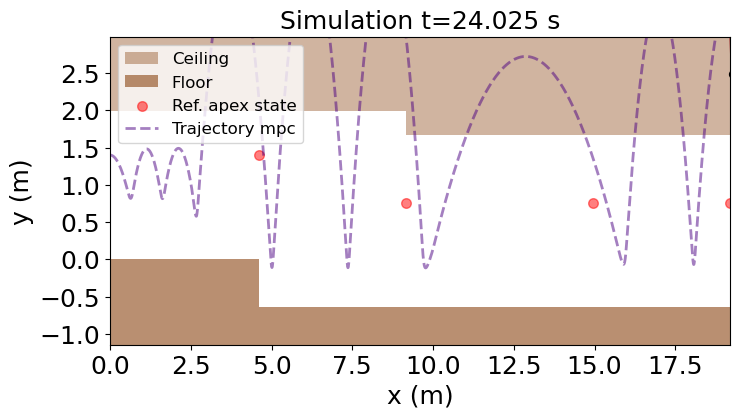}
}
\hfill
\subfigure[Ours]{
\includegraphics[width=0.180\textwidth]{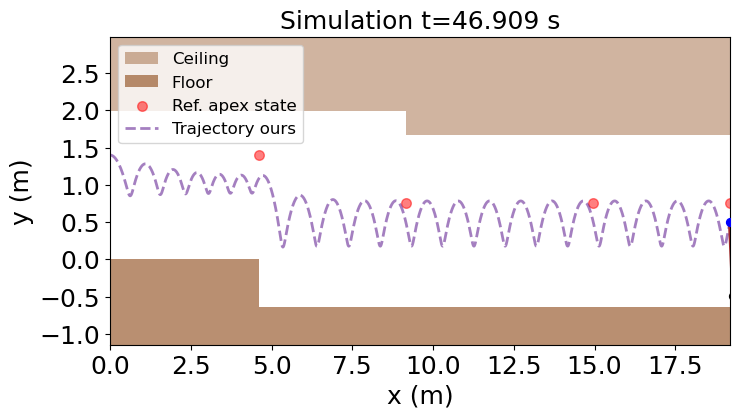}
}
\hfill
}
\end{figure}
\begin{figure}[!htbp]
\floatconts{fig:supple-pogo-sim-19}
{\caption{Pogobot simulation comparisons (trial 19)}}
{
\subfigure[RL-SAC]{
\includegraphics[width=0.180\textwidth]{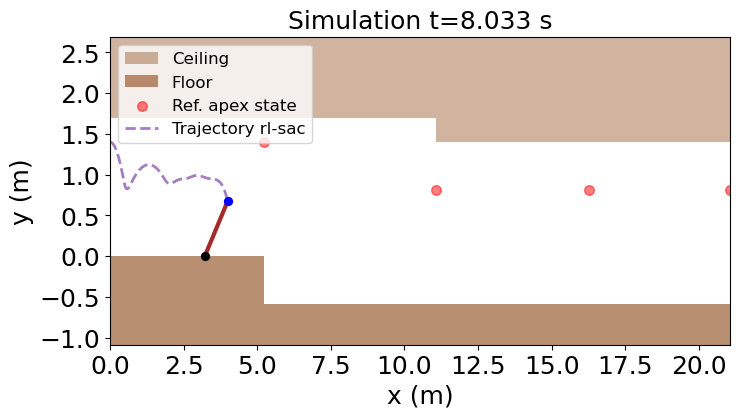}
}
\hfill
\subfigure[RL-PPO]{
\includegraphics[width=0.180\textwidth]{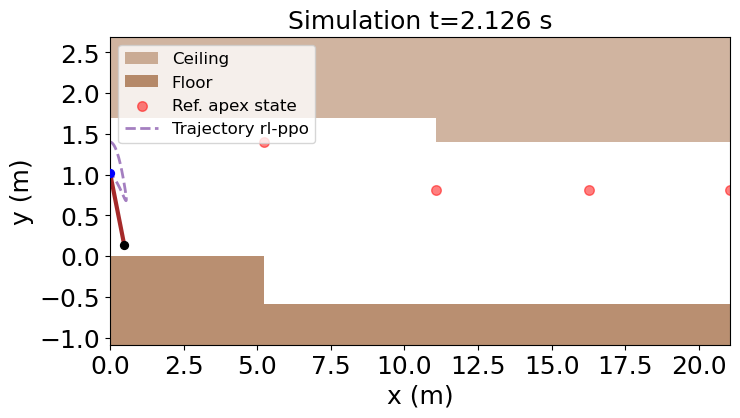}
}
\hfill
\subfigure[RL-DDPG]{
\includegraphics[width=0.180\textwidth]{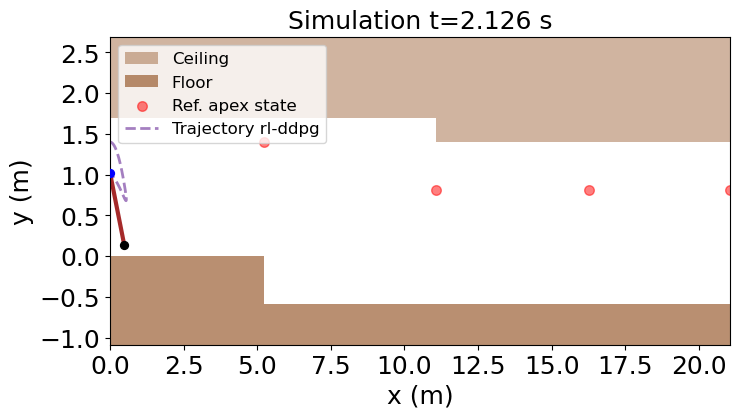}
}
\hfill
\subfigure[MPC]{
\includegraphics[width=0.180\textwidth]{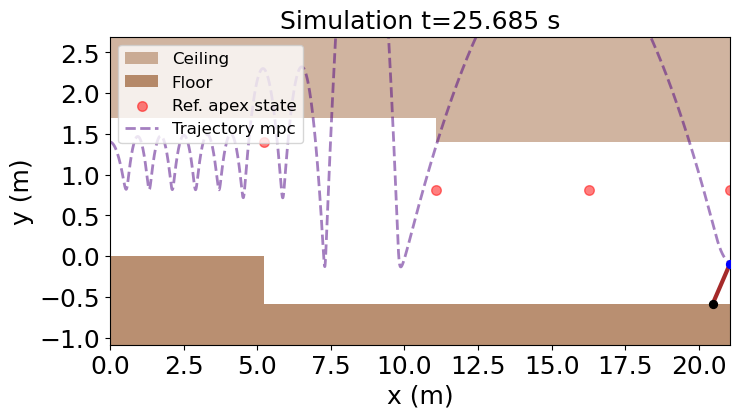}
}
\hfill
\subfigure[Ours]{
\includegraphics[width=0.180\textwidth]{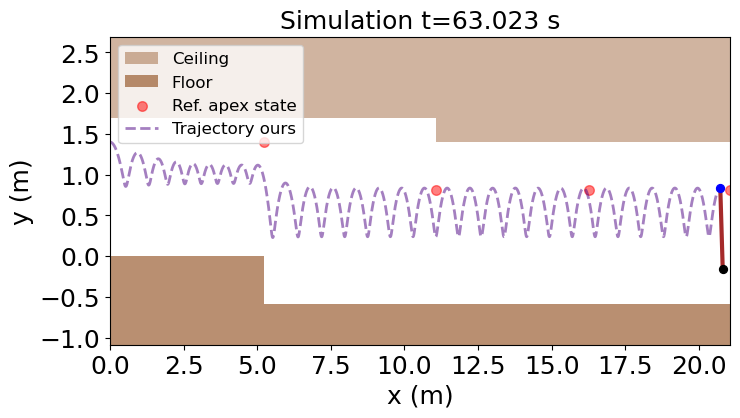}
}
\hfill
}
\end{figure}
\begin{figure}[!htbp]
\floatconts{fig:supple-pogo-sim-20}
{\caption{Pogobot simulation comparisons (trial 20)}}
{
\subfigure[RL-SAC]{
\includegraphics[width=0.180\textwidth]{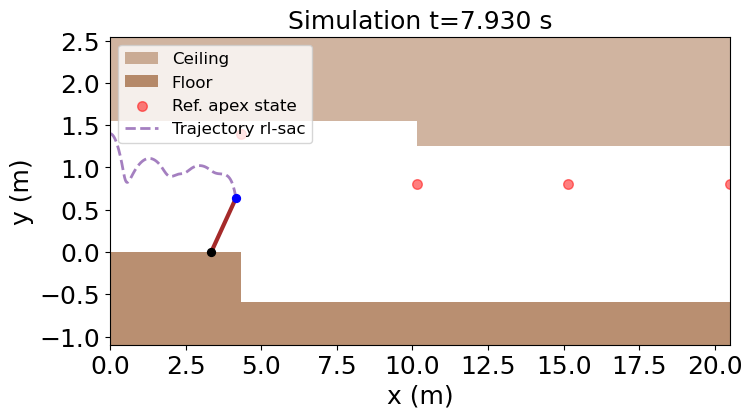}
}
\hfill
\subfigure[RL-PPO]{
\includegraphics[width=0.180\textwidth]{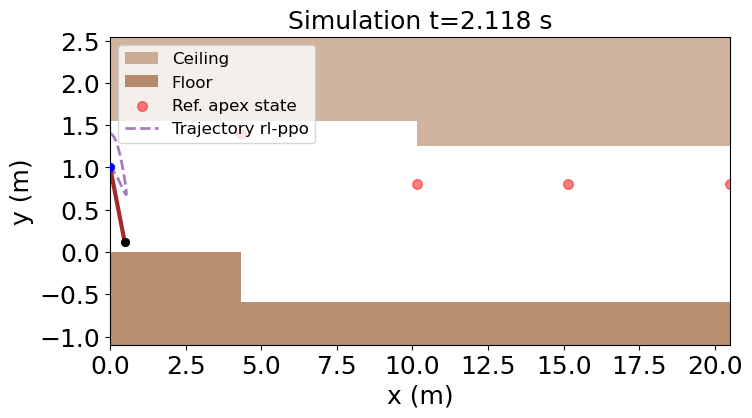}
}
\hfill
\subfigure[RL-DDPG]{
\includegraphics[width=0.180\textwidth]{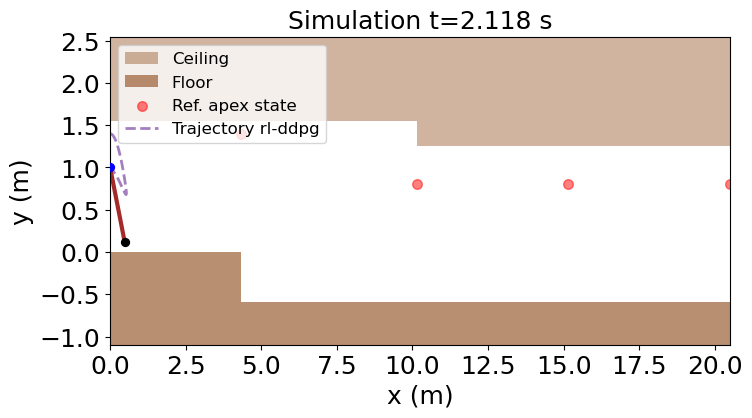}
}
\hfill
\subfigure[MPC]{
\includegraphics[width=0.180\textwidth]{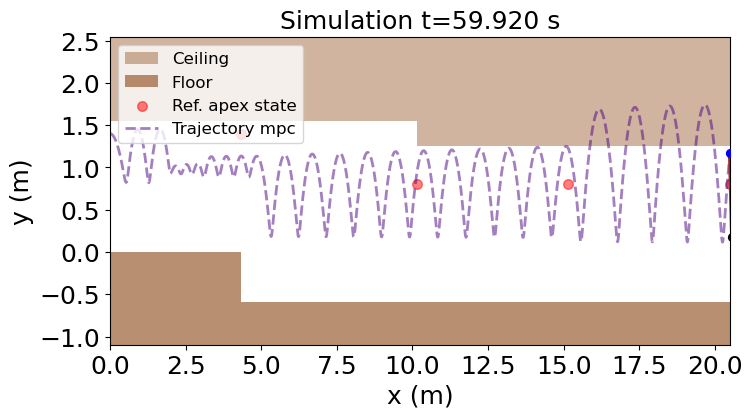}
}
\hfill
\subfigure[Ours]{
\includegraphics[width=0.180\textwidth]{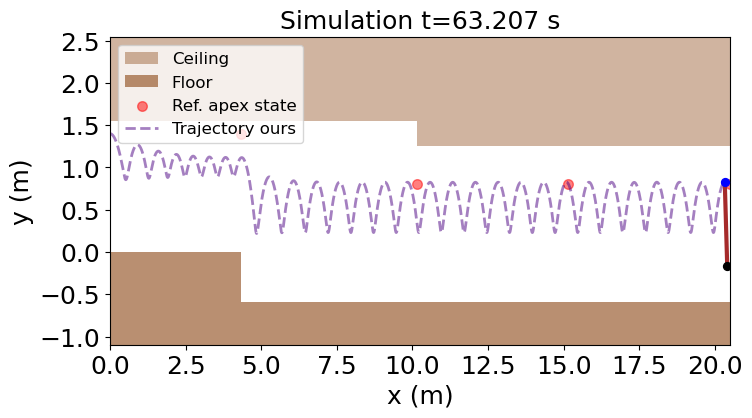}
}
\hfill
}
\end{figure}
\begin{figure}[!htbp]
\floatconts{fig:supple-pogo-sim-21}
{\caption{Pogobot simulation comparisons (trial 21)}}
{
\subfigure[RL-SAC]{
\includegraphics[width=0.180\textwidth]{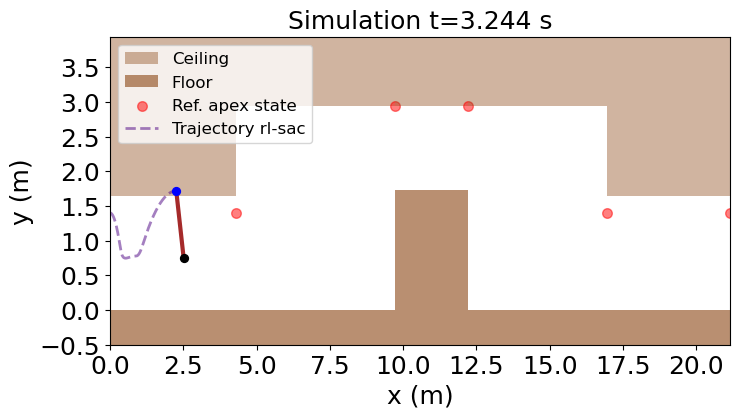}
}
\hfill
\subfigure[RL-PPO]{
\includegraphics[width=0.180\textwidth]{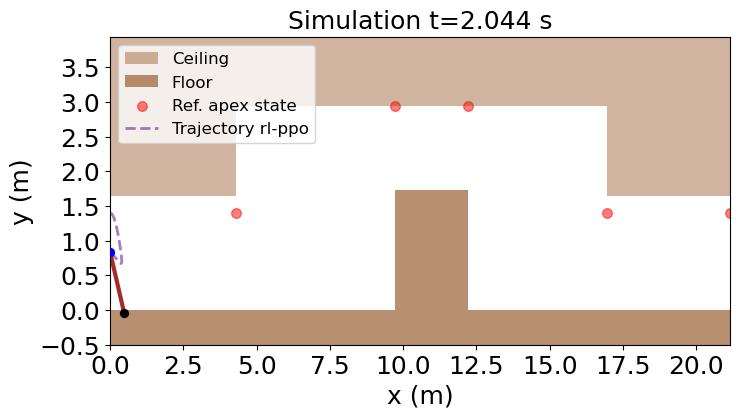}
}
\hfill
\subfigure[RL-DDPG]{
\includegraphics[width=0.180\textwidth]{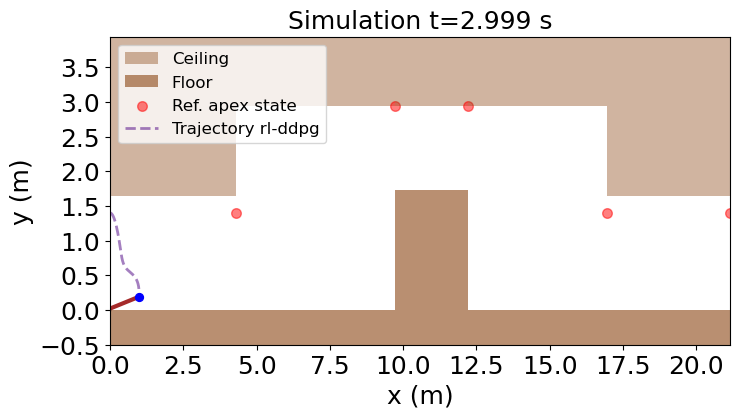}
}
\hfill
\subfigure[MPC]{
\includegraphics[width=0.180\textwidth]{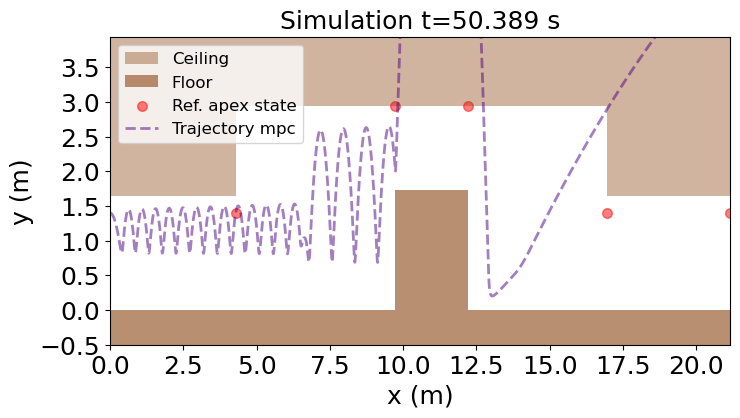}
}
\hfill
\subfigure[Ours]{
\includegraphics[width=0.180\textwidth]{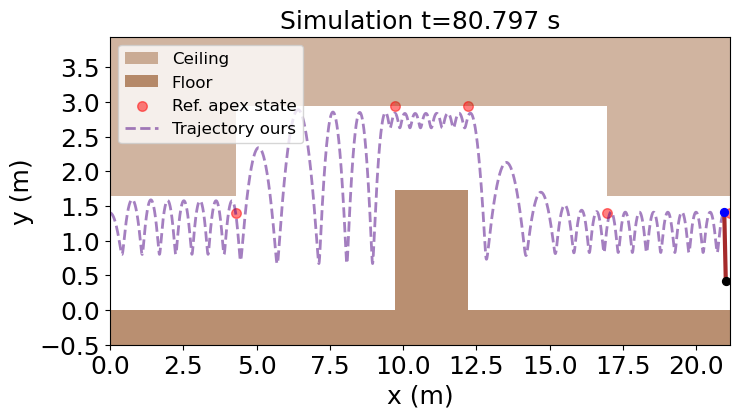}
}
\hfill
}
\end{figure}
\begin{figure}[!htbp]
\floatconts{fig:supple-pogo-sim-22}
{\caption{Pogobot simulation comparisons (trial 22)}}
{
\subfigure[RL-SAC]{
\includegraphics[width=0.180\textwidth]{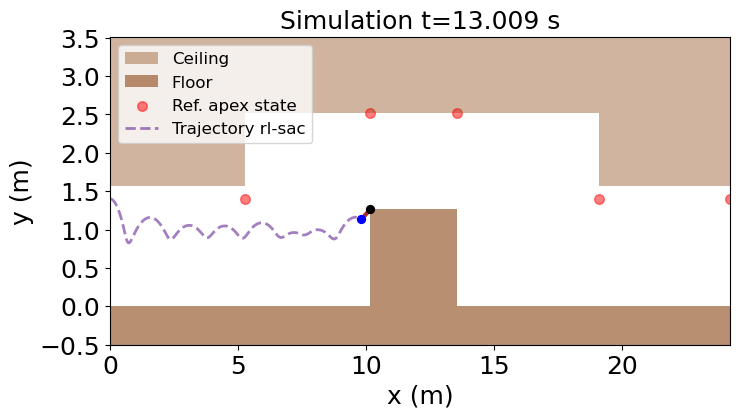}
}
\hfill
\subfigure[RL-PPO]{
\includegraphics[width=0.180\textwidth]{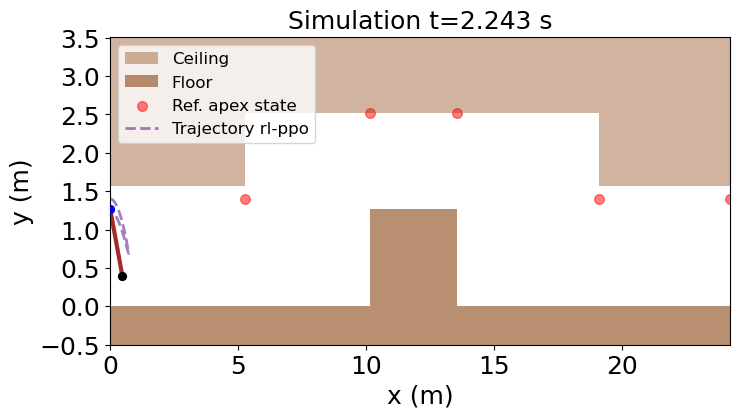}
}
\hfill
\subfigure[RL-DDPG]{
\includegraphics[width=0.180\textwidth]{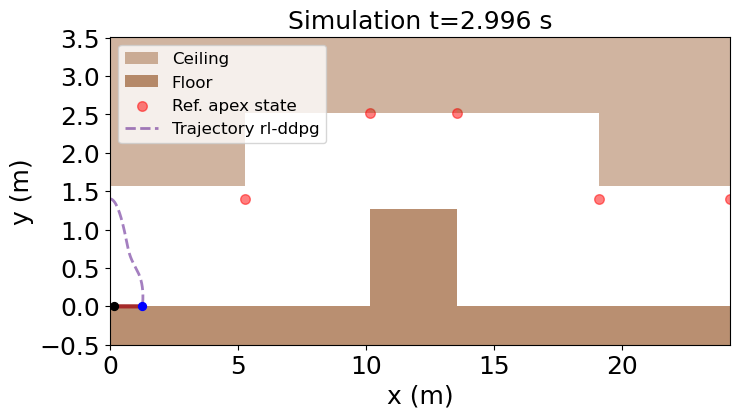}
}
\hfill
\subfigure[MPC]{
\includegraphics[width=0.180\textwidth]{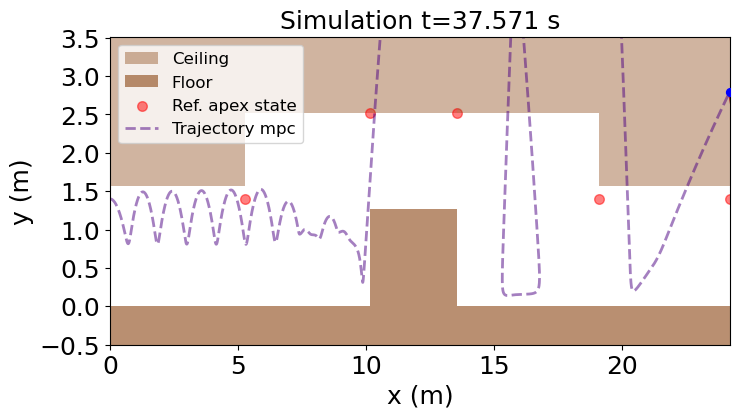}
}
\hfill
\subfigure[Ours]{
\includegraphics[width=0.180\textwidth]{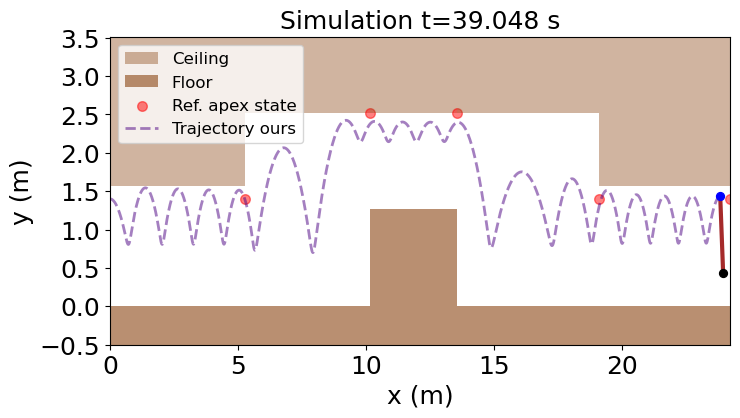}
}
\hfill
}
\end{figure}
\begin{figure}[!htbp]
\floatconts{fig:supple-pogo-sim-23}
{\caption{Pogobot simulation comparisons (trial 23)}}
{
\subfigure[RL-SAC]{
\includegraphics[width=0.180\textwidth]{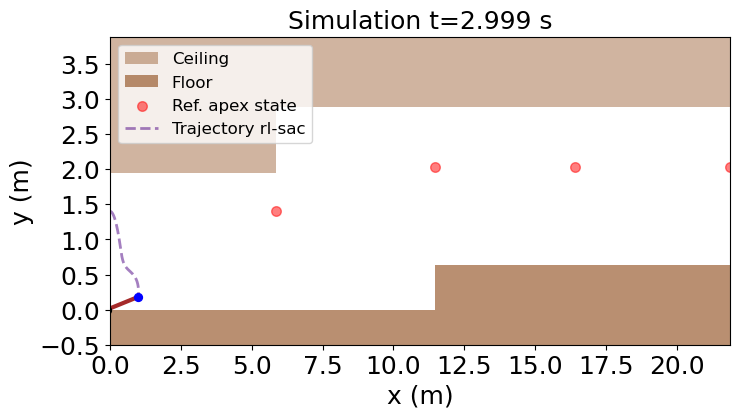}
}
\hfill
\subfigure[RL-PPO]{
\includegraphics[width=0.180\textwidth]{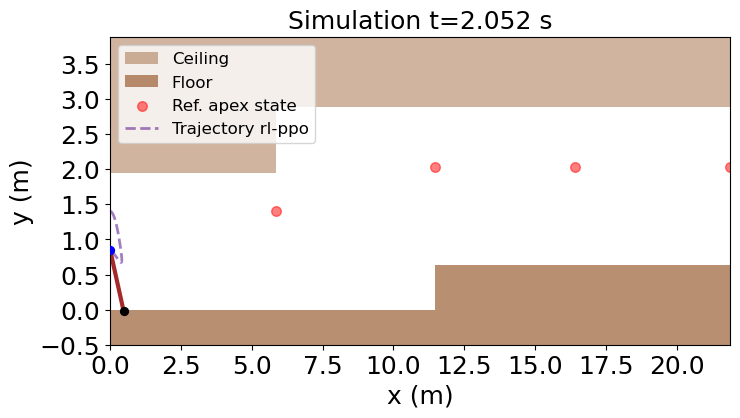}
}
\hfill
\subfigure[RL-DDPG]{
\includegraphics[width=0.180\textwidth]{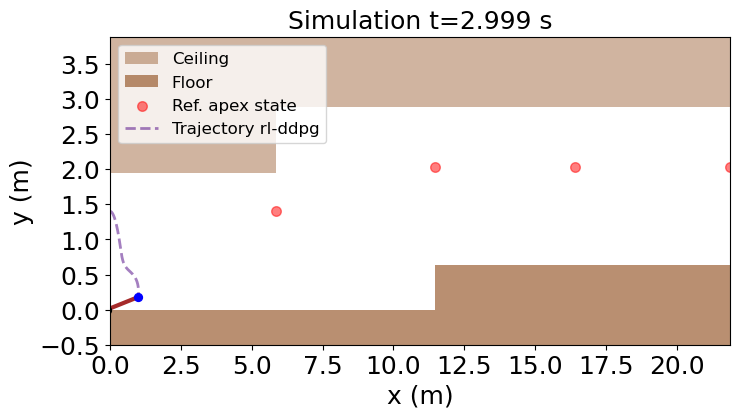}
}
\hfill
\subfigure[MPC]{
\includegraphics[width=0.180\textwidth]{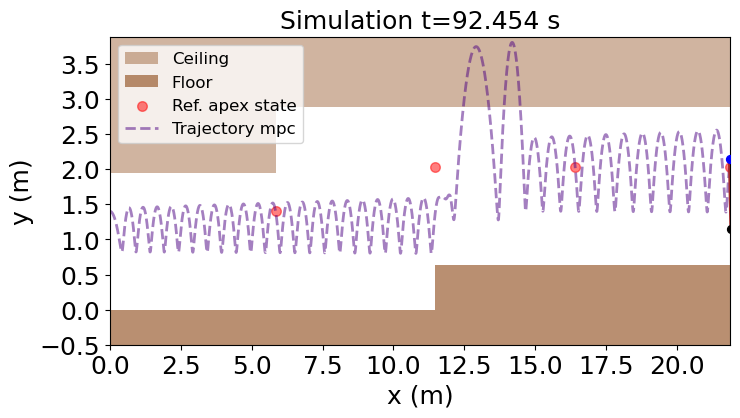}
}
\hfill
\subfigure[Ours]{
\includegraphics[width=0.180\textwidth]{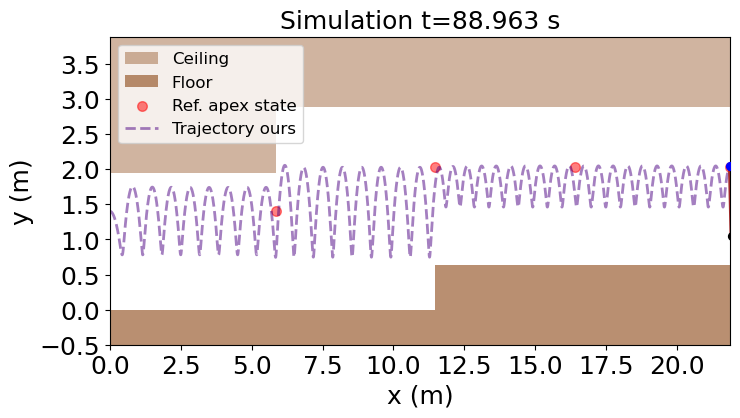}
}
\hfill
}
\end{figure}
\begin{figure}[!htbp]
\floatconts{fig:supple-pogo-sim-24}
{\caption{Pogobot simulation comparisons (trial 24)}}
{
\subfigure[RL-SAC]{
\includegraphics[width=0.180\textwidth]{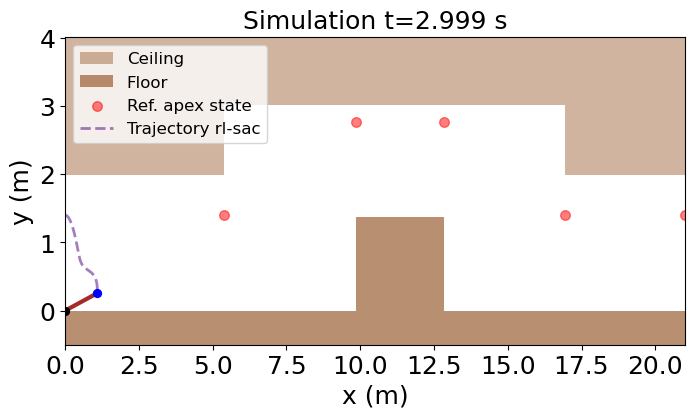}
}
\hfill
\subfigure[RL-PPO]{
\includegraphics[width=0.180\textwidth]{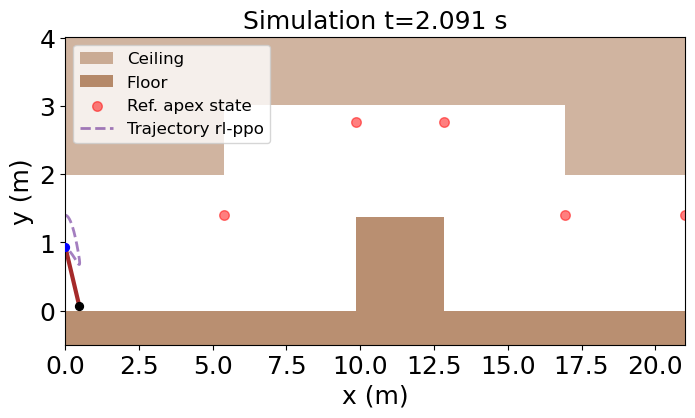}
}
\hfill
\subfigure[RL-DDPG]{
\includegraphics[width=0.180\textwidth]{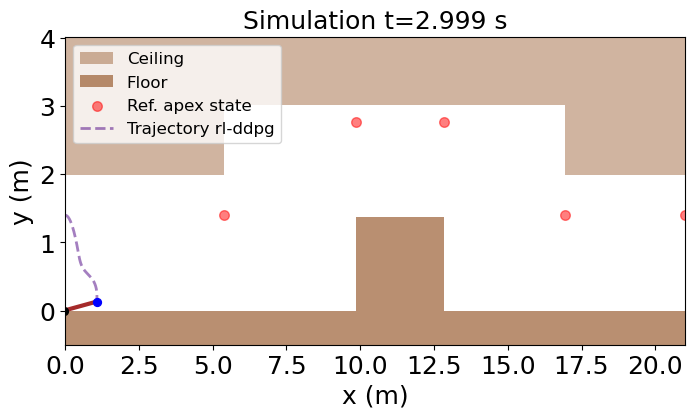}
}
\hfill
\subfigure[MPC]{
\includegraphics[width=0.180\textwidth]{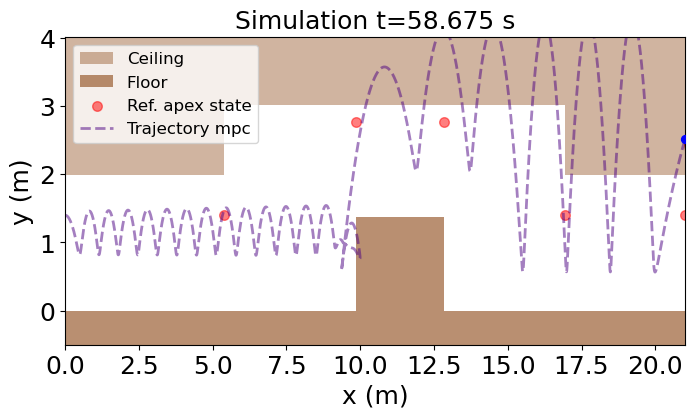}
}
\hfill
\subfigure[Ours]{
\includegraphics[width=0.180\textwidth]{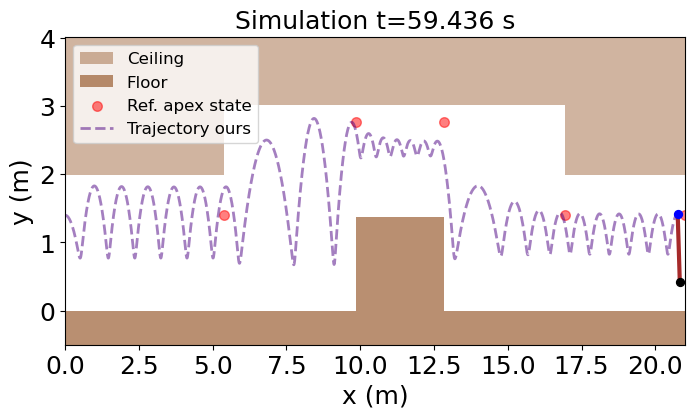}
}
\hfill
}
\end{figure}
\begin{figure}[!htbp]
\floatconts{fig:supple-pogo-sim-25}
{\caption{Pogobot simulation comparisons (trial 25)}}
{
\subfigure[RL-SAC]{
\includegraphics[width=0.180\textwidth]{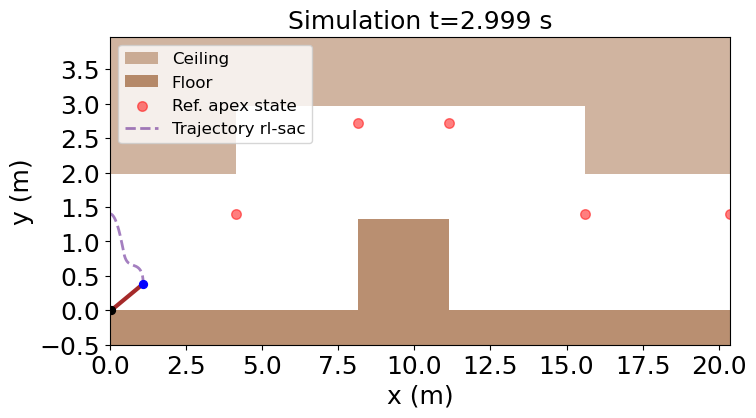}
}
\hfill
\subfigure[RL-PPO]{
\includegraphics[width=0.180\textwidth]{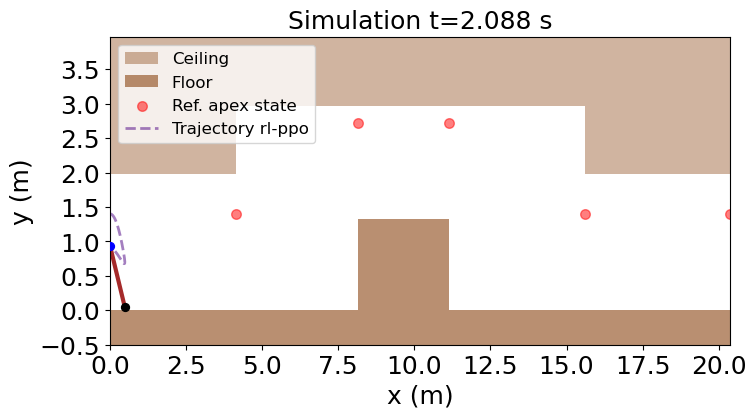}
}
\hfill
\subfigure[RL-DDPG]{
\includegraphics[width=0.180\textwidth]{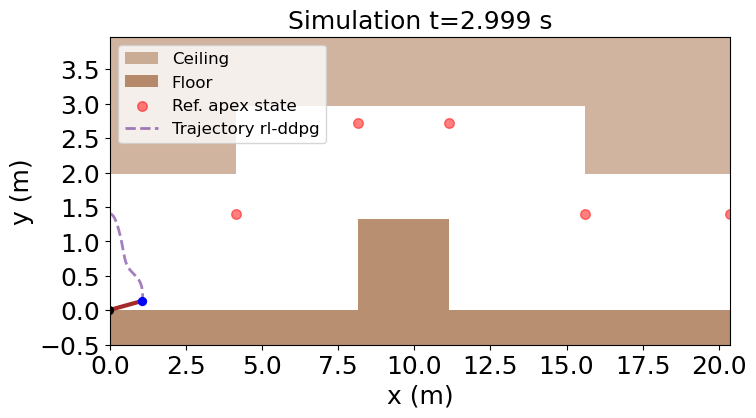}
}
\hfill
\subfigure[MPC]{
\includegraphics[width=0.180\textwidth]{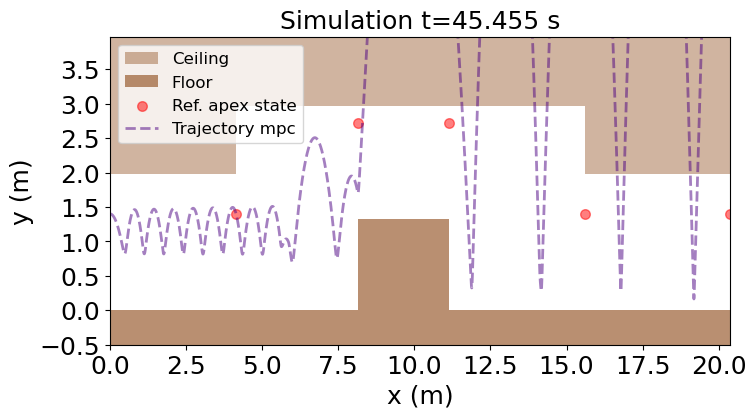}
}
\hfill
\subfigure[Ours]{
\includegraphics[width=0.180\textwidth]{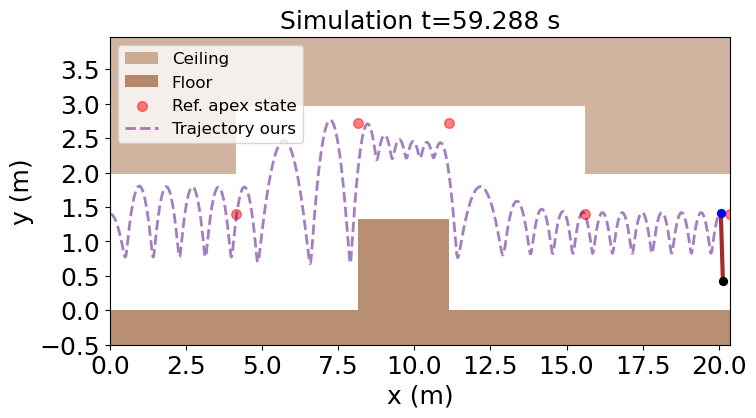}
}
\hfill
}
\end{figure}
\begin{figure}[!htbp]
\floatconts{fig:supple-pogo-sim-26}
{\caption{Pogobot simulation comparisons (trial 26)}}
{
\subfigure[RL-SAC]{
\includegraphics[width=0.180\textwidth]{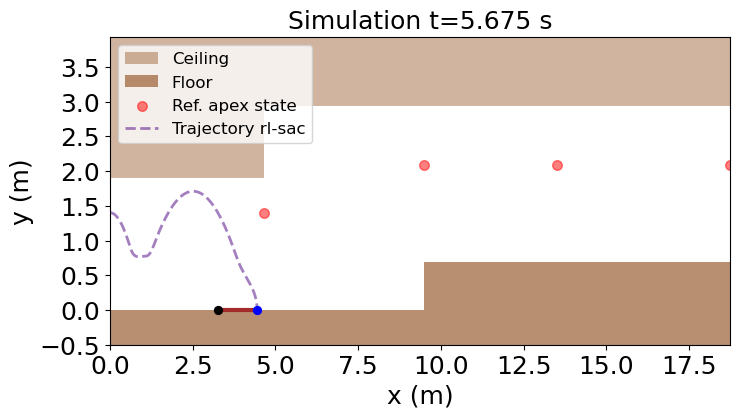}
}
\hfill
\subfigure[RL-PPO]{
\includegraphics[width=0.180\textwidth]{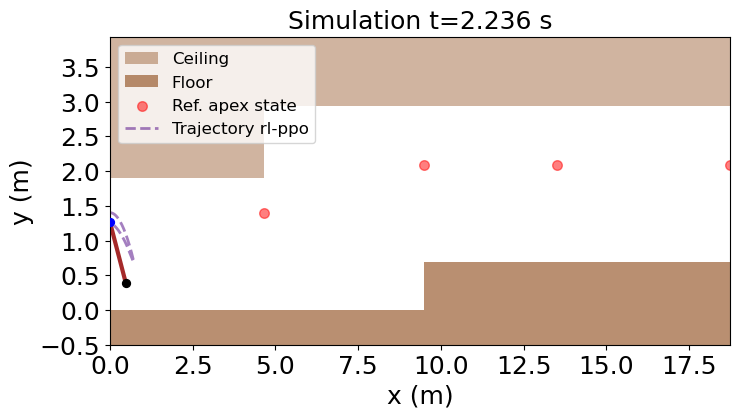}
}
\hfill
\subfigure[RL-DDPG]{
\includegraphics[width=0.180\textwidth]{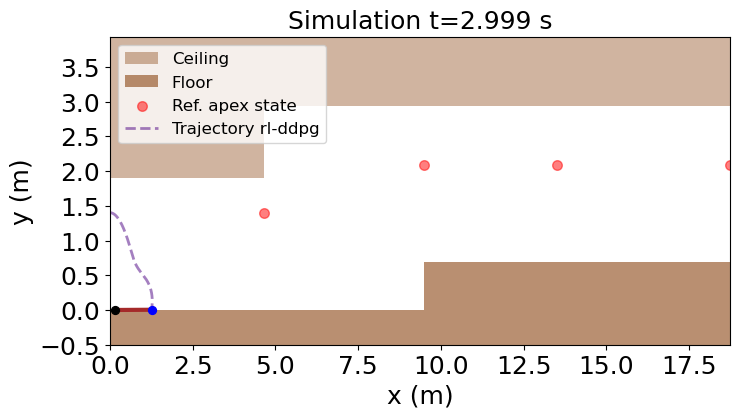}
}
\hfill
\subfigure[MPC]{
\includegraphics[width=0.180\textwidth]{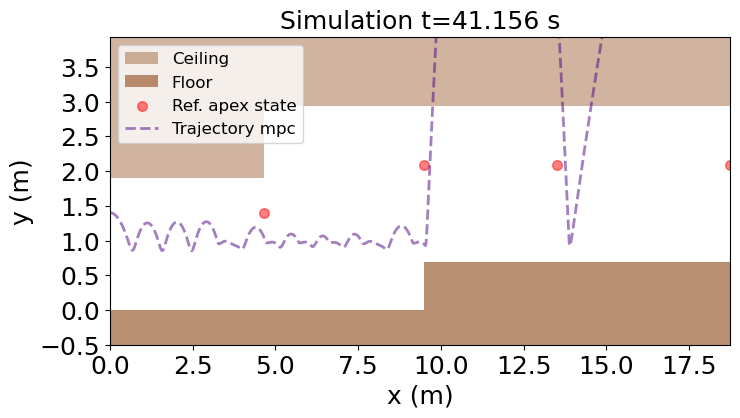}
}
\hfill
\subfigure[Ours]{
\includegraphics[width=0.180\textwidth]{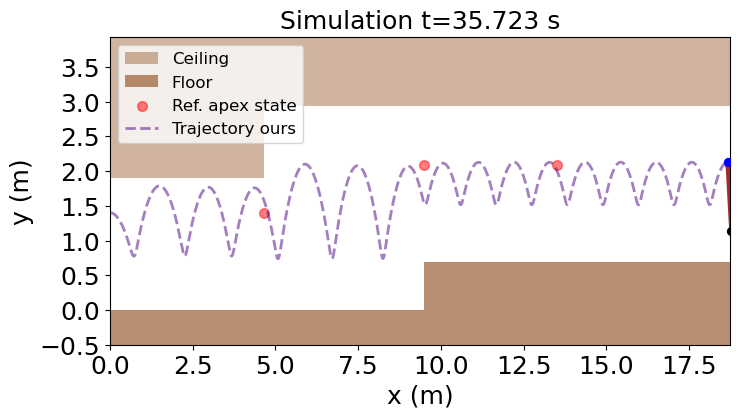}
}
\hfill
}
\end{figure}
\begin{figure}[!htbp]
\floatconts{fig:supple-pogo-sim-27}
{\caption{Pogobot simulation comparisons (trial 27)}}
{
\subfigure[RL-SAC]{
\includegraphics[width=0.180\textwidth]{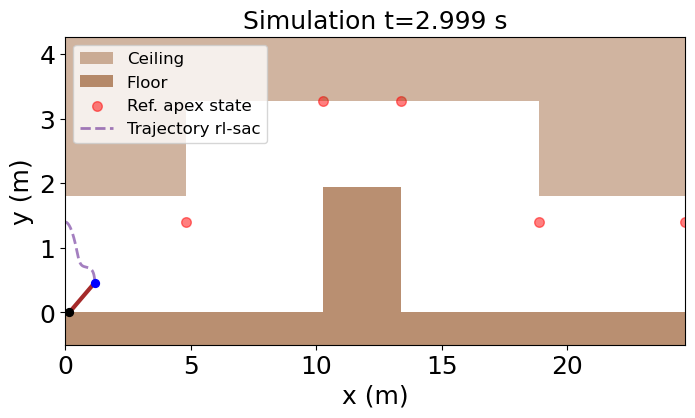}
}
\hfill
\subfigure[RL-PPO]{
\includegraphics[width=0.180\textwidth]{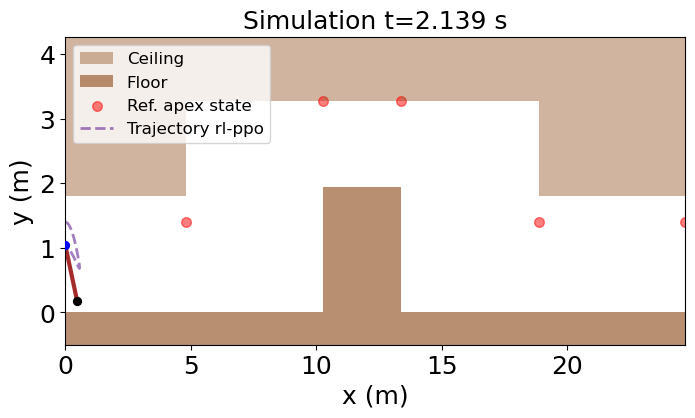}
}
\hfill
\subfigure[RL-DDPG]{
\includegraphics[width=0.180\textwidth]{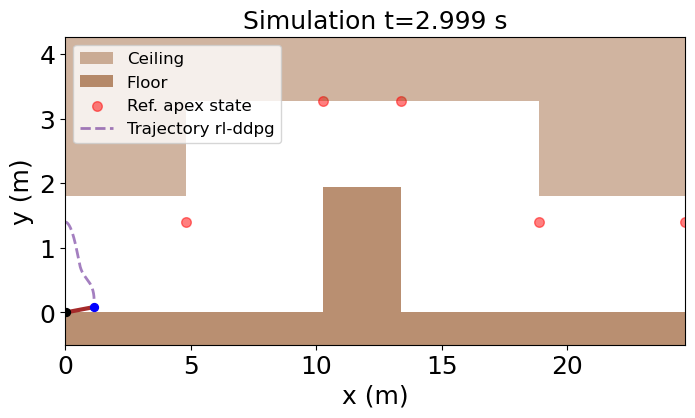}
}
\hfill
\subfigure[MPC]{
\includegraphics[width=0.180\textwidth]{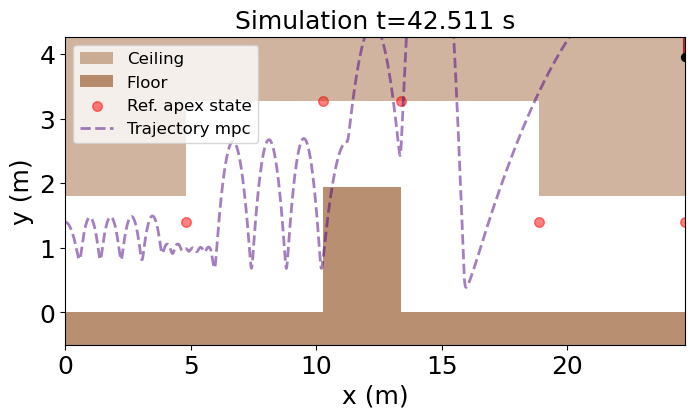}
}
\hfill
\subfigure[Ours]{
\includegraphics[width=0.180\textwidth]{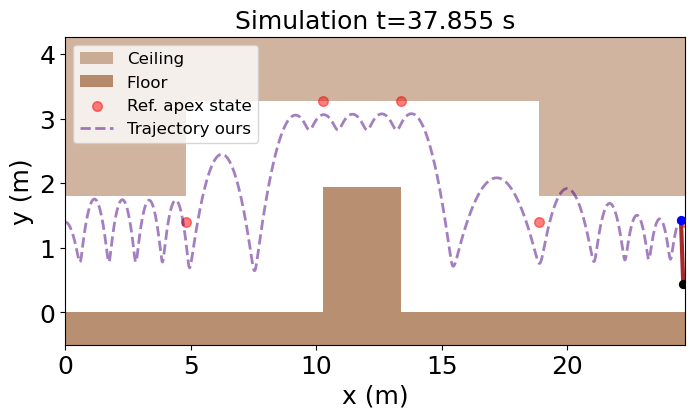}
}
\hfill
}
\end{figure}
\begin{figure}[!htbp]
\floatconts{fig:supple-pogo-sim-28}
{\caption{Pogobot simulation comparisons (trial 28)}}
{
\subfigure[RL-SAC]{
\includegraphics[width=0.180\textwidth]{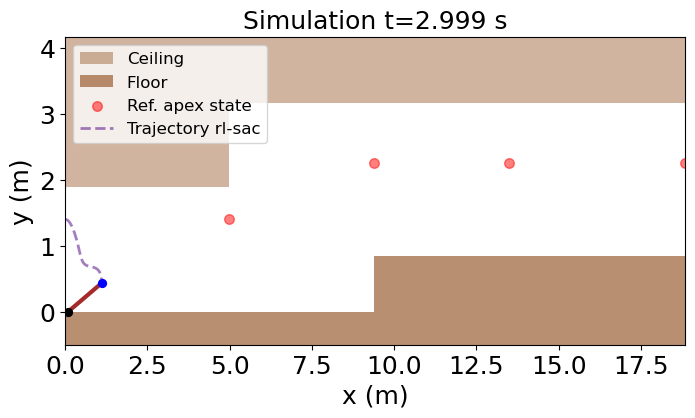}
}
\hfill
\subfigure[RL-PPO]{
\includegraphics[width=0.180\textwidth]{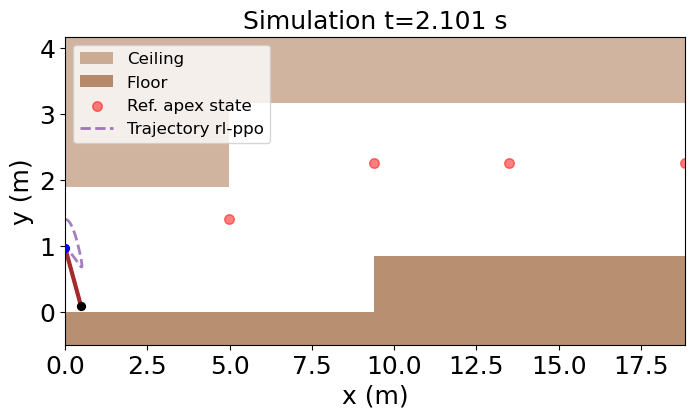}
}
\hfill
\subfigure[RL-DDPG]{
\includegraphics[width=0.180\textwidth]{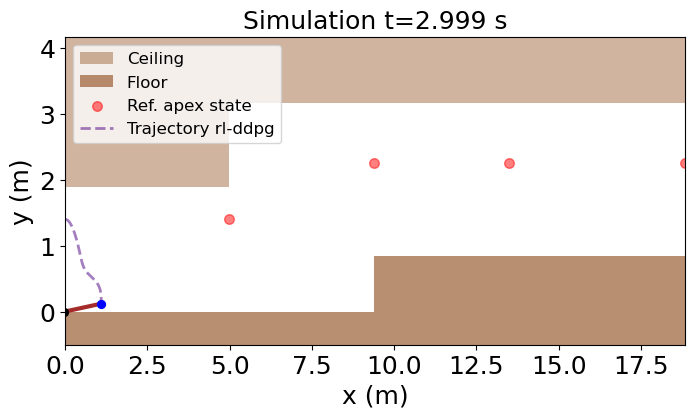}
}
\hfill
\subfigure[MPC]{
\includegraphics[width=0.180\textwidth]{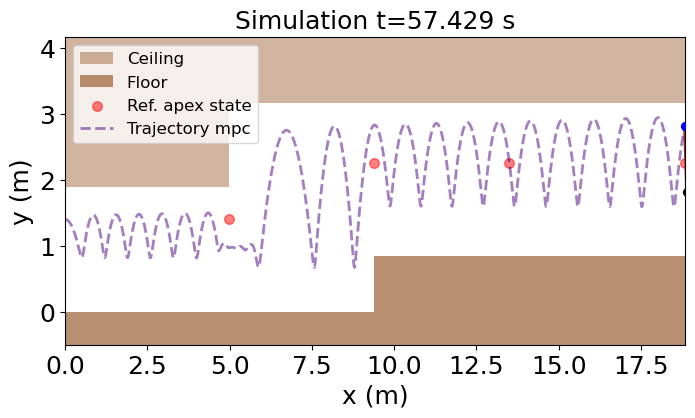}
}
\hfill
\subfigure[Ours]{
\includegraphics[width=0.180\textwidth]{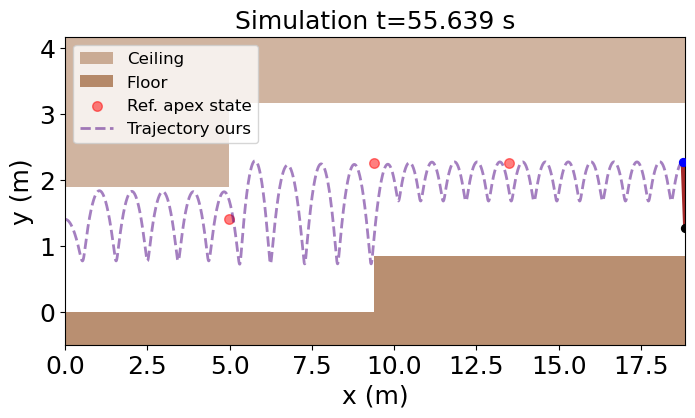}
}
\hfill
}
\end{figure}

\begin{figure}[!htbp]
\floatconts{fig:supple-cgw-sim-50}
{\caption{Bipedal walker simulation comparisons (same target angle)}}
{
\subfigure[RL-SAC]{
\includegraphics[width=0.125\textwidth]{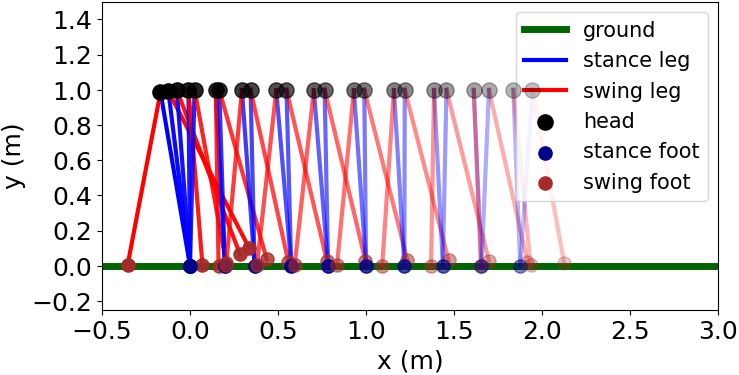}
}
\hfill
\subfigure[RL-PPO]{
\includegraphics[width=0.125\textwidth]{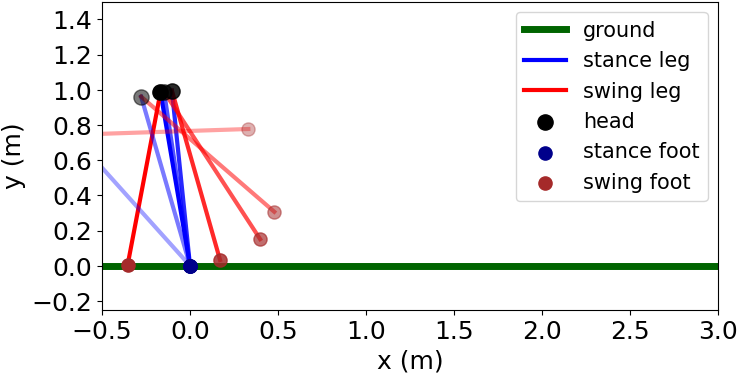}
}
\hfill
\subfigure[RL-DDPG]{
\includegraphics[width=0.125\textwidth]{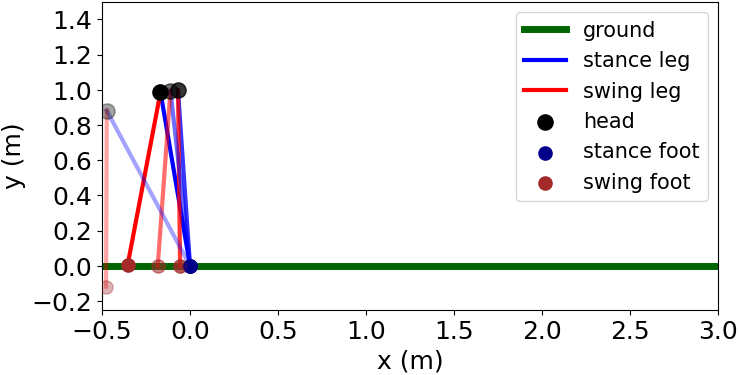}
}
\hfill
\subfigure[MPC]{
\includegraphics[width=0.125\textwidth]{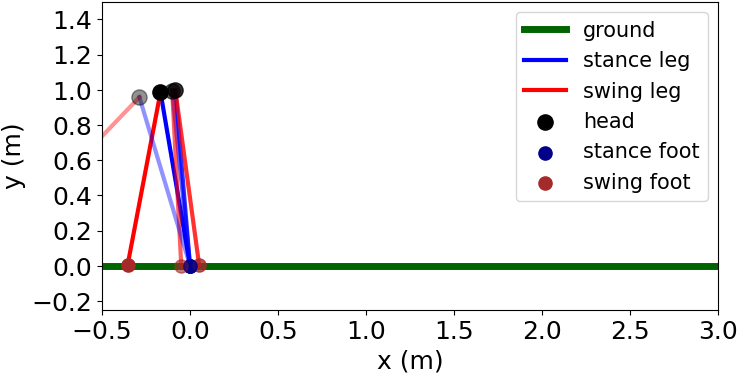}
}
\hfill
\subfigure[QP]{
\includegraphics[width=0.125\textwidth]{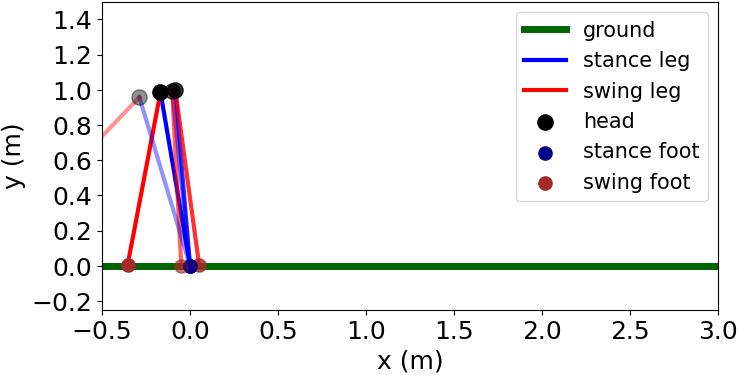}
}
\hfill
\subfigure[HJB]{
\includegraphics[width=0.125\textwidth]{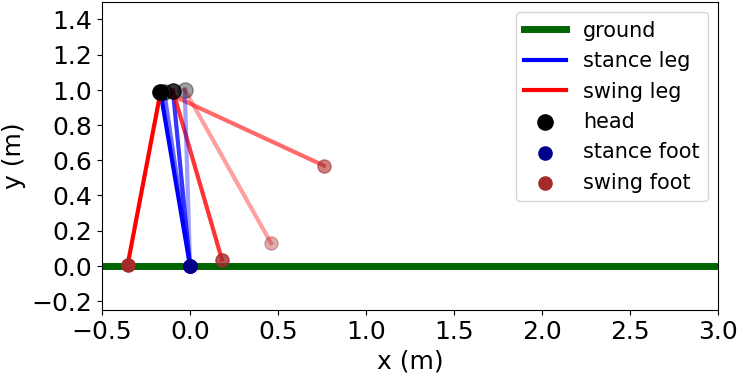}
}
\hfill
\subfigure[Ours]{
\includegraphics[width=0.125\textwidth]{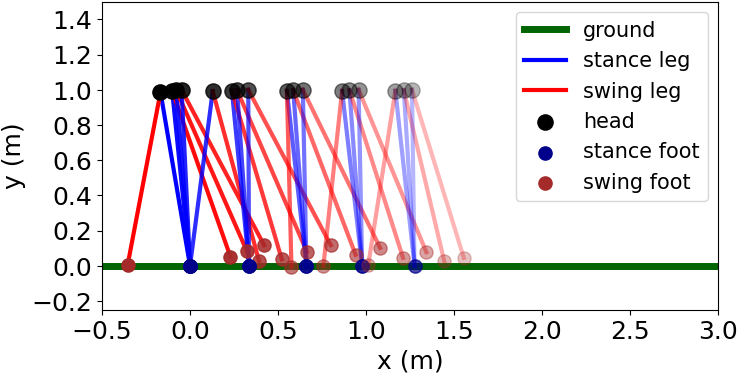}
}
\hfill
}
\end{figure}
\begin{figure}[!htbp]
\floatconts{fig:supple-cgw-sim-80}
{\caption{Bipedal walker simulation comparisons (same target angle)}}
{
\subfigure[RL-SAC]{
\includegraphics[width=0.125\textwidth]{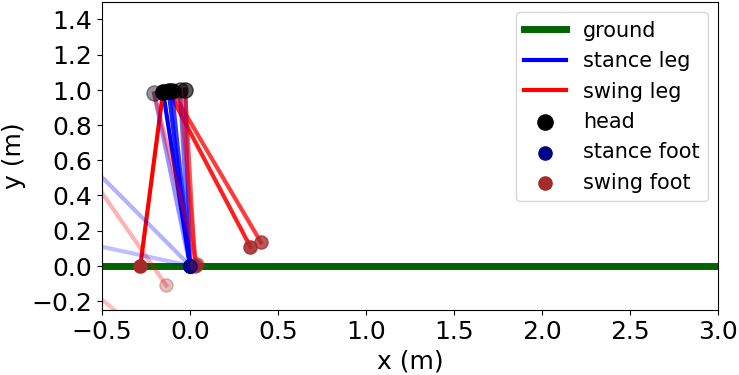}
}
\hfill
\subfigure[RL-PPO]{
\includegraphics[width=0.125\textwidth]{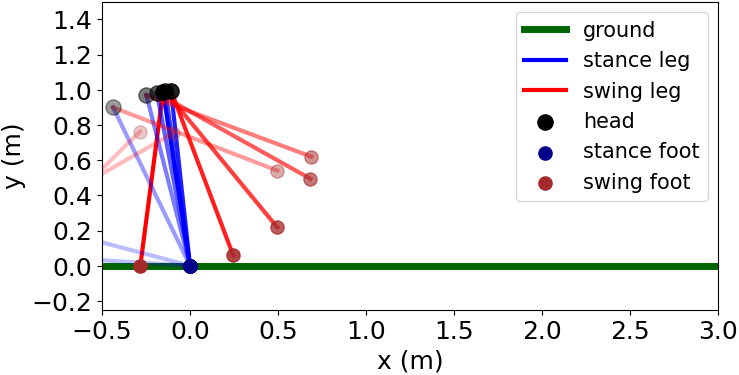}
}
\hfill
\subfigure[RL-DDPG]{
\includegraphics[width=0.125\textwidth]{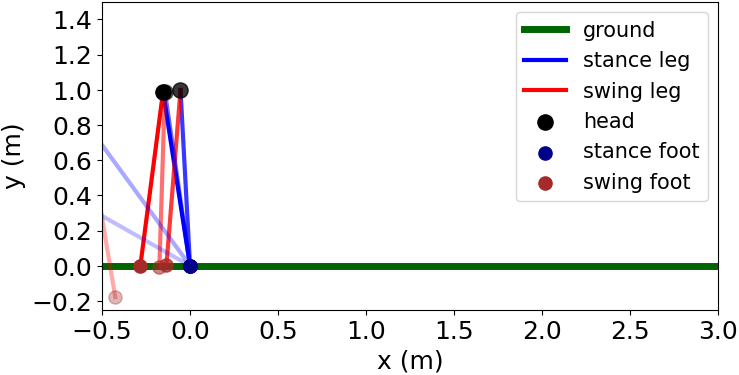}
}
\hfill
\subfigure[MPC]{
\includegraphics[width=0.125\textwidth]{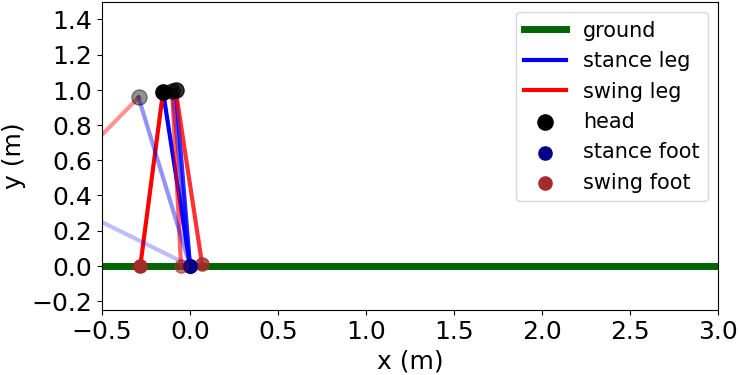}
}
\hfill
\subfigure[QP]{
\includegraphics[width=0.125\textwidth]{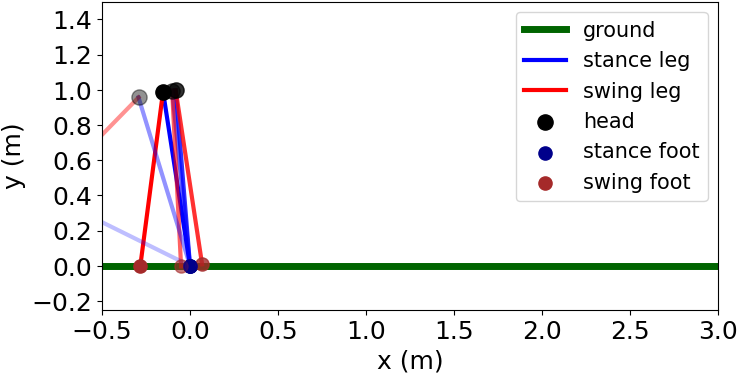}
}
\hfill
\subfigure[HJB]{
\includegraphics[width=0.125\textwidth]{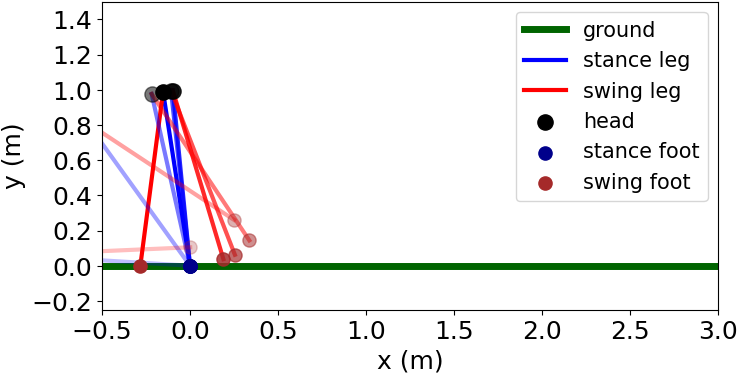}
}
\hfill
\subfigure[Ours]{
\includegraphics[width=0.125\textwidth]{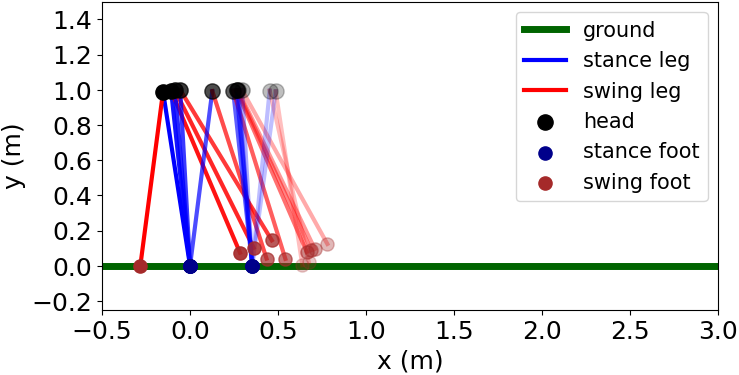}
}
\hfill
}
\end{figure}
\begin{figure}[!htbp]
\floatconts{fig:supple-cgw-simx-100}
{\caption{Bipedal walker simulation comparisons (different target angles)}}
{
\subfigure[RL-SAC]{
\includegraphics[width=0.145\textwidth]{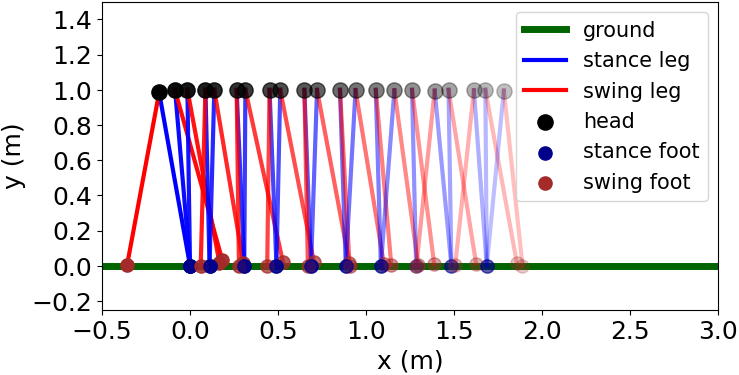}
}
\hfill
\subfigure[RL-PPO]{
\includegraphics[width=0.145\textwidth]{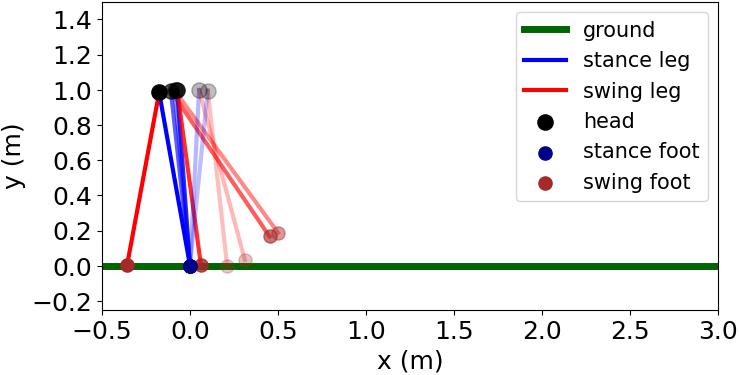}
}
\hfill
\subfigure[RL-DDPG]{
\includegraphics[width=0.145\textwidth]{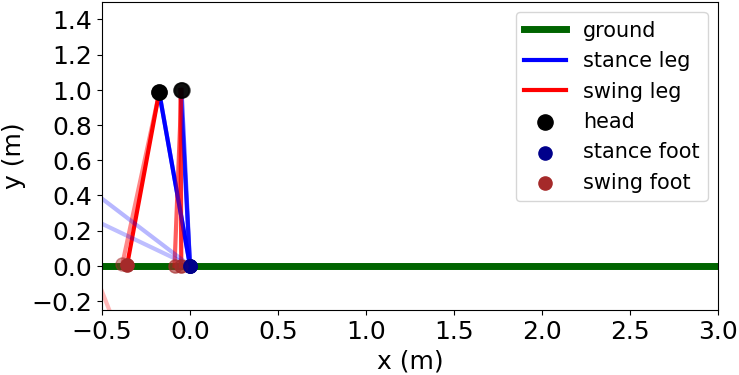}
}
\hfill
\subfigure[MPC]{
\includegraphics[width=0.145\textwidth]{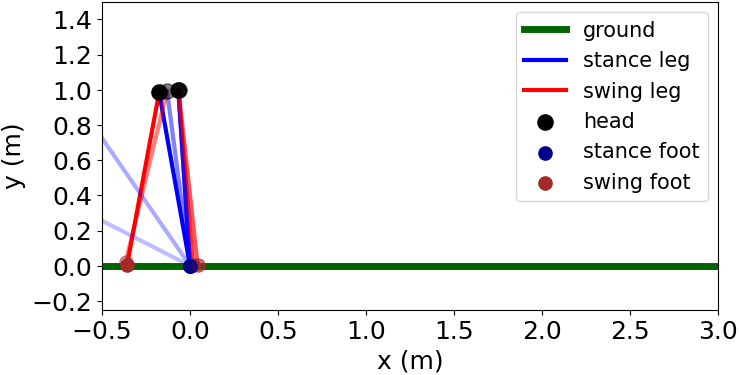}
}
\hfill
\subfigure[QP]{
\includegraphics[width=0.145\textwidth]{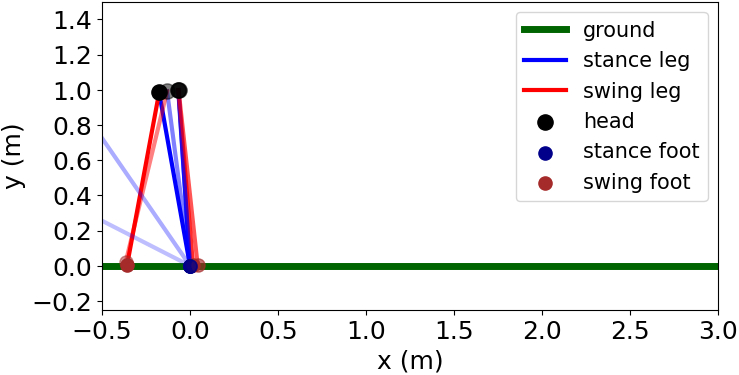}
}
\hfill
\subfigure[Ours]{
\includegraphics[width=0.145\textwidth]{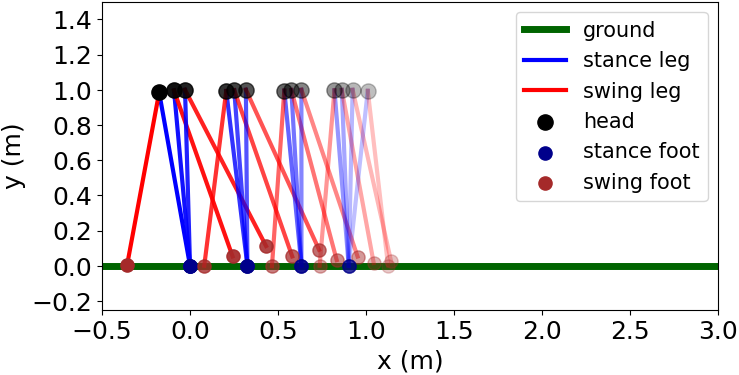}
}
\hfill
}
\end{figure}
\begin{figure}[!htbp]
\floatconts{fig:supple-cgw-simx-130}
{\caption{Bipedal walker simulation comparisons (different target angles)}}
{
\subfigure[RL-SAC]{
\includegraphics[width=0.145\textwidth]{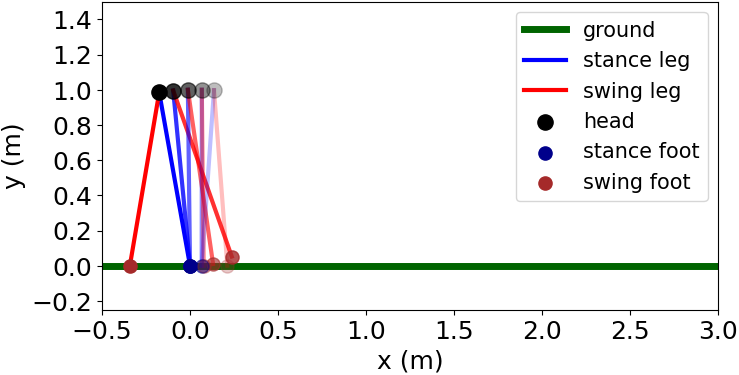}
}
\hfill
\subfigure[RL-PPO]{
\includegraphics[width=0.145\textwidth]{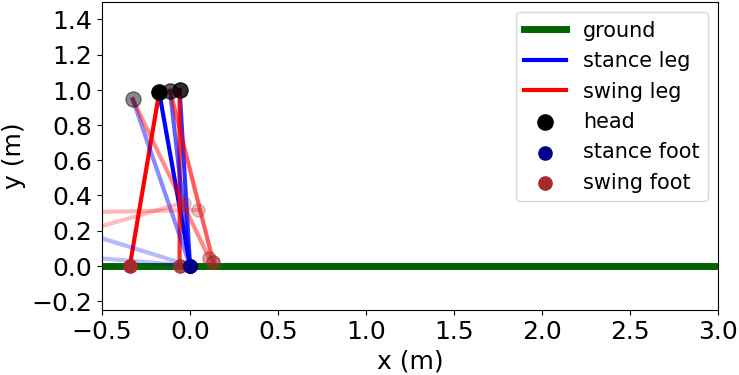}
}
\hfill
\subfigure[RL-DDPG]{
\includegraphics[width=0.145\textwidth]{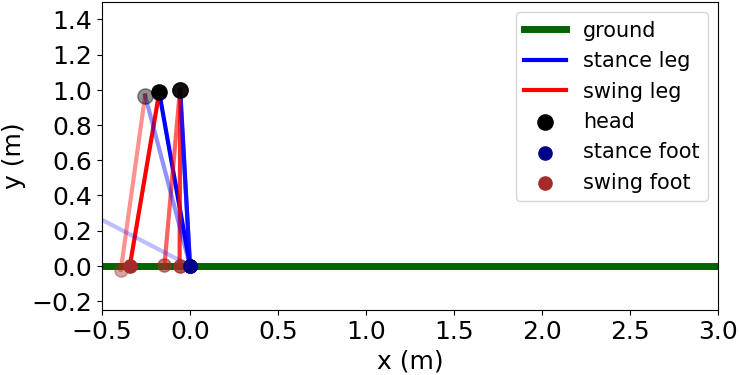}
}
\hfill
\subfigure[MPC]{
\includegraphics[width=0.145\textwidth]{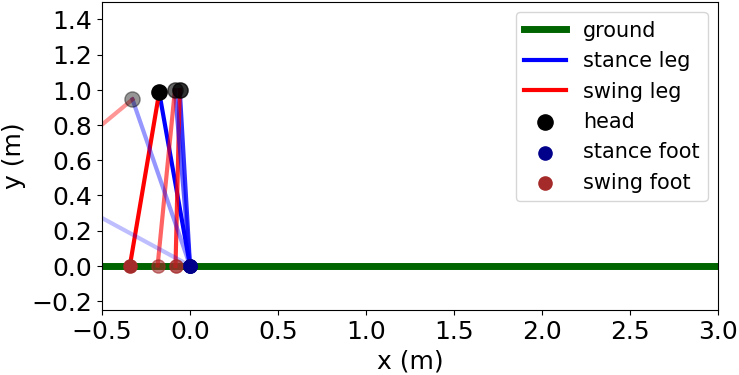}
}
\hfill
\subfigure[QP]{
\includegraphics[width=0.145\textwidth]{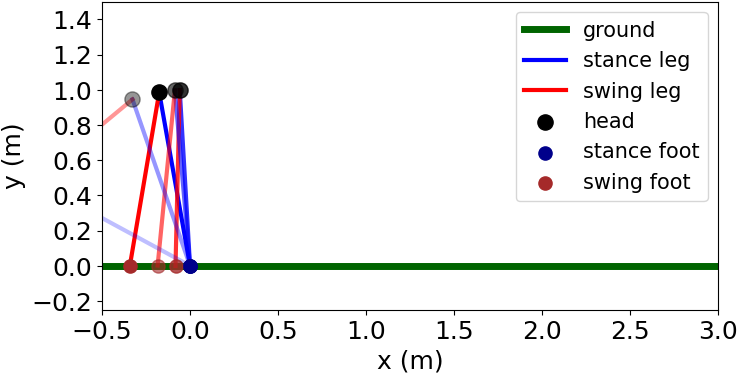}
}
\hfill
\subfigure[Ours]{
\includegraphics[width=0.145\textwidth]{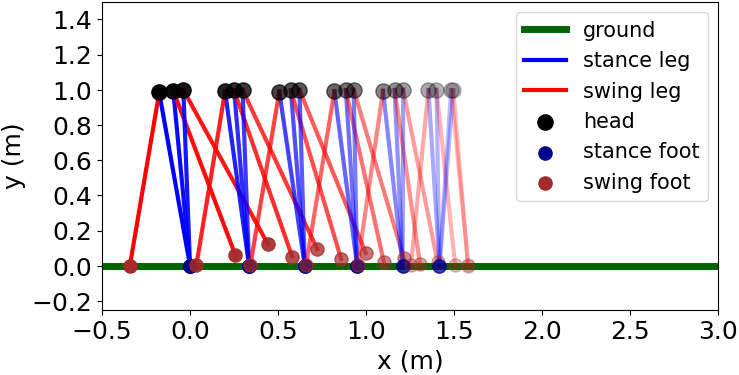}
}
\hfill
}
\end{figure}
\begin{figure}[!htbp]
\floatconts{fig:supple-cgw-simx-260}
{\caption{Bipedal walker simulation comparisons (different target angles)}}
{
\subfigure[RL-SAC]{
\includegraphics[width=0.145\textwidth]{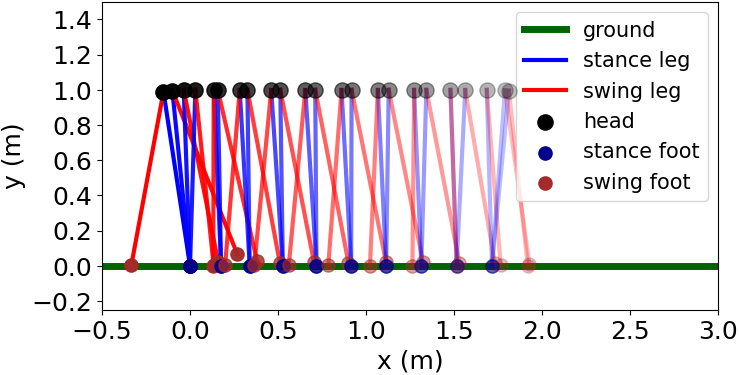}
}
\hfill
\subfigure[RL-PPO]{
\includegraphics[width=0.145\textwidth]{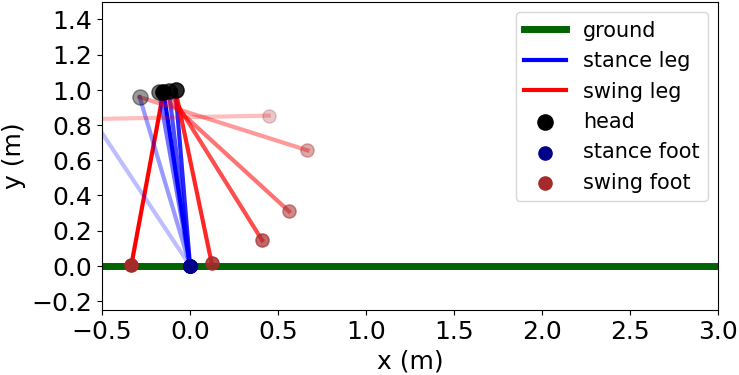}
}
\hfill
\subfigure[RL-DDPG]{
\includegraphics[width=0.145\textwidth]{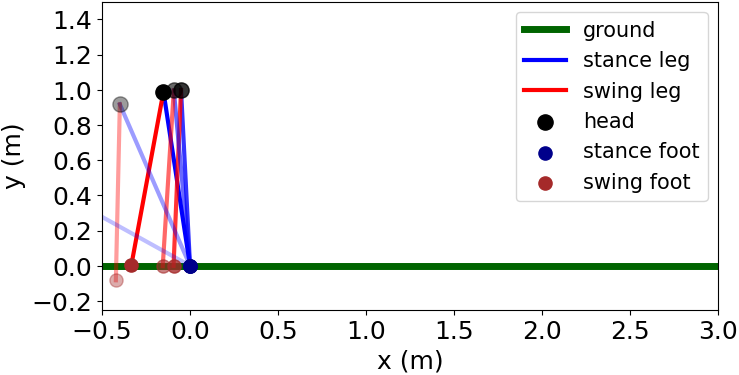}
}
\hfill
\subfigure[MPC]{
\includegraphics[width=0.145\textwidth]{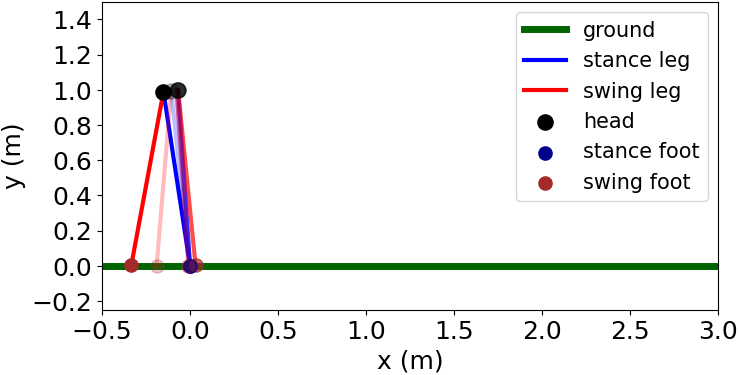}
}
\hfill
\subfigure[QP]{
\includegraphics[width=0.145\textwidth]{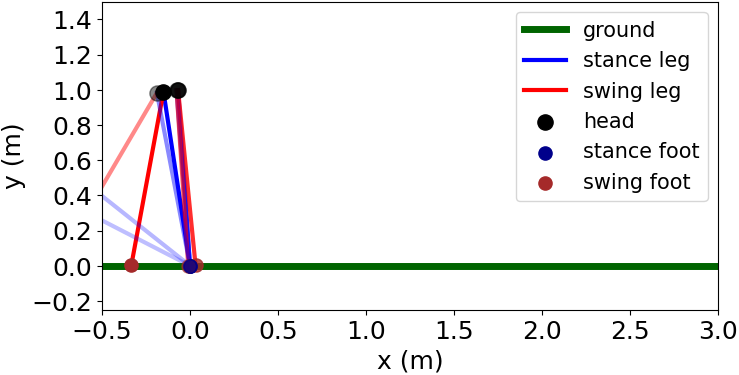}
}
\hfill
\subfigure[Ours]{
\includegraphics[width=0.145\textwidth]{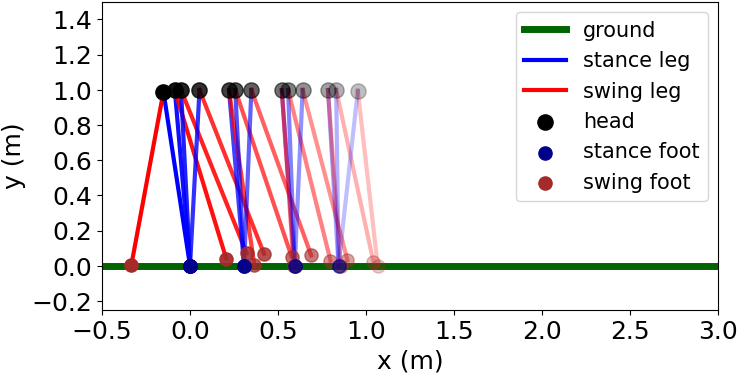}
}
\hfill
}
\end{figure}
\begin{figure}[!htbp]
\floatconts{fig:supple-cgw-simx-430}
{\caption{Bipedal walker simulation comparisons (different target angles)}}
{
\subfigure[RL-SAC]{
\includegraphics[width=0.145\textwidth]{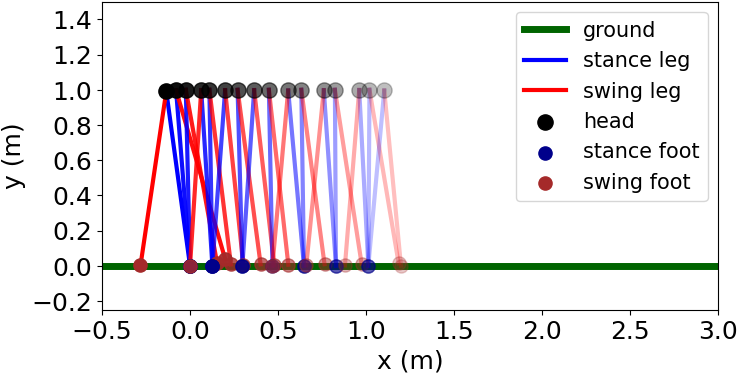}
}
\hfill
\subfigure[RL-PPO]{
\includegraphics[width=0.145\textwidth]{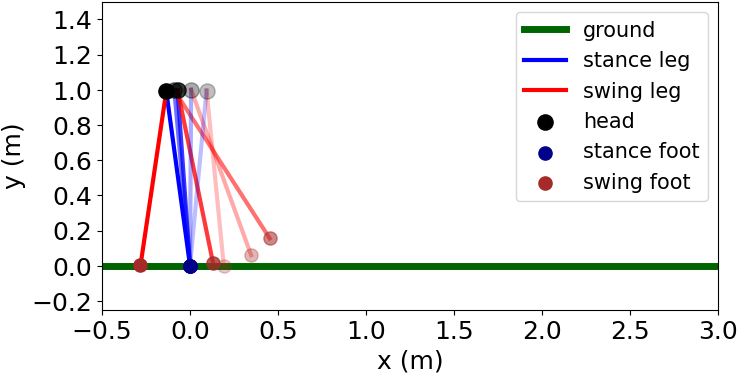}
}
\hfill
\subfigure[RL-DDPG]{
\includegraphics[width=0.145\textwidth]{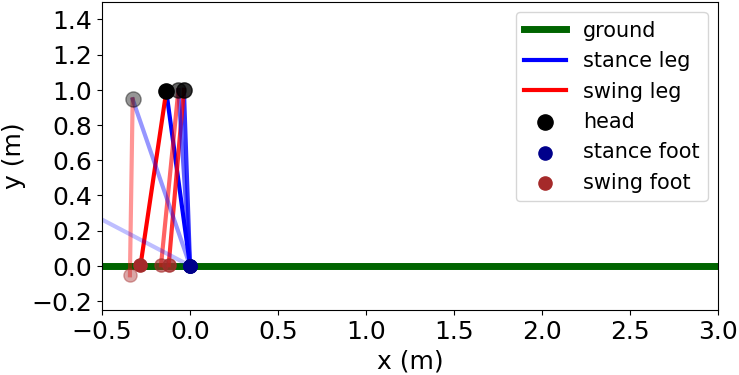}
}
\hfill
\subfigure[MPC]{
\includegraphics[width=0.145\textwidth]{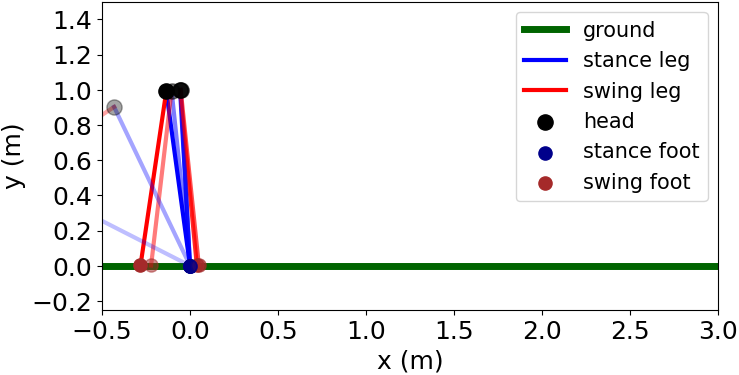}
}
\hfill
\subfigure[QP]{
\includegraphics[width=0.145\textwidth]{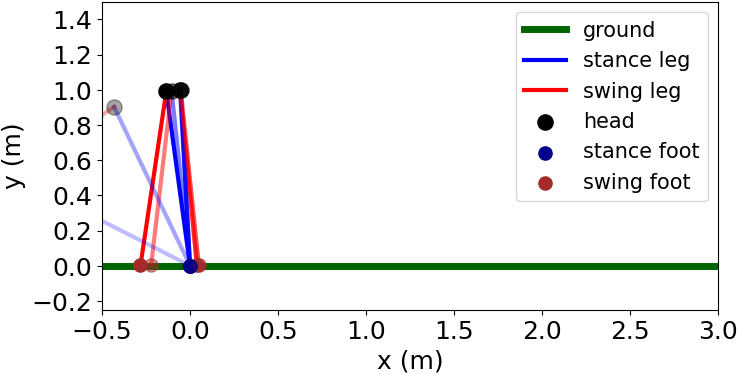}
}
\hfill
\subfigure[Ours]{
\includegraphics[width=0.145\textwidth]{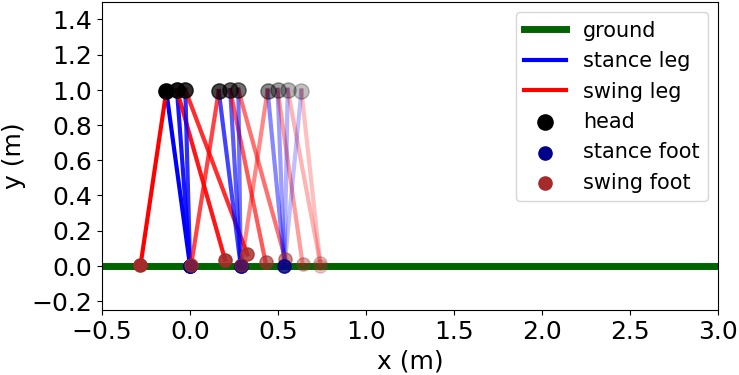}
}
\hfill
}
\end{figure}




\end{document}